# Feature extraction using Latent Dirichlet Allocation and Neural Networks: A case study on movie synopses

**Despoina I. Christou**

Department of Applied Informatics

University of Macedonia

Dissertation submitted for the degree of

*BSc in Applied Informatics*

**2015**

to my parents and my siblings…

# Acknowledgements

I wish to extend my deepest appreciation to my supervisor, Prof. I. Refanidis, with whom I had the chance to work personally for the first time. I always appreciated him as a professor of AI, but I was mostly inspired by his exceptional ethics as human. I am grateful for the discussions we had during my thesis work period as I found them really fruitful to gain different perspectives on my future work and goals.

I am also indebted to Mr. Ioannis Alexander M. Assael, first graduate of Oxford MSc Computer Science, class of 2015, for all his expertise suggestions and continuous assistance.

Finally, I would like to thank my parents, siblings, friends and all other people who with their constant faith in me, they encourage me to pursue my interests.

# Abstract

Feature extraction has gained increasing attention in the field of machine learning, as in order to detect patterns, extract information, or predict future observations from big data, the urge of informative features is crucial. The process of extracting features is highly linked to dimensionality reduction as it implies the transformation of the data from a sparse high-dimensional space, to higher level meaningful abstractions. This dissertation employs Neural Networks for distributed paragraph representations, and Latent Dirichlet Allocation to capture higher level features of paragraph vectors. Although Neural Networks for distributed paragraph representations are considered the state of the art for extracting paragraph vectors, we show that a quick topic analysis model such as Latent Dirichlet Allocation can provide meaningful features too. We evaluate the two methods on the CMU Movie Summary Corpus, a collection of 25,203 movie plot summaries extracted from Wikipedia. Finally, for both approaches, we use K-Nearest Neighbors to discover similar movies, and plot the projected representations using T-Distributed Stochastic Neighbor Embedding to depict the context similarities. These similarities, expressed as movie distances, can be used for movies recommendation. The recommended movies of this approach are compared with the recommended movies from IMDB, which use a collaborative filtering recommendation approach, to show that our two models could constitute either an alternative or a supplementary recommendation approach.

# Contents

# Chapter

# 1. Introduction

## 1.1 Overview

Thanks to the enormous amount of electronic data that is the digitization of old material, the registration of new material, sensor data and both governmental and private digitization intentions in general, the amount of data available of all sorts has been expanding and increasing for the last decade. Simultaneously, the need for automatic data organization tools and search engines has become obvious. Naturally, this has led to an increased scientific interest and activity in related areas such as pattern recognition and dimensionality reduction, fields related mostly to feature extraction. Although the history of text categorization dates back to the introduction of computers, it is only from the early 90's that text categorization has become an important part of the mainstream research of text mining, thanks to the increased application-oriented interest and to the rapid development of more powerful hardware. Categorization has successfully proved its strengths in various contexts, such as automatic document annotation (or indexing), document filtering (spam filtering in particular), automated metadata generation, word sense disambiguation, hierarchical categorization of Web pages and document organization, just to name a few. Probabilistic Models and Neural Networks consist the two state of the art methods to extract features.

The efficiency, scalability and quality of document classification algorithms heavily rely on the representation of documents (Chapter 4). Among the set-theoretical, algebraic and probabilistic approaches, the vector space model (TF – IDF scheme in Section 2.1) representing documents as vectors in a vector space is used most widely. Dimensionality reduction of the term vector space is an important concept that, in addition to increasing efficiency by a more compact document representation, is also capable of removing noise such as synonymy, polysemy or rare term use. Examples of dimensionality reduction include Latent Semantic Analysis (LSA in Section 2.2) and Probabilistic Latent Semantic Analysis (PLSA in Section 2.3). Deerwester et al. in 1990 [1] proposed one of the most basic approaches to topic modelling, called LSA or LSI. This method is based on the theory of linear algebra and uses the bag-of words assumption. The core of the method is to apply SVD to the co-occurrence count matrix of documents and terms (often referred to as the term-document matrix), to obtain a reduced dimensionality representation of the documents. In 1999 Thomas Hofmann [2] suggested a model called probabilistic Latent Semantic Indexing (PLSI/PLSA) in which the topic distributions over words were still estimated from co-occurrence statistics within the documents, but introduced the use of latent topic variables in the model. PLSI is a probabilistic method, and has shown itself superior to LSA in a number of applications, including Information Retrieval (IR). Since then there have been an increasing focus on using probabilistic modelling as a tool rather than using linear algebra. A more interesting approach that can be suitable for low dimensional representation is the generative probabilistic model of text corpora, namely Latent Dirichlet Allocation (LDA) by Blei, Ng, Jordan [3] in 2003 (Section 2.2). LDA models every topic as a distribution over the words of the vocabulary, and every document as a distribution over the topics, thereby one can use the latent topic mixture of a document as a reduced representation. According to Blei et al. [3] PLSI has some shortcomings with regard to overfitting and generation of new documents. This was one of the motivating factors to propose Latent Dirichlet Allocation (LDA) [BNJ03], a model that quickly became very popular, widely used in IR, Data Mining, Natural Language Processing (NLP) and related topics. Probabilistic topic models, such as latent Dirichlet allocation (Blei et al., 2003) and probabilistic latent semantic analysis (Hofmann, 1999, 2001), model documents as finite mixtures of specialized distributions over words, known as topics. An important assumption underlying these topic models is that documents are generated by first choosing a document-specific distribution over topics, and then repeatedly selecting a topic from this distribution and drawing a word from the topic selected. Word order is ignored and each document is modelled as a "bag-of-words". The weakness of this approach, however, is that word order is an important component of document structure, and is not irrelevant to topic modelling. For example, two sentences may have the same unigram statistics but be about quite different topics. Information about the order of words used in each sentence may help disambiguate possible topics. For further information see Chapter 2.

N-gram language models [4] [5] decompose the probability of a string of text, such as a sentence, or document into a product of probabilities of individual words given some number of previous words. Put differently, these models assume that documents are generated by drawing each word from a probability distribution specific to the context consisting of the immediately preceding words, assuming they use local linguistic structure. As to figure out these features, we use the Autoencoder Neural Network (Chapter 3) which is highly used as a machine learning technique that uses paragraph vectors to represent the network input. Paragraph Vectors [6] (see Section 4.2) are closely related to n-grams and are able to capture this disadvantage of the bag-of-words models, the linguistic structure. Precisely, paragraph vector is an unsupervised algorithm that learns fixed-length feature representations from variable-length pieces of texts, such as sentences, paragraphs, and documents. Moreover, Neural Networks can represent dimensionality reduction models, when their hidden layer(s) are of smaller size that the input. Precisely, we use Autoencoder Neural Network, because of its structure that assumes the output to have the same size as the input. This characteristic helps interpreting out data more precisely, as the hidden layer is a compact representation of the data. For further information see Chapter 3 and 4.

Capturing features, either with a probabilistic model or with a Neural Network, it is important to find similarities in the extracted features. K-Nearest Neighbors is such an approach that uses cosine distance as a similarity measure. Furthermore, with T-SNE, a nonlinear dimensionality reduction technique that is well suited for embedding high-dimensional data into a space of two or three dimensions we are able to transfer the extracted features in the 2-dimensional space, where we can visualize the features in a scatter plot. The representing data captures respective distance similarities as the KNN does, but in a more human friendly interpretable way. For further information see Chapter 5 and 6.

## 1.2 Motivation

Feature extraction gains increasing attention in the field of machine learning and pattern recognition, as the urge of informative features is crucial in order to detect patterns and extract information.
Thinking of recommendation systems, the problem comes to the initial information needed as to recommend the relevant object for our scope. At the beginning we do not have data from users while in the following time the user-data may be sparse for certain items [7]. Speaking for movies recommendation, feature extraction from movie plots constitutes a way for clustering relevant movies as to recommend the relevant ones according to their plots, namely their genre. In this thesis we show that feature extraction either from a probabilistic model, such as the state of the art Latent Dirichlet Allocation, or from a Neural Network, namely the Autoencoder, can constitute a significant base for recommendation systems.
Among different recommendation approaches (content-based, collaborative filtering and hybrid [8]), collaborative filtering techniques have been most widely used (IMDB recommendation method), largely because they are domain independent, require minimal, if any, information about user and item features and yet can still achieve accurate predictions [9] [10]. Even though they do manage some prediction, the accuracy of rating predictions is highly increased with content information as proved in [7] [11] [12] [13]. Moreover, in [14] the probabilistic topic model of LDA is used on movie text reviews to show that extracted features from texts can provide additional diversity. Meanwhile, both LDA and Autoencoder can be used as recommendations algorithms. Precisely, in content-based and in collaborative filtering recommendation methods, the algorithms used are divided in memory-based and in model-based approaches. Memory-based methods predict users' preference based on the ratings of other similar users (paragraph vectors-Autoencoder, See Section 4.2), while model-based methods rely on a prediction model by using Clustering (i.e. LDA) [15].
In this thesis, we want to testify that LDA and Autoencoder, both used as methods for feature extraction, present astonishing movie recommendations, reaffirming that a system which makes use of content might be able to make predictions for this movie even in the absence of ratings [16]. Moreover, we compare the two methods with the recommending results of the collaborating filtering process of IMDB as to discover a comparison among them (see Conclusions, Chapter 7).

# Chapter 2

# 2. Latent Dirichlet Allocation

Topic modeling is a classic problem in natural language processing and machine learning. In this chapter we present Latent Dirichlet Allocation, one of the most successful generative latent topic models developed by Blei, Ng and Jordan [3]. Latent Dirichlet Allocation (LDA) can be a useful tool for the statistical analysis of document collections and other discrete data. Specifically, LDA is an unsupervised graphical model which can discover latent topics in unlabeled data [17]. This exact characteristic of LDA renders this model prior to collaborative models, as for the process of filtering the information of large datasets, the "many users" needed for the collaborative process are difficult to exist from the beginning.

## 2.1 History

The main inspiration behind LDA is to find a proper modeling framework of the given discrete data, namely text corpora in this thesis. The most essential objective is the dimensionality reduction of our data, so to ensure efficient backings productive operations, such as information retrieval, clustering, etc. Baeza-Yates and Ribeiro-Neto, 1999 have made significant progress in this field. What IR researchers proposed was to reduce every document in our corpus to a vector of real numbers, each of which represents ratios of tallies. Below, we review the three most compelling methodologies, which preceded the LDA model.

### 2.1.1 TF – IDF scheme

*Term Frequency – Inverse Document Frequency* scheme, is the most widely used weighting scheme in the vector space model, which was introduced in 1975 by G. Salton, A. Wong, and A. C. S. Yang [18] as a model for automatic indexing. Vector space model is an algebraic model for representing text documents as vectors of identifiers, for example, index terms, filtered information, etc. TF-IDF scheme is used in information filtering, information retrieval, indexing and relevant rankings.
In Vector Space Model, the corpus is represented as follows:

$$D \equiv \{\vec{d}_i\}_{i=1}^{M} \tag{2.1.1}$$

Here, the documents are represented as finite dimensional vectors, where $\vec{d}_i = (w_{i1}, \ldots, w_{iV})$ and V the vocabulary size. The weight $w_{ij}$, represents how much term $t_j$ contributes to the content of the document $\vec{d}_i$ and is equal to the tf-idf value.

TD - IDF scheme models every document of the corpus as real valued vectors. The weights reflect the contribution to the content of the document. Its value increases proportionally to the number of times a word shows up in the archive, yet is balanced by the frequency of the word in the corpus, which serves to adjust for the fact that some words appear more often in general.

*Term Frequency*: the number of times a term occurs in a document

*Inverse Document Frequency*: terms that occur very frequently in a document lessen the weight of this term and relatively terms that occur rarely increase the weight of the corresponding term.

Term Frequency – Inverse Document Frequency (TF - IDF) is the product of two statistics, the Term Frequency and the Inverse Document Frequency. The procedure for implementing the tf-idf scheme has some minor differences concurring its application, but generally we can assume the tf-idf value, as follows:

$$t_d = f_{t,d} * \log\left(\frac{D}{f_{t,D}}\right) \tag{2.1.2}$$

where $f_{t,d}$ equals the number of times term t appears in document d, D is the size of the corpus, and $f_{t,D}$ equals the number of documents in which t appears in D [19].

Once, the queries have all been Boolean, implying that documents either match one term or they do not. Answering the problem of too few or too many results, rank-order the documents in a collection with respect to a query was needed. There comes the TF-IDF. Terms with high TF-IDF numbers imply a strong relationship with the document they appear in, suggesting that if that term were to appear in a query for example, the document could be of interest to the user. For instance if we want to find a term highly related to a document, the $f_{t,d}$ should be large, while the $f_{t,D}$ should be small. In that way, the $\log\left(\frac{D}{f_{t,D}}\right)$ would be rather large and subsequently $t_d$ would be likewise large. This case implies that t is an important term in document d but not in the corpus D, so term t has a large discriminatory power. Next important case is the $t_d$ to be equal to zero, with the basic definition declaring that this term occurs in all documents.

The similitude of two documents (or a document and a query) can be found in several ways using the tf-idf weights with the most common one the cosine similarity.

$$sim(\vec{d}_i, \vec{d}_j) = \frac{\vec{d}_i \cdot \vec{d}_j}{\|\vec{d}_i\|\|\vec{d}_j\|} \tag{2.1.3}$$

In a typical information retrieval context, given a query and a corpus, the assignment is to find the most relevant documents from the collection to the query. The first step is to calculate the weighted tf-idf vectors to represent each documents and the query, then to compute the cosine similarity score for these vectors and finally rank and return the documents with the highest score with respect to the query.

Disadvantage consist the fact that this representation reveals little in the way of inter- or intra-document statistical structure, and the intention of reducing the description length of documents is only mildly alleviated. One way of penalizing the term weights for a document in accordance with its length can be the document length normalization.

### 2.1.2 Latent Semantic Analysis (LSA)

Latent Semantic Analysis (LSA) introduced in 1990 by Deerwester et al. [1] is a method in Natural Language Processing, particularly a vector space model, for extracting and representing the contextual significance of words by statistical computations applied to a large corpus. It is also known as Latent Semantic Indexing (LSI).

LSA has proved to be a valuable tool in many areas of NLP and IR and has been used for finding and organizing search results, summarization, clustering documents, spam filtering, speech recognition, patent searches, automated essay evaluation (i.e. PTE tests), etc.

The intuition behind LSA is that words that are close in meaning will occur in similar pieces of text, so by extracting such relationships among words, documents and corpus we can assume words into coherent passages.

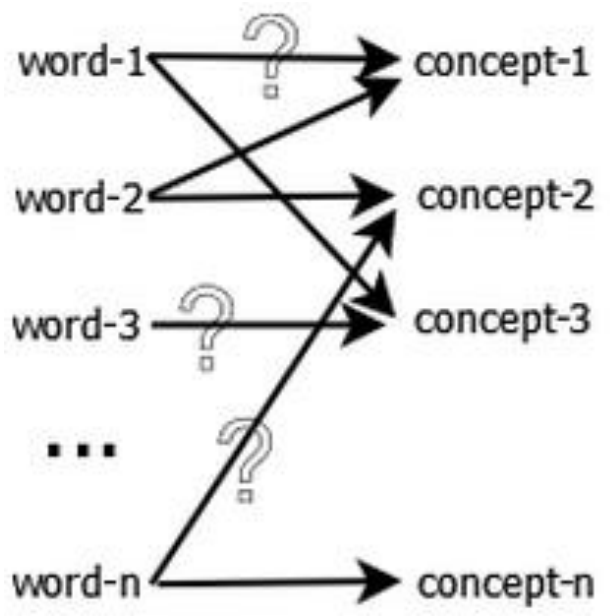

*Figure 1: Words correlated to coherent texts [20]*

As we notice in *Figure 1*, a term can correspond to multiple concepts as languages have different words that mean the same thing (synonyms), words with multiple meanings, and many ambiguities that obscure the concepts to the point where people can have a hard time understanding.

LSA endeavors to solve this problem by mapping both words and documents into a latent semantic space and transferring the comparison in this space.

| Index Words | Titles | | | | | | | | |
|---|---|---|---|---|---|---|---|---|---|
| | T1 | T2 | T3 | T4 | T5 | T6 | T7 | T8 | T9 |
| book | | | 1 | 1 | | | | | |
| dads | | | | | | 1 | | | 1 |
| dummies | | 1 | | | | | | 1 | |
| estate | | | | | | | 1 | | 1 |
| guide | 1 | | | | | 1 | | | |
| investing | 1 | 1 | 1 | 1 | 1 | 1 | 1 | 1 | 1 |
| market | 1 | | 1 | | | | | | |
| real | | | | | | | 1 | | 1 |
| rich | | | | | | 2 | | | 1 |
| stock | 1 | | 1 | | | | | 1 | |
| value | | | | 1 | 1 | | | | |

*Figure 2: Count Matrix [1]*

Firstly, a *Count Matrix* is constructed. The Count Matrix is a word by document matrix, where each index word is a row and each title is a column, with each cell representing the number of times that word occurs in the document. The matrices generated though this step tend to be very large, but also very sparse. In this point we highlight that LSA removes the most frequent used words, known as stop words, given better results than previous vector models, as its vocabulary consist of words that contribute much meaning.

Secondly, raw counts in the matrix at Figure 2, are *modified with TF-IDF*, as to weight more the rare words and less the most common ones. (See Section 2.1.1)

Once the count matrix is ready, we apply the *Singular Value Decomposition* (SVD) method to the matrix. SDM is a method of *dimensionality reduction* of our matrix, namely it reconstructs our matrix with the least possible information, by throwing away the noise and maintaining the strong patterns.
The SVD of a matrix D is defined as the product of three matrices:

$$D = U\Sigma V^T \tag{2.1.2.1}$$

$$D(n*m) = U(n*n)\Sigma(n*m)V(m*m) \tag{2.1.2.2}$$

, where U matrix represents the coordinates of each word in our latent space, the $V^T$ matrix stands for the coordinates of each document in our latent space, and the Σ matrix of singular values is a hint of the dimensions or "concepts" we need to include. From the Σ matrix we keep only k (number of concepts) Eigen values, a procedure reflecting the major associative patterns in the data, while ignoring the less important influence and noise. (2.1.2.2) equation becomes as follows:

$$D(n*m) = U(n*k)\Sigma(k*k)V(k*m) \tag{2.1.2.3}$$

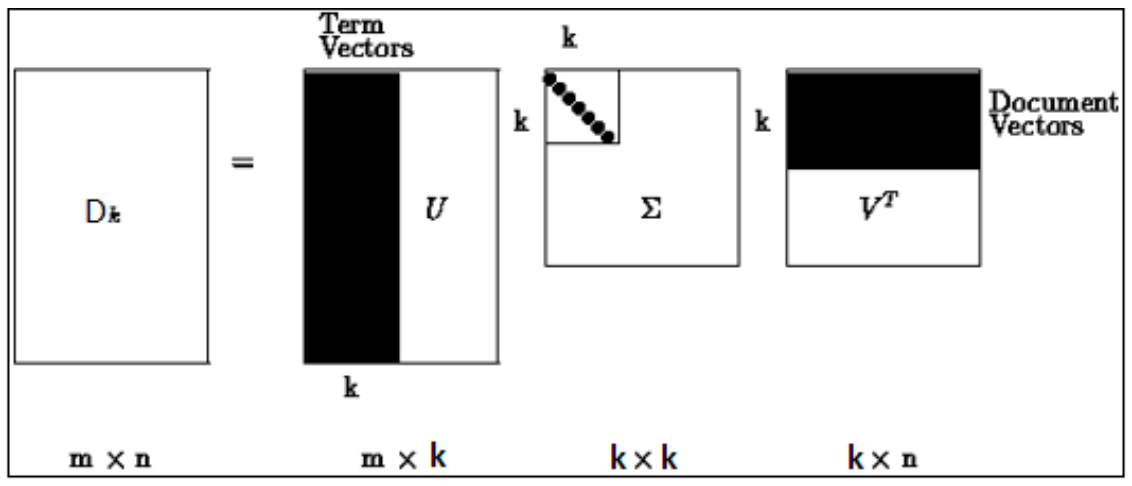

*Figure 3: Dimension Reduction with SVD [1]*

Lastly, terms and documents are converted to point s in k-dimensional space and given the reduced space, we can compute the similarity between doc-doc, word-word (synonymy and polysemy), and word-document, by clustering our data.

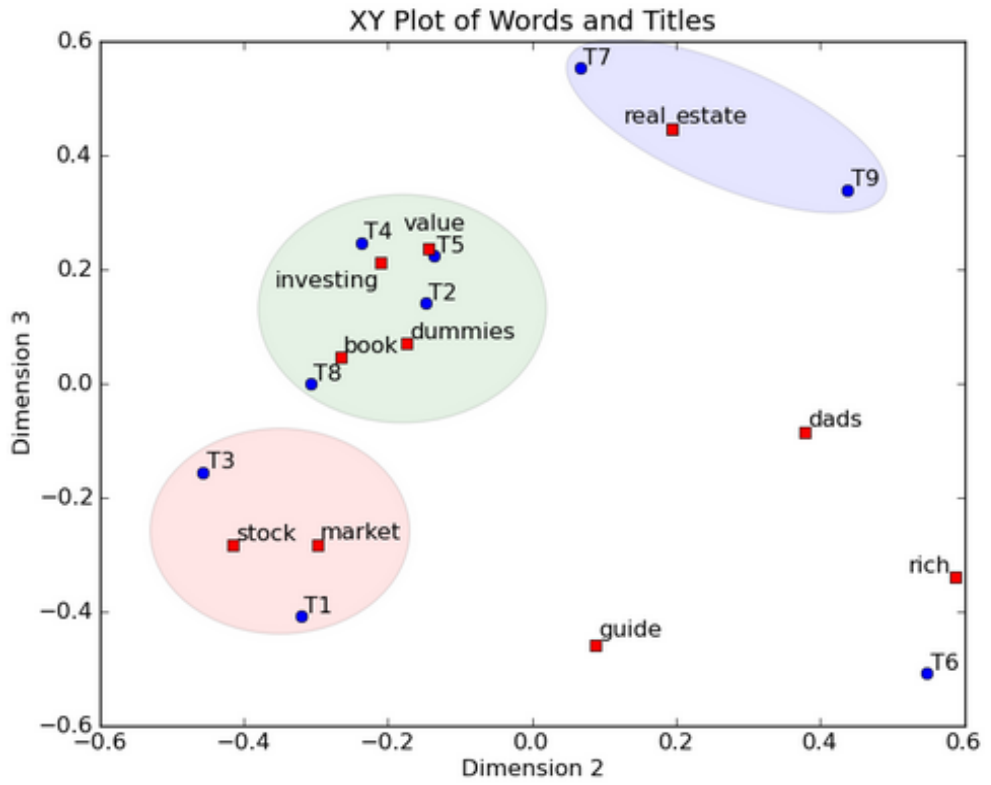

*Figure 4: Clustering*

Moreover, in the Information Retrieval field we can find similar documents across languages, after analyzing a base set of translated documents and we can also find coherent documents to a query of terms after its conversion into the low-dimensional space. A challenge to be addressed in vector space models is twofold: synonymy and polysemy [2]. Synonymy leads to poor recall as they will have small cosine but related, while polysemy concludes to poor precision as they will have large cosine, but not truly related. LSA as a vector space model cannot retrieve documents coherent to a query unless they have common terms. LSA partially address this, as the query and document are transformed in a lower dimensional "concept" space and are represented by a similar weighted combination of the SVD variables.

However, LSA even though it manages to extract proper correlations of words and documents, it handles the text with the bag of words model (BOW), making no use of word order, syntax and morphology. Another disadvantage is the highly connection with SVD, a computationally intensive method, having O(n^2k^3) complexity. Moreover, SVD hypothesizes that words and documents follow the normal (Gaussian) distribution, when a Poison distribution is usually observed. Nevertheless, we can better perform many of the above using Probabilistic Latent Semantic Analysis (PLSA), a preferred method over LSA.

### 2.1.3 Probabilistic Latent Semantic Analysis (PLSA)

Probabilistic Latent Semantic Analysis (PLSA), or Probabilistic Latent Semantic Indexing (PLSI), in the field of information retrieval) embeds the idea of LSA into a probabilistic framework. Contradictory to LSA, which is a linear algebra method, PLSA sets its foundations in statistical inference.
PLSA originated in the domain of statistics in 1999 by Thomas Hofmann [2], and later on 2003 in the domain of machine learning by Blei, Ng, Jordan [3]. PLSA is considered as a generative model in which the topic distributions over words are still estimated from co-occurrence statistics within the documents as in LSA, but the use of latent topic variables is introduced in the model. Probabilistic Latent Semantic Analysis has its most prominently applications in information retrieval, natural language processing and machine learning. From 2012 PLSA has been also used in the bioinformatics.
PLSA practices a generative latent class model to accomplish a probabilistic mixture decomposition. It models every document with a topic-mixture. This mixture is assigned to the documents individually, without a generative process and the mixture weights (parameters) are learned by expectation maximization.
*The aspect model*: The aspect model is the statistical model in which PLSA relies on (Hofmann, Puzicha, & Jordan, 1999). It assumes that every document is a mixture of latent (K) aspects. The aspect model is a latent variable model for co-occurrence data which associates an unobserved class variable $z_k \in (z_1, \dots, z_K)$ with each observation, which is the occurrence of a word in a particular document.

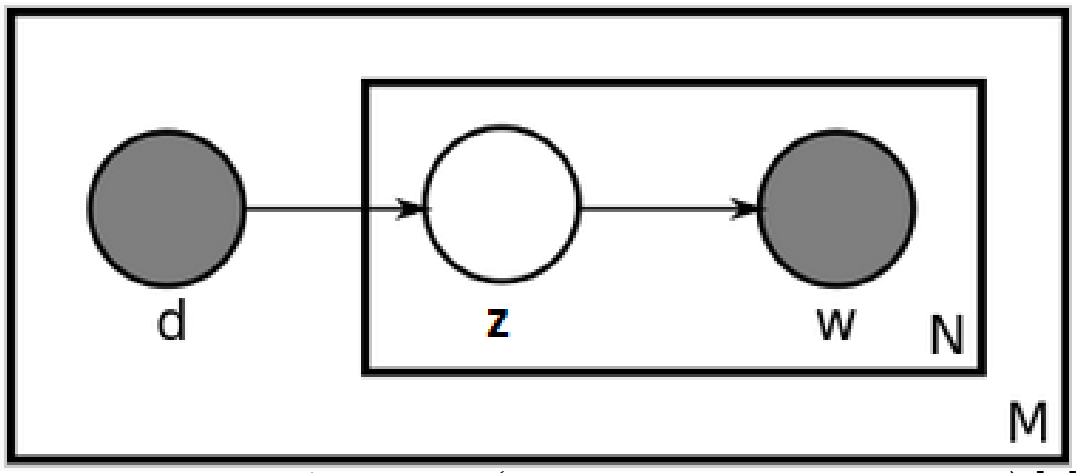

*Figure 5: PLSA model (asymmetric formulation)* [3]

Figure 5 illustrates the *generative model* for the word-document co-occurrence. First PLSA selects a document $d_i$, with probability $P(d_i)$, then is picks a latent variable $z_k$ with probability $P(z_K, d_i)$ and finally generates a word $w_j$ with probability $P(w_j, z_k)$. The result is a joint distribution model, presented in the expression:

$$P(w_j | d_i) = \sum_{k=1}^{K} P(z_k | d_i) \, P(w_j | z_k) \qquad (2.1.3.1)$$

The generation process assumes that for each document $d_i \in (d_1, \dots, d_M)$, for each token position $j \in (1, \dots, N)$ we first pick o topic z from the multinomial distribution $P(z_k|d_i)$ and then we choose a term w from the multinomial distribution $P(\, w_j|z_k)$.

The model can be equally parameterized by a perfectly symmetric model for documents and words. We can see this below:

$$P(\, w_j|d_i) = \sum_{z=1}^{Z} P(\mathrm{z})P(d|z)\,P(w|z) \qquad (2.1.3.2)$$

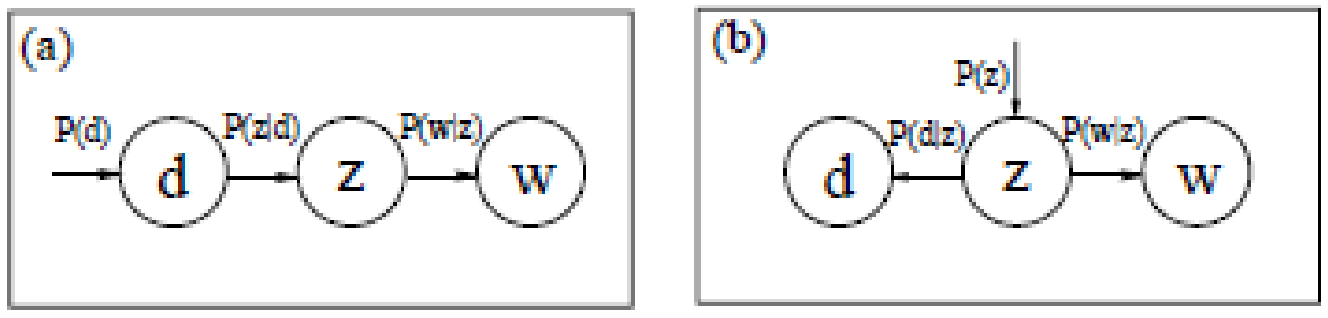

*Figure 6: asymmetric (a) and symmetric (b) PLSA representation* [2]

PLSA as a statistical latent variable model introduces a conditional independence assumption, namely that d and w are independent set as for the associated latent variable. The parameters of this model are $P(\, w_j|z_k)$ and $P(z_k|d_i)$ so according to the above conditional independence assumption the number of parameters equals $wz + zd$, namely it grows linearly with the number of documents.
The parameters are estimated by Maximum Likelihood, with the *Expectation Maximization* (EM) be the typical procedure (Dempster, Laird, & Rubin, 1977). EM substitutes two steps: the Expectation (E) step where the posterior probabilities of the latent variables are calculated and the Maximization (M) step where the parameters are redesigned. The (E)-step implies to apply the Bayes' formula in the (2.1.3.1) equation and (M)-step calculates the expected complete data-log-likelihood, which depends on the outcome of the first step as to update the parameters.
(E)-step:

$$P(z_k|d_i,\, w_j) = \frac{P(\, w_j|z_k)P(z_k|d_i)}{\sum_{i=1}^{K} P(\, w_j|z_i)P(z_i|d_i)}, \qquad (2.1.3.3)$$

(M)-step:

$$\mathbf{E}\left[\mathcal{L}^c\right] = \sum_{i=1}^{N}\sum_{j=1}^{M} n(d_i, w_j) \sum_{k=1}^{K} P(z_k \mid d_i, w_j)\, \log\left[P(w_j \mid z_k)P(z_k \mid d_i)\right] \qquad (2.1.3.4)$$

The two steps are alternated until a termination condition. This can be either the convergence of the parameters or an early stopping, namely the cancel of updating the parameters when their performance is not improving.

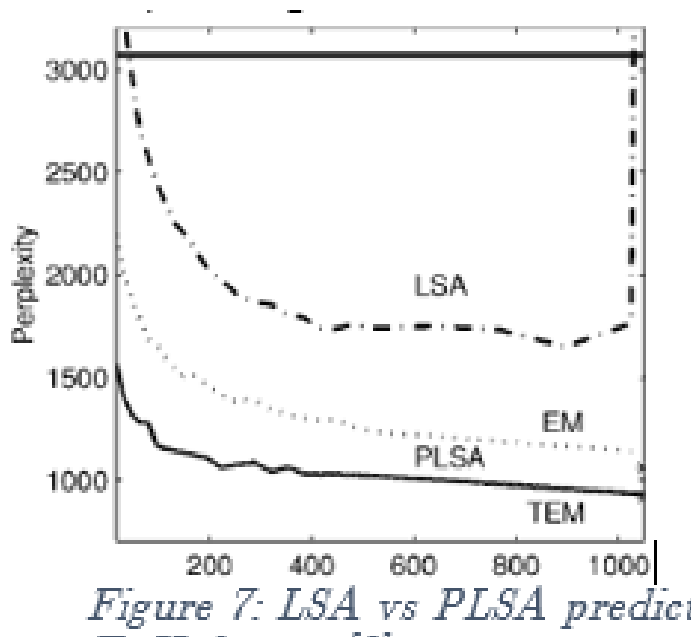

*Figure 7: LSA vs PLSA prediction, T. Hofmann [6]*

PLSA was influenced by LSA, as a solution for the unsatisfactory statistical foundation. The main difference between the two is the objective function; LSA is based on a Gaussian assumption, while PLSA relies on the likelihood function of multinomial sampling and aims at an explicit maximization of the predictive power of the model. In Figure 7 [2], we notice the difference in LSA's and PLSA's perplexity in a collection of 1033 documents, clearly revealing PLSA's predictive supremacy over LSA. As noted in the disadvantages of LSA in Section 2.1.2, the Gaussian distribution hardly stands for count data. As for the dimensionality reduction, PLSA performs its method by having K aspects, while LSA keeps only K singular values for its method. Consequently, when the selection of proper value of K in LSA is heuristic, PLSA's model selection (aspect model) can determine an optimal K maximizing its predictive power. Regarding the computational complexity of LSA (O (n^2k^3)) and PLSA (O (n*k*i), with i counting the iterations), the SVD in LSA can be computed precisely, while EM is an iterative method which only affirms to find a local maximum of the likelihood function. However Hofmann's experiments have shown that when using regularization techniques, as 'early stopping' in order to avoid the overfitting, presented both in global and local maximums, even a "poor" local maximum in PLSA might performs better than the LSA's solution.

## 2.2 Latent Dirichlet Allocation (LDA)

Latent Dirichlet Allocation (LDA), a generative probabilistic topic model for collections of discrete data, is the most representative example of topic modeling and it was first presented as a graphical model for topic discovery by Blei, Ng, Jordan in 2003 [3], when they put into question the Hofmann's PLSI model [2]. Due to its high modularity, it can be easily extended, giving much interest to its study. LDA is also an instant of a mixed membership model [21], so according to the Bayesian analysis as each document is associated with a distribution, it can be associated with multiple components respectively. LDA as a topic model intends on detecting the thematic information of large archives and can be adapted as to find patterns in generic data, images, social networks, and bioinformatics.

### 2.2.1 LDA intuition

Years ago the main issue was to find information. Nowadays, human knowledge and behavior are in high percentage digitized in the web and more precisely in scientific articles, books, blogs, emails, images, sound, video, and social networks, transferring the issue on how to interact with these electronic archives each time yet more efficiently. Here comes the topic modeling, with LDA the most characteristic topic modeling algorithm.

*Topic Modeling* does not require any prior annotations or labeling of the archives as topic models are able to detect patterns in an unstructured collection of documents and to organize these electronic archives at a scale that would be inconceivable by human annotation. Topic models as statistical models have their origins to PLSI, created by Hofmann in 1999 [2]. Latent Dirichlet Allocation (LDA) [3] an amelioration of PLSI, is probably the most common topic model currently in use as others topic models are generally extensions on LDA.

*The objective of LDA is* to discover short descriptions of the collection's individuals, reducing the plethora of the initial information into a smaller interpretable space, while keeping the essential statistical dependences to facilitate efficient processing of the data. If speaking for text corpora as our collection, Latent Dirichlet Allocation assumes *that each document of the collection exhibits multiple topics* [3], with the topic probabilities providing an explicit representation of the corpus.

LDA assumes that all documents in the collection share the same set of topics, but each document exhibits those topics with different proportion. We can better understand LDA's objective in Blei's representation of the hidden topics in a scientific document in Figure 8 below:

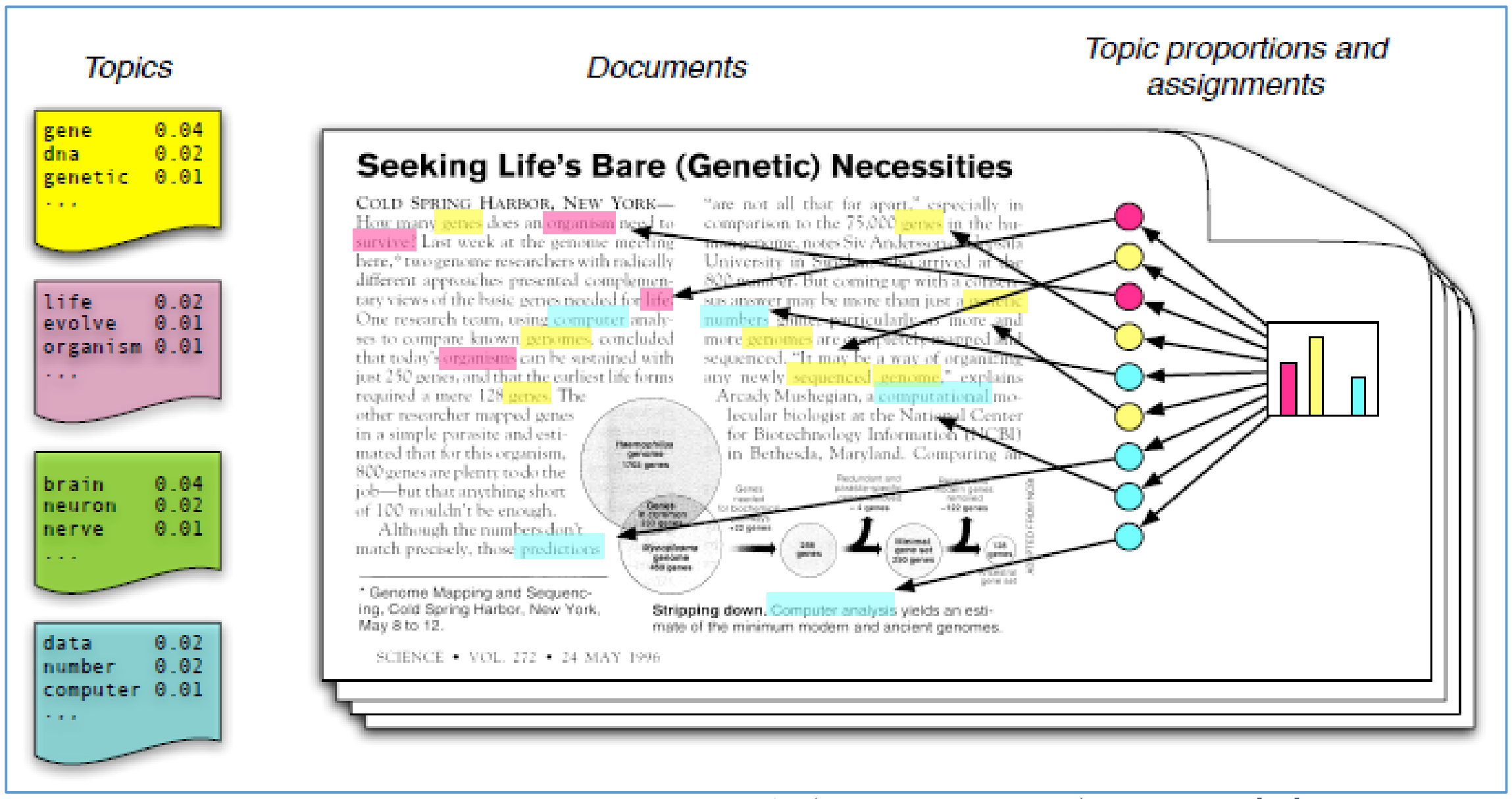

*Figure 8: The intuition behind LDA. (Generative process) by D. Blei [17]*

In Figure 8, words from different topics are highlighted with different colors. For example yellow represents words about genetics, pink words about evolutionary life, green words about neurons and blue words about data analysis, declaring the nature of the article and the mixture of multiple topics which is the intuition of the model. LDA as a generative probabilistic model treats data as observations that arrive from a generative probabilistic process that includes hidden variables. These hidden or latent variables reflect the thematic structure of the collection, the topics, with the histogram at right in Figure 8 representing the document's topic distribution. Moreover, the word mixture of each topic is represented at left, also given the probabilities of each word's exhibition in every topic.

This procedure assumes that the order of words does not necessarily matter. Even though documents would be unreadable with their words shuffled, we are able to discover the thematically coherence terms, we are interested about (see Section 2.2.2 about bag of words). At this sense, same words can be observed in multiple topics but with different probabilities. Every topic contains a probability for every word, so even though one word does not have high probability in one topic, it might have a greater one in another.

The main intuition behind LDA as said above is that documents exhibit multiple topics. In Figure 8, one can notice that documents in LDA are modeled by assuming we know the topic mixture of all documents and the word mixture of all the topics. Having this knowledge, the model assumes that the document at first is empty, then choose a topic from the topic mixture and then a word from the word mixture of that topic, repeating this process until the document is shaped. This exact procedure is repeated for every document in the corpus. This process constitutes the generative character of the LDA model, namely its attribute of generating observations, given some hidden variables.

The generative process of LDA can be seen in the following algorithm [22]:

```
1: "topic plate"
2: for all topics k ∈ [1, K] do
3:     sample mixture components φ⃗_k ~ Dir(β⃗)
4: "document plate"
5: for all documents m ∈ [1, M] do
6:     sample mixture proportion ϑ⃗_m ~ Dir(α⃗)
7:     "word plate"
8:     for all words n ∈ [1, N_m] in document m do
9:         sample topic index z_{m,n} ~ Multinomial(ϑ⃗_m)
10:        sample term for word w_{m,n} ~ Multinomial(φ⃗_{z_{m,n}})
11: return
```

*Algorithm 1: Generative process of LDA*

| | |
|---|---|
| $M$ | number of documents in the corpus |
| $K$ | number of latent topics/ mixture components |
| $V$ | number of terms $t$ in vocabulary |
| $N$ | number of words in the corpus, i.e. $N = \sum_{m=1}^{M} N_m$ |
| $\vec{\alpha}$ | hyperparameter on the mixing proportions |
| $\vec{\beta}$ | hyperparameter on the mixture components |
| $\vec{\vartheta}_m$ | parameter notation for $p(z\|d = m)$, the mixture component of topic $k$. One component for each topic, $\underline{\Theta} = \{\vec{\vartheta}_m\}_{m=1}^{M}$ |
| $\vec{\varphi}_k$ | parameter notation for $p(t\|z = k)$, the topic mixture proportion for document $m$. One proportion for each document, $\underline{\Phi} = \{\vec{\varphi}_k\}_{k=1}^{K}$ |
| $N_m$ | document length (document-specific), modeled with a Poisson distribution with constant parameter $\xi$ |
| $z_{m,n}$ | mixture indicator that chooses the topic for the $n^{\text{th}}$ word in document $m$ |
| $w_{m,n}$ | term indicator for the $n^{\text{th}}$ word in document $m$ |

*Figure 9: LDA notation*

The algorithm has no information about the topics. The inferred topic distributions are generated by computing the hidden structure from the observed documents. The problem of inferring the latent topic proportions is translated to compute the posterior distribution of the hidden variables given the documents.

For the generative process of LDA first we have to model each document with a Poisson distribution over the words. This first step is not presented in the Algorithm 1, but is considered as known in line 8 and it can also be seen in LDA's notation in *Figure 9*. Second, for each topic we assume the random variable $\vec{\varphi}_k$, a dirichlet distribution of the words in topic k, parameterized by $\vec{\beta}$, a V-dimension vector of positive reals, summing up to one, that is to be estimated. Third, for each document we compute the random variable $\vec{\theta}_m$, a dirichlet distribution of the topics occurring in document m, parameterized by $\vec{\alpha}$, a K-dimension vector of positive reals, summing up to one. Then for each document, the following steps are repeated for every word n of document m: First we identify the topic of the word as a multinomial distribution given $\vec{\theta}_m$, the topic distribution of document m and secondly we choose a term from $\vec{\varphi}_k$, the word distribution of the chosen topic in the forward step. These two steps show that documents exhibit the latent topics in different proportions, while each word is drown from one of the topics. Finally this process is repeated for all the documents in our corpus with the goal to estimate the posterior (conditional) distribution of the hidden variables by first defining the joint probability distribution of the observed and the latent random variables.

The generative process for LDA corresponds to the following joint distribution of the latent and the observed variables for a document m:

$$p(\vec{w}_m, \vec{z}_m, \vec{\vartheta}_m, \underline{\Phi}|\vec{\alpha}, \vec{\beta}) = \overbrace{\underbrace{\prod_{n=1}^{N_m} p(w_{m,n}|\vec{\varphi}_{z_{m,n}})p(z_{m,n}|\vec{\vartheta}_m)}_{\text{word plate}} \cdot p(\vec{\vartheta}_m|\vec{\alpha})}^{\text{document plate (1 document)}} \cdot \underbrace{p(\underline{\Phi}|\vec{\beta})}_{\text{topic plate}} \qquad (2.2.1.1)$$

where the $w_{m,n}$ are the only observable variables, namely a bag of words organized by document m each time, with all the others being the latent ones. Equation (2.2.1.1) denotes there are three levels to the LDA representation. The hyperparameters α and β are corpus level parameters, assumed to be sampled once in the process of generating a corpus. The variable $\vec{\theta}_m$ are document-level variables, computed once per document. Finally, the variables $z_{m,n}$ and $w_{m,n}$ are word level variables and are sampled once for each word in each document smoothed by p$\left(\Phi|\vec{\beta}\right)$.and $w_{m,n}$ are word level variables and are sampled once for each word in each document smoothed by p$\left(\Phi|\vec{\beta}\right)$.

The joint distribution for the whole corpus is described by the (2.2.1.2):

$$\prod_{k=1}^{K} p(\vec{\varphi}_k|\vec{\beta}) \left(\prod_{m=1}^{M} p(\vec{\theta}_m|\vec{\alpha}) \left(\prod_{n=1}^{N} p(\vec{z}_{m,n}|\vec{\theta}_m)\, p(\vec{w}_{m,n}|\vec{\varphi}_{zm,n})\right)\right) \qquad (2.2.1.2)$$

The interpretation of the (2.2.1.1) equation can be better understood by LDA's graphical representation, first presented as such model by Blei et al. [3]:

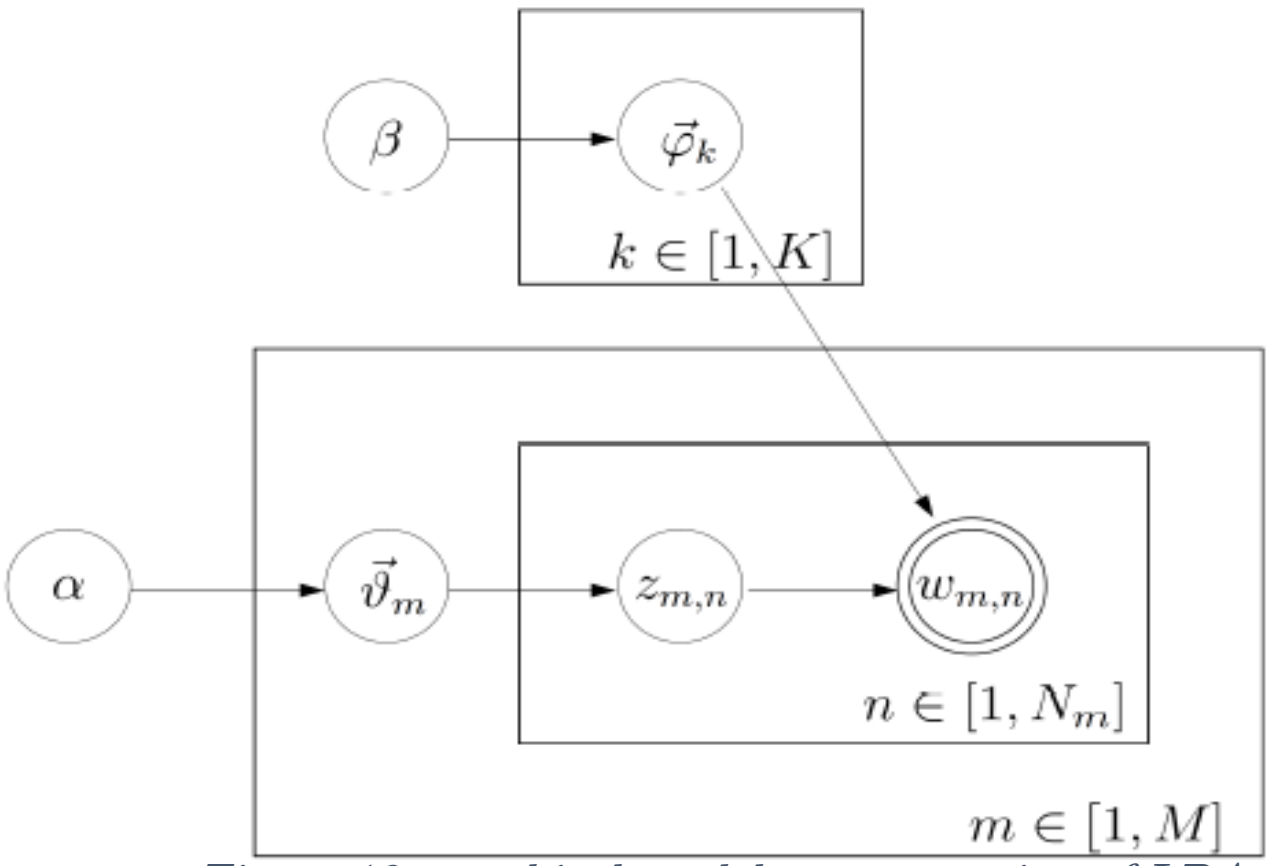

*Figure 10: graphical model representation of LDA*

LDA as a probabilistic graphical model shows the statistical assumptions, behind the generative process described above, which relies on the Bayesian Networks. In Figure 10, the boxes plates represent loops.

The outer plates represent the documents, while the inner plates represent the repeated choice of topics and words within a document [3] [23]. Moreover every single circled node declares a hidden variable, while the unique double circled node of w represents and the only observed variable, the words of the corpus. Figure 10 illustrates the conditional dependences that define the LDA model. Specifically, the topic indicator $\vec{z}_{m,n}$ depends on $\vec{\theta}_m$, the topic proportions of document m. Futhermore, the observed word $\vec{w}_{m,n}$ depends on $\vec{z}_{m,n}$ , the topic indicator and $\vec{\varphi}_k$, the word distribution of the topic that $\vec{z}_{m,n}$ indicates.

The choice of the topic assignment $z_{m,n}$ and the choice of n word from the word distribution of topic m, $\vec{w}_{m,n}$ , are represented as multinomial distributions of $\vec{\theta}_m$ and $\vec{\varphi}_k$ respectively. Here, the *multinomial distribution* is assigned to the Multinomial with just one trial. In this case the multinomial distribution is equivalent to the categorical distribution. In probability theory and statistics, a categorical distribution (also called a "generalized Bernoulli distribution" or, less precisely, a "discrete distribution") is a probability distribution that describes the result of a random event that can take on one of K possible outcomes, with the probability of each outcome separately specified. There is not necessarily an underlying ordering of these outcomes, but numerical labels are attached for convenience in describing the distribution, often in the range 1 to K. Note that the K-dimensional categorical distribution is the most general distribution over a K-way event; any other discrete distribution over a size-K sample space is a special case. The parameters specifying the probabilities of each possible outcome are constrained only by the fact that each must be in the range 0 to 1, and all must sum to 1.

In the LDA model, we assume that the topic mixture proportions for each document are drawn from some distribution. So, what we want is a distribution on multinomials. That is, k-tuples of non-negative numbers that sum to one. We can represent the topic proportions of a document m, $\vec{\theta}_m$, and and the word distributions of a topic k, $\vec{\varphi}_k$, as a M-1 topic simplex and V-1 vocabulary simplex repsectively. As a simplex we assume the geometric interpretation of all these multinomials. More precisely we assume that the single values of θ and ϕ in Figure 10 come from a dirichlet distribution.
A *dirichlet distribution* is conjugate to the multinomial distribution parameterized by a vector α of positive reals with its density function giving the belief that the probabilities of K rival events are $x_i$ given that each event has been observed $\alpha_i$-1 times. In Bayesian statisitcs, a dirichlet distribution is used as a prior distribution, namely a probability distribution of an event we have no evidence about.

The representation of the k-dimensional Dirichlet random variable θ and of V-dimensional Dirichlet random variable ϕ in the simplex, is deeper analyzed in Section 5.4. as to understand how the concetration parameters α and β can influence the results of LDA

## 2.2.2 LDA and Probabilistic models

Latent Dirichlet Allocation was introduced by Blei et al. [3] as a probabilistic graphical model. In this Section we describe the basic disciplines that prevail in the probabilistic models. As described in Section 2.2.1 LDA follows a generative process to discover the hidden variables by ascribing statistical inference to the data. Specifically, using statistical inference we can invert the generative process and obtain a probability distribution of a document's topic mixture. These principles – *generative processes, hidden variables, and statistical inference* – are the foundation of probabilistic models [24].

Probability theory holds the foundations as to model our beliefs about different possible states of a situation, and to re-estimate them when new evidence comes to the forefront. Even though probability theory has existed since the 17th century, our ability to use it effectively in large problems involving many inter-related variables is fairly recent, and is due to the development of a framework known as Probabilistic Graphical Models (PGMs). LDA constitutes the state of the art such model and is subsequent to the Probabilistic LSA.

Both LDA and PLSA, as most of the probabilistic models rely on the *"bag of words"* assumption that is the order of words is irrelevant. Many words, especially nouns and verbs that only seldom occur outside a limited number of contexts, have one specific meaning or at least only a few, not depending on the position in the text. In this basis the bag-of-words assumption captures the topic mixture and can effectively describe the topical aspects of the document collection.

This "bag of words" assumption equals the exchangeability property.

*Exchangeability property* assumes that the joint distribution is invariant to permutation. If p is a permutation of the integers from 1 to N, we can say that a finite set of random variables $\{z_1, \dots, z_N\}$ is said to be exchangeable [25]:

$$\mathrm{p}(z_1, \dots, z_N) = \mathrm{p}((z_{\pi(1)}, \dots, z_{\pi(N)})) \qquad (2.2.21)$$

where $\pi\,(\cdot)$ is a permutation function on the integers $\{1 \dots \mathrm{N}\}$.

Respectively, an infinite sequence of random variables is infinitely exchangeable if every finite subsequence is exchangeable.

De Finetti's representation theorem [26] states that the joint distribution of an infinitely exchangeable sequence of random variables is as if a random parameter were drawn from some distribution and then the random variables in question were independent and identically distributed, conditioned on that parameter. Note that an assumption of exchangeability is weaker than the assumption that the random variables are independent and identically distributed. Rather, exchangeability essentially can be interpreted as meaning "conditionally independent and identically distributed", where the conditioning is with respect to an underlying latent parameter of a probability distribution. That is to say that the conditional probability distribution $p(x_1, \dots, x_k|\Theta) = \prod_{i=1}^{K} p(x_i|\Theta)$ is easy to express, while the joint distribution usually cannot be decomposed.

We are going to examine LDA respectively to its precedent probabilistic model PLSA as to better view the process of the probabilities models in time, as also to highlight the advantages of LDA over the others similar models.
First, the exchangeability property was exploited in PLSA only on the level of words. As one can argue that not only words but also documents are exchangeable, a probabilistic model should capture the exchangeability of both words and documents. LDA evolved from PLSA by extending the exchangeability property to the level of documents by applying Dirichlet priors on the multinomial distributions $\vec{\theta}_m$ $\vec{\varphi}_k$.

LDA assumes that words are generated by the fixed conditional distributions over topics with those topics be infinitely exchangeable within a document. By de Finetti's theorem, the probability of a sequence of words and topics must therefore have the form:

$$\mathrm{p(w, z)} = \int p(\theta)(\prod_{n=1}^{N} \sum p(z_n|\theta)p(w_n|z_n))d\theta \qquad (2.2.2.2)$$

where θ is the random parameter of a multinomial over topics.

As noted by Blei, et al. [3] *"PLSI is incomplete in that it provides no probabilistic model at the level of documents. In PLSI, each document is represented as a list of numbers (the mixing proportions for topics), and there is no generative probabilistic model for these numbers."* Latent Dirichlet allocation models the documents as finite mixtures over an underlying set of latent topics with specialized distributions over words which are inferred from correlations between words, independently of the word order (bag of words). Namely, LDA appends a Dirichlet prior on the per-document topic distribution as to address the criticized inefficiency of PLSA inferred above.
Let notice the probabilistic graphical model of the two models to infer the differences.

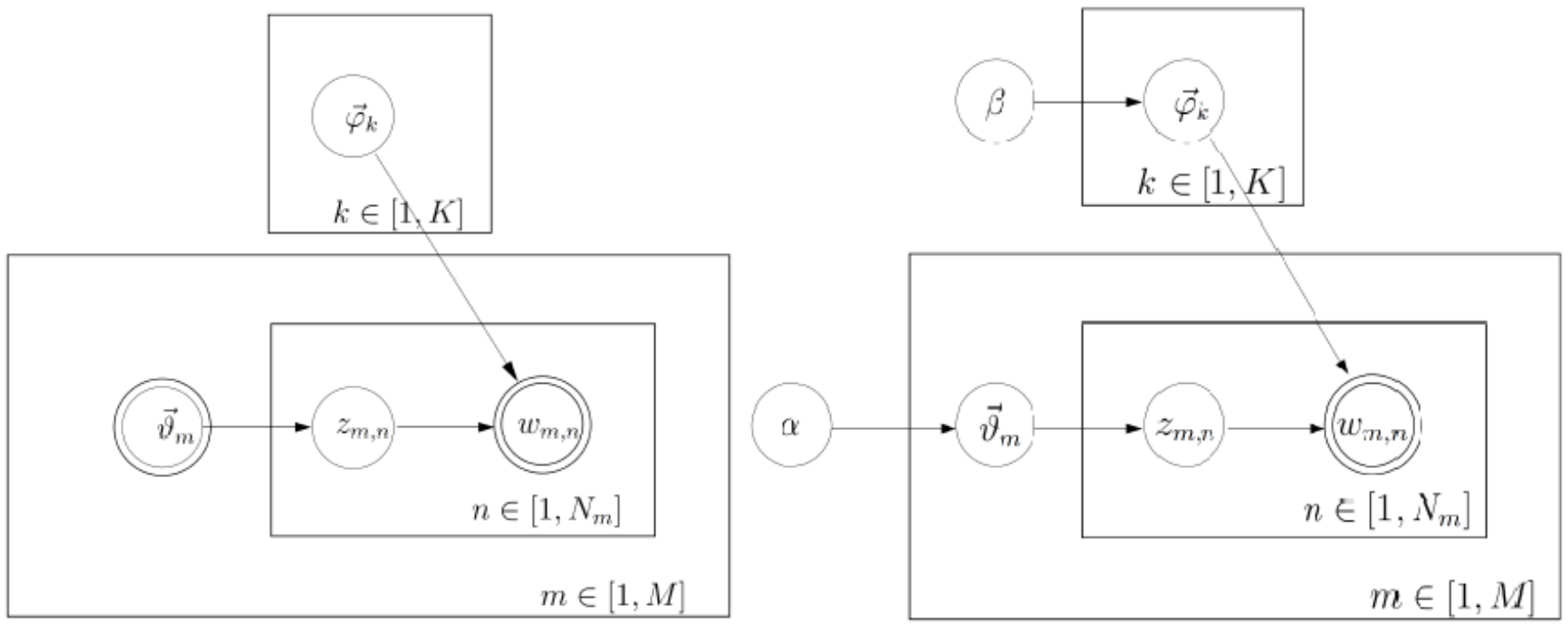

*Figure 11: graphical representation of PLSA (at left) and LDA (at right)*

In PLSA $\theta_m$ is a multinomial random variable and the model learns these topic mixtures only for the training documents M, thus PLSA is not a fully generative model, particularly at the level of documents, since there is no clear solution to assign probability to a previously unseen document. This lead to the consequence of the linear growth of the number of parameters to be estimated with the number of training documents. The parameters for a k-topic PLSA model are k multinomial distributions of size V and M mixtures over the k hidden topics. This gives kV + kM parameters and thus grows linearly in M. The linear growth in parameters implies that the model is prone to overfitting. Even though tempering heuristic is proposed by [2] to smooth the parameters of the model for acceptable predictive performance, it has been shown that overfitting can occur even then [27].
LDA as noted in [3] overcomes both of these problems by treating the topic mixture weights as a k parameter hidden random variable sampled from a Dirichlet distribution, applicable also for unseen documents. Furthermore, the number of parameters is k + kV in a k-topic LDA model, which do not grow with the size of the training corpus, thereby avoids overfitting. Moreover, the count for a topic in a document can be much more informative than the count of individual words belonging to that topic.

Illustrating these differences in a latent space, we can see how a document is geometrically represented under each model. Each document is a distribution over words, so below we observe each distribution as a point on a (V-1) simplex, namely document's word simplex.

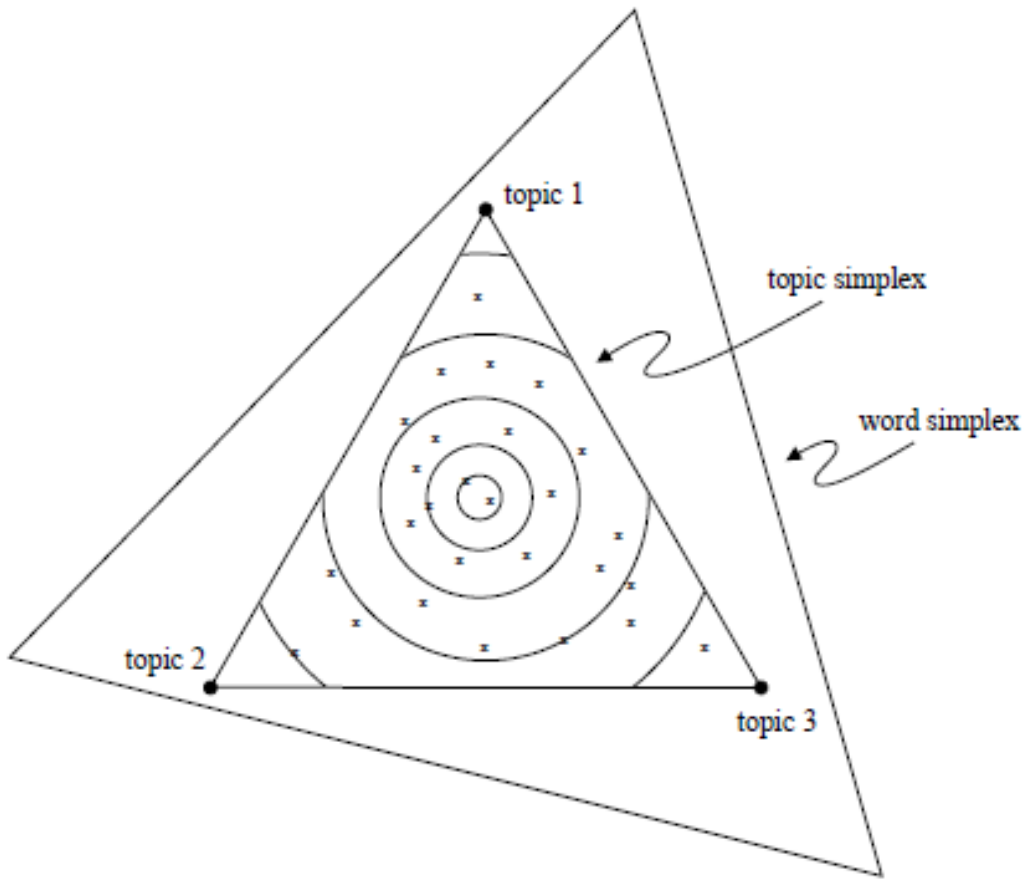

*Figure 12: The topic simplex for three topics embedded in the word simplex for three words [20].*

The mixture of unigrams model in Figure 12 maps each corner of the word simplex to the word probability, equal to one. Each topic corner is then chosen randomly and all the words of the document are drawn from the distribution corresponding to that point.
In PLSI topics are themselves drawn from a document-specific distribution, denoted by spots. LDA though, draws the topics from a distribution with a randomly chosen parameter, different for every document, denoted by the contour lines in Figure 12.

## 2.2.3 Model Inference

In LDA, documents are represented as a mixture of topics and each topic has some particular probability of generating a set of words. Thus, LDA assumes data to arise from a generative process that includes hidden variables. This generative process defines a joint probability distribution over both the observed and the hidden random variables, transferring the inferential problem to *compute the posterior distribution* of the hidden variables given the observed variables, which are the documents.
The posterior distribution is:

$$p(\theta, z|w, \alpha, \beta) = \frac{p(\theta, z, w|\alpha, \beta)}{p(w|\alpha, \beta)} \quad (2.2.3.1)$$

The numerator is the joint distribution of all the random variables, which can be easily computed for any setting of the hidden variables. The denominator is the marginal probability of the observations, which is the probability of seeing the observed corpus under any topic model. Posterior distribution cannot be directly computed. Thus, to solve the problem we need approximate inference algorithms as to form an alternative distribution which is close to the posterior. Such algorithms are divided in the sampling based algorithms and the variational algorithms. *Sampling based algorithms* collect samples from the posterior to approximate it with an empirical distribution, while *variational methods* place a parameterized class of distributions over the hidden structure and then find the member of that class that is closest to the posterior, using Kullback-Leibler divergence [28]. Thus, the inference problem is turned into an optimization problem.

The most popular approximate posterior inference algorithms are the mean field variational method by Blei [3], the expectation propagation by Minka and Lafferty [29] and Gibbs Sampling [30], with many others or extensions of the above.

### 2.2.3.1 Gibbs Sampler

Gibbs sampling is commonly used as a method of statistical inference, especially Bayesian inference. Gibbs sampling is named after the physicist Josiah Willard Gibbs, in reference to an analogy between the sampling algorithm and statistical physics. The algorithm was described by the Geman brothers in 1984 [31], some eight decades after Gibbs's death. Part of the Gibbs Sampler work presented is based on the thesis of István Bíró [32].

Gibbs sampling is applicable when the joint distribution is not known surely or is difficult to sample from the beginning, but the conditional distribution of each variable is known and is easy to sample from. The Gibbs sampling algorithm generates an instance from the distribution of each variable in turn, conditional on the current values of the other variables. It can be shown that the sequence of samples constitutes a Markov chain, and the stationary distribution of that Markov chain is just the recherche joint distribution [33].

Subsequently, Gibbs Sampler or Gibbs Sampling is an algorithm based on Markov Chain Monte Carlo (MCMC) simulation. As noted in [33], MCMC algorithms can simulate high-dimensional probability distributions by the stationary behavior of a Markov chain. The process generates one sample per transmition in the chain. The chain starts from an initial random state, then after a burn-in period it stabilizes by eliminating the influence of initialization parameters. The MCMC simulation of the probability distribution p($\vec{x}$) is as follows: dimensions $x_i$ are sampled alternately one at a time, conditioned on the values of all other dimensions denoted by $\vec{x}_{-i} = (x_1, \dots, x_{i-1}, x_{i+1}, \dots x_n)$, which is:

1. Choose dimension i (random or by cyclical permutation)
2. Sample $x_i$ from p( $x_i | \vec{x}_{-i}$)

As stated above the conditional distribution of each variable is known and can be calculated as follows:

$$p(x_i|\vec{x}_{\neg i}) = \frac{p(\vec{x})}{\int p(\vec{x})\,\mathrm{d}x_i} \quad \text{with} \quad \vec{x} = \{x_i, \vec{x}_{\neg i}\} \qquad (2.2.3.1.1)$$

In order to construct a Gibbs sampler for LDA, one has to estimate the probability distribution $p(\vec{z}|\vec{w})$ for $\vec{z} \in K^N, \vec{w} \in V^N$, where N denotes the set of word-positions in the corpus. This distribution is directly proportional to the joint distribution:

$$p(\vec{z}|\vec{w}) = \frac{p(\vec{z},\vec{w})}{\sum_z p(\vec{z},\vec{w})}. \qquad (2.2.3.1.2)$$

This joint distribution cannot be inferred because of the denominator, which is a summation over $K^N$ terms. This is why we use Gibbs Sampling, as it requires only the full conditionals $p(z_i|\vec{z}_{-i},\vec{w})$ as to infer $p(\vec{z}|\vec{w})$. So, first we have to estimate the joint distribution:

$$p(\vec{w},\vec{z}|\vec{\alpha},\vec{\beta}) = p(\vec{w}|\vec{z},\vec{\beta})p(\vec{z}|\vec{\alpha}), \qquad (2.2.3.1.3)$$

Because the first term is independent of $\vec{\alpha}$ due to the conditional independence of $\vec{w}$ and $\vec{\alpha}$ given $\vec{z}$, while the second term is independent of $\vec{\beta}$. The elements of the joint distribution can now be managed independently. The first term can be obtained from $p(\vec{w}|\vec{z},\Phi)$, which is simply:

$$p(\vec{w}|\vec{z},\underline{\Phi}) = \prod_{i=1}^{N} \varphi_{z_i,w_i} = \prod_{z=1}^{K}\prod_{t=1}^{V} \varphi_{z,t}^{N_{zt}}. \qquad (2.2.3.1.4)$$

The N words of the corpus are observed according to independent multinomial trials with parameters conditioned on the topic indices $z_i$. In the second equation we split the product over words into one product over topics and one over the vocabulary, separating the contributions of the topics. The term $N_{zt}$ denotes the number of times that term t has been observed with topic z. The distribution $p(\vec{w}|\vec{z},\vec{\beta})$ is obtained by integrating over Φ, which can be done using Dirichlet integrals within the product over z, as can be seen below:

$$p(\vec{w}|\vec{z},\vec{\beta}) = \int p(\vec{w}|\vec{z},\underline{\Phi})p(\underline{\Phi}|\vec{\beta})\,\mathrm{d}\underline{\Phi}$$

$$= \int \prod_{z=1}^{K} \frac{1}{\Delta(\vec{\beta})} \prod_{t=1}^{V} \varphi_{z,t}^{N_{zt}+\beta_t-1}\,\mathrm{d}\vec{\varphi}_z$$

$$= \prod_{z=1}^{K} \frac{\Delta(\vec{N}_z+\vec{\beta})}{\Delta(\vec{\beta})}, \quad \vec{N}_z = \{N_{zt}\}_{t=1}^{V} \qquad (2.2.3.1.5)$$

The topic distribution $p(\vec{z}|,\vec{\alpha})$ can be derived from $p(\vec{z}|,\Theta)$:

$$p(\vec{z}|\underline{\Theta}) = \prod_{i=1}^{N} \vartheta_{d_i,z_i} = \prod_{m=1}^{M}\prod_{z=1}^{K} \vartheta_{m,z}^{N_{mz}} \qquad (2.2.3.1.6)$$

where $d_i$ refers to the document word $w_i$ belongs to and $N_{mz}$ refers to the number of times that topic z has been observed in document m. Integrating out Θ, we obtain:

$$p(\vec{z}|\vec{\alpha}) = \int p(\vec{z}|\underline{\Theta})p(\underline{\Theta}|\vec{\alpha})\,\mathrm{d}\underline{\Theta}$$

$$= \int \prod_{m=1}^{M} \frac{1}{\Delta(\vec{\alpha})} \prod_{z=1}^{K} \vartheta_{m,z}^{N_{mz}+\alpha_z-1}\,\mathrm{d}\vec{\vartheta}_m$$

$$= \prod_{m=1}^{M} \frac{\Delta(\vec{N}_m + \vec{\alpha})}{\Delta(\vec{\alpha})}, \quad \vec{N}_m = \{N_{mz}\}_{z=1}^{K} \tag{2.2.3.1.7}$$

Consequently, the joint distribution becomes:

$$p(\vec{w}, \vec{z}|\vec{\alpha}, \vec{\beta}) = \prod_{z=1}^{K} \frac{\Delta(\vec{N}_z + \vec{\beta})}{\Delta(\vec{\beta})} \cdot \prod_{m=1}^{M} \frac{\Delta(\vec{N}_m + \vec{\alpha})}{\Delta(\vec{\alpha})} \tag{2.2.3.1.8}$$

Now, by using (2.2.3.1.8) equation we can determine the update equation for the hidden variable:

$$p(z_i = k|\vec{z}_{\neg i}, \vec{w}) = \frac{p(\vec{z}, \vec{w})}{p(\vec{z}_{\neg i}, \vec{w})} = \frac{p(\vec{w}|\vec{z})}{p(\vec{w}|\vec{z}_{\neg i})} \cdot \frac{p(\vec{z})}{p(\vec{z}_{\neg i})}$$

$$\propto \frac{\Delta(\vec{n}_k + \vec{\beta})}{\Delta(\vec{n}_{z,\neg i} + \vec{\beta})} \cdot \frac{\Delta(\vec{n}_m + \vec{\alpha})}{\Delta(\vec{n}_{m,\neg i} + \vec{\alpha})}$$

$$= \frac{\frac{\Gamma(N_{kt}+\beta_t)}{\Gamma(\prod_{v=1}^{V}(N_{kv}+\beta_v))} \cdot \frac{\Gamma(N_{mk}+\alpha_k)}{\Gamma(\prod_{z=1}^{K}(N_{mz}+\alpha_z))}}{\frac{\Gamma(N_{kt}-1+\beta_t)}{\Gamma(\prod_{v=1}^{V}(N_{kv}+\beta_v)-1)} \cdot \frac{\Gamma(N_{mk}-1+\alpha_k)}{\Gamma(\prod_{z=1}^{K}(N_{mz}+\alpha_z)-1)}}$$

$$= \frac{N_{kt}^{\neg i} + \beta_t}{\sum_{v=1}^{V}(N_{kv} + \beta_v) - 1} \cdot \frac{N_{mk}^{\neg i} + \alpha_k}{\sum_{z=1}^{K}(N_{mz} + \alpha_z) - 1} \tag{2.2.3.1.9}$$

where the superscript $N^{-i}$ denotes that the word or topic with index $i$ is excluded from the corpus when computing the corresponding count. Note that only the terms of the products over m and k contain the index $i$, while all the others are cancelled out.

Having this knowledge we can calculate the values of Θ, Φ from the given state of the Markov chain as are the current samples of p($\vec{z}|\vec{w}$). This can be done as a posterior estimation, by predicting the distribution of a new topic-word pair ($\vec{z}$= k, $\vec{w}$ = t) that is observed in a document m, given state ($\vec{z}|\vec{w}$):

$$p(\tilde{z}=k,\tilde{w}=t|\vec{z},\vec{w}) = p(\tilde{w}=t|\tilde{z}=k,\vec{z},\vec{w}) \cdot p(\tilde{z}=k|\vec{z},\vec{w})$$

$$= \frac{p(\tilde{z}=k,\tilde{w}=t,\vec{z},\vec{w})}{p(\vec{z}|\vec{w}))} = \frac{\frac{\Gamma(N_{kt}+1+\beta_t)}{\Gamma(\prod_{v=1}^{V}(N_{kv}+\beta_v)+1)} \cdot \frac{\Gamma(N_{mk}+1+\alpha_k)}{\Gamma(\prod_{z=1}^{K}(N_{mz}+\alpha_z)+1)}}{\frac{\Gamma(N_{kt}+\beta_t)}{\Gamma(\prod_{v=1}^{V}(N_{kv}+\beta_v))} \cdot \frac{\Gamma(N_{mk}+\alpha_k)}{\Gamma(\prod_{z=1}^{K}(N_{mz}+\alpha_z))}}$$

$$= \frac{N_{kt}+\beta_t}{\sum_{v=1}^{V}(N_{kv}+\beta_v)} \cdot \frac{N_{mk}+\alpha_k}{\sum_{z=1}^{K}(N_{mz}+\alpha_z)}. \quad (2.2.3.1.10)$$

Using the decomposition in (2.2.3.1.10), we can interpret its first factor in the first line as $\varphi_{k,t}$ and its second factor as $\theta_{m,k}$ , hence:

$$\varphi_{k,t} = \frac{N_{kt}+\beta_t}{\sum_{v=1}^{V}(N_{kv}+\beta_v)} \quad (2.2.3.1.11)$$

$$\vartheta_{m,k} = \frac{N_{mk}+\alpha_k}{\sum_{z=1}^{K}(N_{mz}+\alpha_z)} \quad (2.2.3.1.12)$$

Gibbs Sampler algorithm, using (2.2.3.1.9), (2.2.3.1.11), and (2.2.3.1.12) equations can approximate and infer the initial wanted posterior distribution. Moreover, we should state some keys about Gibbs sampling algorithm. During the initial stage of the sampling process namely the burn-in period [30], the Gibbs samples have to be discarded because they are poor estimates of the posterior. After the burn-in period, the successive Gibbs samples start to approximate the target posterior of the topic assignments. To get a representative set of samples from this distribution, a number of Gibbs samples are saved at regularly spaced intervals, to prevent correlations between samples.

# Chapter 3

# 3. Autoencoders

Autoencoder, Autoassociator or Diabolo network is a special sort of artificial neural networks. Neural Networks as stated by W. McCulloch and W. Pitts [34] are inspired by biological neural networks and they are used to estimate or approximate functions given large data as input, generally unknown. Neural networks are adaptive to these inputs and subsequently capable of learning [20]. In this chapter after a review to the neural networks, we present Autoencoder, represented as a feedforward neural net most of the times, as a method for dimensionality reduction and feature extraction. More precisely, the encoder network is used during both training and deployment, while the decoder network is only used during training. The purpose of the encoder network is to discover a compressed representation of the given input. In the next chapters we are going to apply both LDA (described in Chapter 3) and Autoencoder on a movie database as to compare the two methods in their results to dimensionality reduction and feature extraction.

The autoencoder is an unsupervised learning algorithm with the aim to reconstruct the input in the output through hidden layer(s) of lower dimension. This compressed representation of the data leads to a representation in a reduced space and renders the model capable of discovering latent concepts through our data, which is exactly what we want to capture. Moreover, this exact characteristic, accordingly to LDA, gives interest to observe its results respectively to these of collaborative filtering models (see Section 6).

## 3.1 Neural Networks

Neural networks use learning algorithms that are inspired by the procedure that our brain learns, but they are evaluated by how well they work for practical applications such as speech recognition, object recognition, image retrieval and the ability of recommendation. Neural networks, first described by W. McCulloch and W. Pitts [34] as a possible approach for AI applications, is a system consisting of neurons and adaptive weights which represent the conceptual connections between the neurons, that are refined through a learning algorithm, capable of approximating the non-linear functions of their inputs. These weight-values comprise the flexible part of the neural network and define its behavior. With appropriately network functions, various learning assignments can be performed by minimizing a cost function over the network function.

Below we can see the simplest possible neural network, the 'single neuron' [35] [36]:

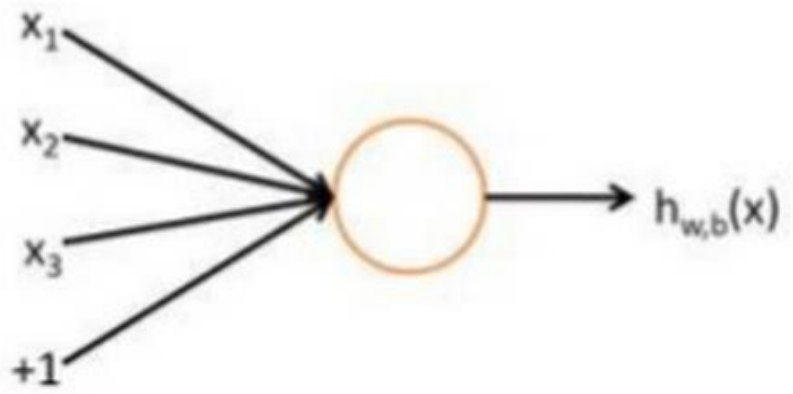

*Figure 13: 'Single neuron'*

This "neuron" is a computational unit that takes as input $x_1, x_2, x_3$ (and a +1 intercept term), and outputs $h_{W,b}(X) = f(W^T x) = f(\sum_{i=1}^{3} W_i x_i + b)$ , where $f: \Re \rightarrow \Re$ is called the activation function.

Neural Networks use basic functions that are nonlinear functions of a linear combination of the inputs, as stated by Bishop [20]. Subsequently, as to from a basic neural network model, first we have to construct M linear combinations of the inputs $x_1, \ldots, x_D$:

$$a_j = \sum_{i=1}^{D} w_{ji}^{(1)} x_i + w_{j0}^{(1)} \tag{3.1.1}$$

where $j = 1, \ldots M$ and (1) superscript denoted the first layer of the network. The $w_{ij}$ parameter is the weight of the input, while the $w_{j0}$ represents a bias. Finally, $\alpha_j$ represents the activation. Each activation is then transformed via a nonlinear activation function h (·), usually chosen to be a sigmoidal function:

$$z_j = h(a_j) \tag{3.1.2}$$

The next layer of the neural net is the linear combination of all $z_j$, $j = 1, \ldots, M$

$$a_k = \sum_{j=1}^{M} w_{kj}^{(2)} z_j + w_{k0}^{(2)} \tag{3.1.3}$$

The output is the transformation of each $\alpha_k$ through a proper activation function, chosen accordingly to the nature of our data. Let σ be the final activation function k output is presented as:

$$y_k = \sigma(a_k) \tag{3.1.4}$$

with the overall network represented below:

$$y_k(\mathbf{x}, \mathbf{w}) = \sigma \left( \sum_{j=1}^{M} w_{kj}^{(2)} h \left( \sum_{i=1}^{D} w_{ji}^{(1)} x_i + w_{j0}^{(1)} \right) + w_{k0}^{(2)} \right) \tag{3.1.5}$$

As an **activation function** we usually use the sigmoid or the hyperbolic tangent (tanh) functions.

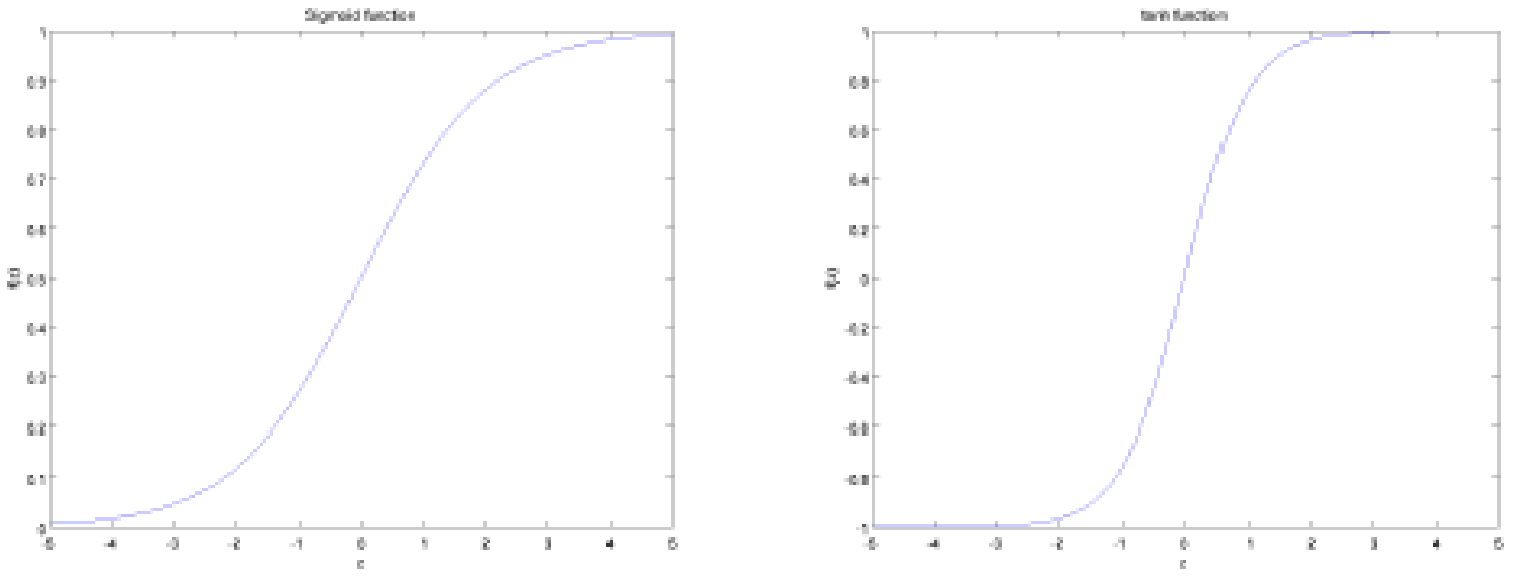

$$\sigma(a) = \frac{1}{1+e^{-\alpha}} \qquad (3.1.6)$$

$$\tanh(\alpha) = \frac{e^{\alpha}-e^{-\alpha}}{e^{\alpha}+e^{-\alpha}} \qquad (3.1.7)$$

*Figure 14: Sigmoid (left) and tanh (right) activation functions [20]*

The tanh(z) function is a rescaled version of the sigmoid, and its output range is [−1, 1] instead of [0, 1]. The functions take as input the weighted sum α, of the values coming from the units connected to it.
Note that the output values for the σ function range between but never make it to 0 and 1. This is because $e^{-\alpha}$ is never negative, and the denominator of the fraction tends to 0 as α gets very big in the negative direction, and tends to 1 as it gets very big in the positive direction. This tendency comes easily as the middle ground between 0 and 1 is rarely seen because of the sharp (near) step in the function. Because of it looking like a step function, we can think of it firing and not-firing as in a perceptron: if a positive real is input, the output will generally be close to +1 and if a negative real is input the output will generally be close to -1.

In general, a network function linked to a neural network declares the relationship between input and output layers, parameterized by the weights. By finding out proper network functions, various learning tasks can be performed by minimizing a cost function over the network function that is the weights.

A neural network is put together by linking many of our simple "neurons," so that the output of a neuron can be the input of another. For example, here is a small neural network

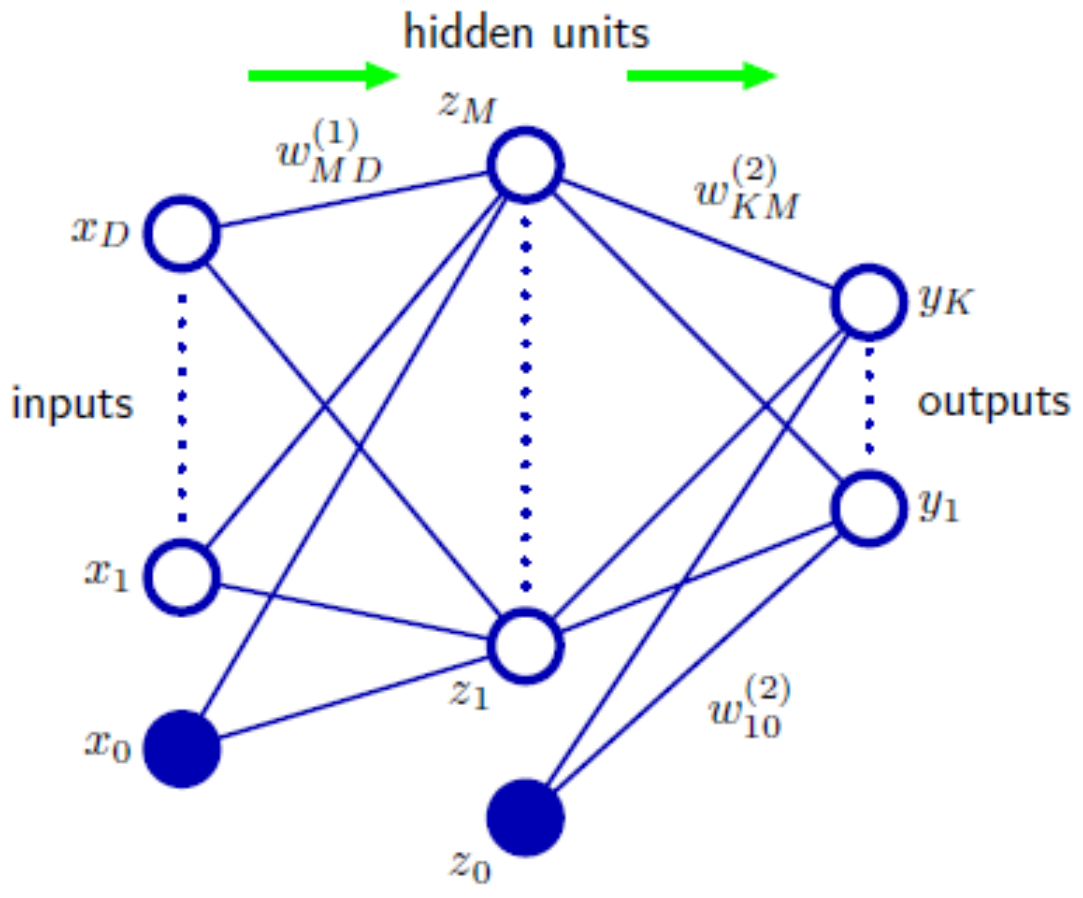

*Figure 15: 3-layer neural network [20]*

*Figure 14* illustrates a three layer neural net, where x denotes the input layer, y the output layer and the middle layer of the nodes, z, is the hidden layer, we are mainly interested in this thesis, as it represents the representation of our data in a lower dimension space. The shaded nodes represent the biases and are not considered as inputs. Moreover, this is a feedforward neural network. Thus, the information 'travels' only one direction.

When speaking for a **feedforward neural network**, namely a neural net whose information move in one direction and do not shape any directed loops or cycles, by repeating (3.1.1) and (3.1.2) equations we can form a neural network with multiple hidden layers. Of course, neural nets can have more than one outputs as seen in *Figure 15*.
*Multilayer feedforward neural networks* can be used to perform feature learning and prediction, since they learn a representation of their input at the hidden layer(s), which is subsequently used for classification or regression at the output layer.

## 3.2 Autoencoder

*Autoencoders or Autoassociators* are a special kind of artificial neural networks and provide a fundamental paradigm for unsupervised learning. They were first introduced in 1986 by Hinton et al. [37] to address the problem of *backpropagation* (See Section 3.3) without a teacher", by using the input data to avoid the need for having a teacher and are later related to the concept of Sparse Coding, presented by Olshausen et al. [38] in 1996. Their input and output layer have the same size and there is a smaller hidden layer in between. Autoencoder tries to reconstruct the input vector in the output layer with as much accuracy by learning to map the input to itself (auto-encoder) in a lower dimension space. Thus, the network is evaluated by evaluating the input through the hidden layer to the output layers. Because the goal is to reconstruct the input-vector in the output layer as precisely as possible, the network is back-propagated with the error between the reconstruction and the original pattern. The smaller sized hidden layer has to represent the larger input data. Therefore, the system learns a compressed representation of the data. The activation of the hidden layer provides a compressed representation of the data, the *encoding* of the data.
More recently, autoencoders have come to the forefront in the deep architecture of neural networks, as if put one on top of each other [39], they create stacked autoencoders capable to learn deep networks [40] [41]. These deep architectures are shown to present great results on a number of challenging classification and regression problems.
Autoencoder, in its simplest representation is a feedforward, non-recurrent neural network, very similar to a multilayer perceptron. The difference is that in Autoencoder the output layer is of the same size as the input layer is, so instead of training the neural net to predict some target value 'y', the autoencoder is trained to reconstruct its own inputs (auto-encoder).
The framework of the Autoencoder as represented by Baldi [42]:

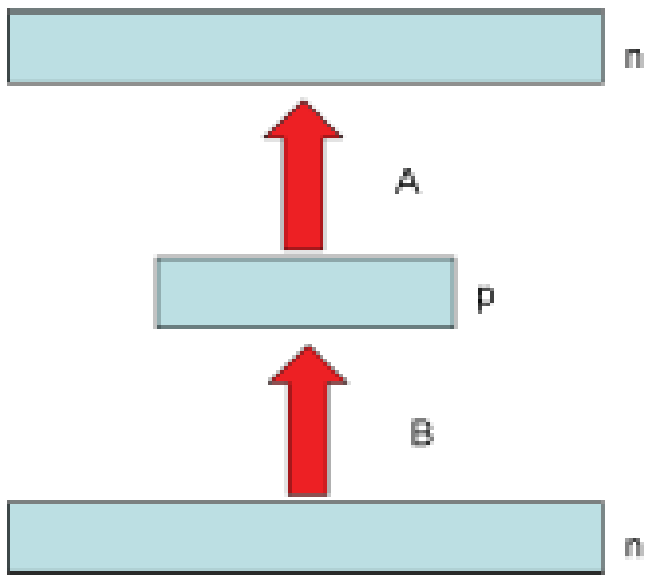

*Figure 16: Autoencoder's architecture*

*Figure 16* illustrates the general architecture behind an Autoencoder neural net that is defined by n, p, A, B, X, Δ, F, G. F represents the input/output layer denoted as a $F^n$ vector and G declares the hidden layer, the bottleneck, denoted as $G^p$ vector, with n, p be positive integers. A is a transformation class from $G^p$ to $F^n$, B is a similar class from $F^n$ to $G^p$. $X = \{x_1, \dots, x_m\}$ is a set of m training vectors in $F^n$ and Δ is dissimilarity or distortion function defined over $F^n$.

Given these tuples, the corresponding problem is to find $A \in A$ *and* $B \in B$ that minimize the overall error (distortion) function:

$$\min E(A, B) = \min_{A,B} \sum_{t=1}^{m} E(x_t) = \min_{A,B} \sum_{t=1}^{m} \Delta\big(A \circ B(x_t), x_t\big) \qquad (3.2.1)$$

The difference with a multilayer perceptron where the external target $y_t$ are considered as different to the input x, can be better understand below, with the minimization problem become as follows:

$$\min E(A,B) = \min_{A,B} \sum_{t=1}^{m} E(x_t, y_t) = \min_{A,B} \sum_{t=1}^{m} \Delta(A \circ B(x_t), y_t) \quad (3.2.2)$$

During the training phase of every neural network, the weights (and hence F) are successively modified, via one of several possible algorithms, in order to minimize E.

Generally there are two options for the Autoencoder network. Mainly when we refer to an Autoencoder neural network we assume that the layer of the hidden nodes, often referred as bottleneck, (Thompson et al., 2002), are of smaller dimension than the input and output layer. So, in *Figure 16*, we can see precisely this with p<n. This structure results in the compression of data into a smaller dimension and then decompressing into the output space, resulting in a butterfly structure:

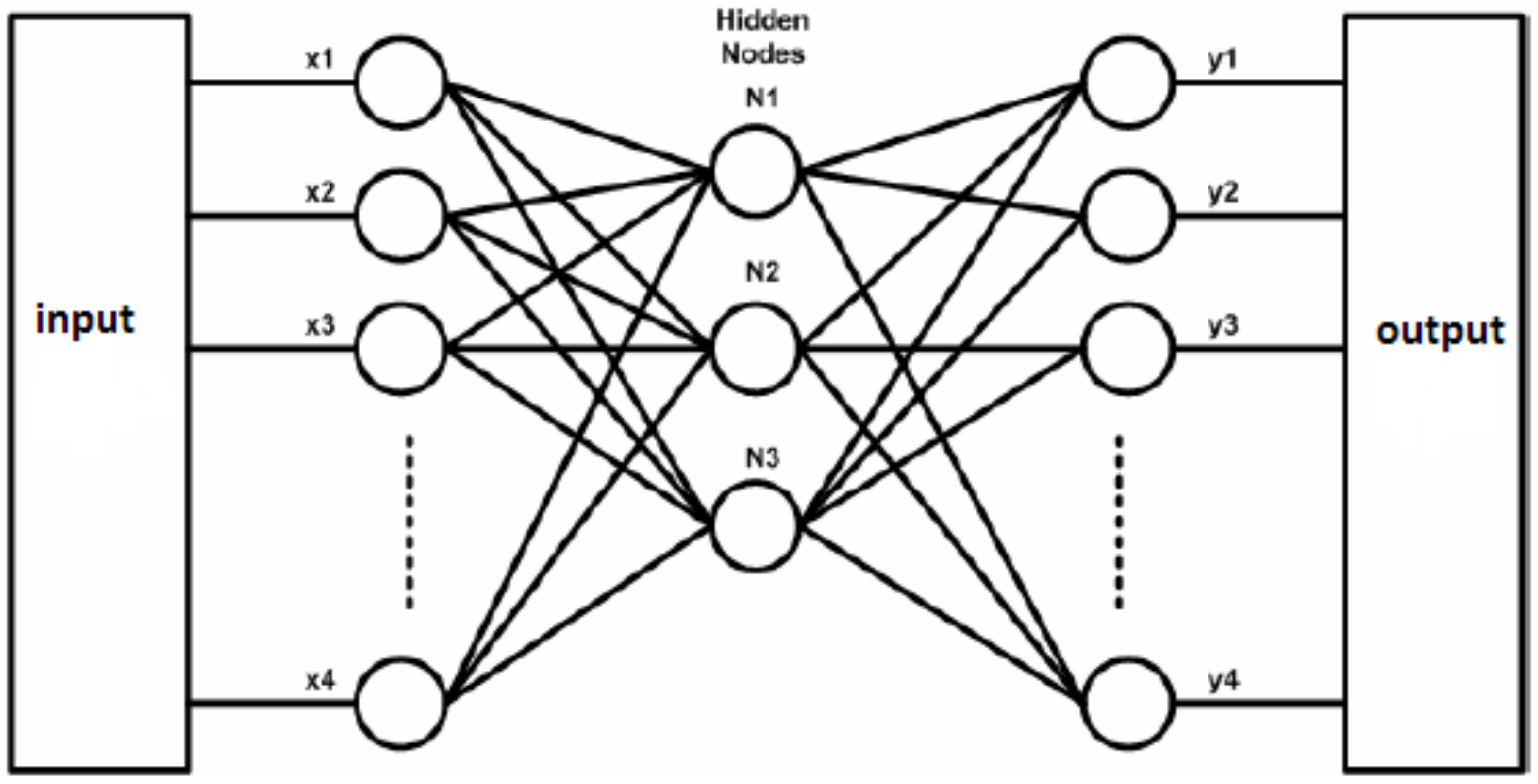

*Figure 17: Autoencoder butterfly scheme [43]*

In *Figure 17*, we notice that the hidden layers are narrower that the input-output layers and subsequently the activations of the final hidden layer can be regarded as a compressed representation of the input. Moreover, as we limit the number of the hidden units, we can discover interesting structure about the data.

More general speaking Autoencoders can be presented also as deep networks with a symmetric topology and an odd number of hidden layers, containing an encoder, a low dimensional representation and a decoder, with the center small layer operating as the bottleneck.

If the hidden layers are narrower (have fewer nodes) than the input/output layers, then the activations of the final hidden layer can be regarded as a compressed representation of the input. All the usual activation functions from MLPs can be used in autoencoders; Significantly, when the number of hidden units is smaller than either the number of input or output units, then the transformations that the network can generate are not the most general possible linear transformations from inputs to outputs because information is lost in the dimensionality reduction at the hidden units. In this case we use sigmoid functions like the logical or the hyperbolic function. Moreover, if linear activations are used, or only a single sigmoid hidden layer, then the optimal solution to an auto-encoder is strongly related to principal component analysis (PCA) [44].

On the other hand, hidden layers can be of bigger size than the input layer. In this case the autoencoder can potentially learn the identity function and become useless; however, experimental results have shown that such autoencoders might still learn useful features in this case [45].

For the above reasons, autoencoder network is preferred in recall applications as it can map linear and nonlinear relationships between all of the inputs.

Even though Autoencoders are unsupervised networks, their assumption that the output 'y' is known, renders a supervised comportment to the model. This is why Autoencoders use the Backpropagation method as their training method (See Section 3.3), as Backpropagation is a method of implementing a gradient descent method for E that requires a known, desired output for each input value in order to calculate the loss function gradient.
Consequently, autoencoders are simple learning circuits which aim to transform inputs into outputs with the least possible amount of distortion .If we force the output of a multi-layer neural network to be the same as its input and we put a tiny layer in the middle to form a bottleneck, then the value of this bottleneck layer is forced to be an efficient code of the input. Due to its many layers and the non-linearities, the codes found by an autoencoder can be compact and powerful.

## 3.3 Backpropagation method

Backpropagation method, the abbreviation for "backward propagation of errors", is a typical method of **training** artificial neural networks [37] used in combination with an optimization method like **gradient descent**. The method evaluates the gradient of a loss-error function according to all the weights E(w) in the network. The gradient is then driven to the chosen optimization method which in turn uses it to update the weights, in an attempt to minimize the loss function. Backpropagation requires a known, desired output for each input value in order to calculate the loss function gradient, as it is subsequently explained. This is why Backpropagation method is usually used as a supervised method, although it is also used in some unsupervised networks such as autoencoders where the output is considered known.

Backpropagation's goal subsequently, is to calculate the gradient of E(w). This process to minimize the error function can be divided in two distinct steps:

1. *Evaluate the derivatives* of the function E(w), with respect to the weights.
2. Use the derivative from step (1) to *update the weights*.

Let assume we have the following neural net: $(\vec{x}, \vec{y}\,)$ where $\vec{x}$ corresponds to the input layer, $\vec{y}$ to the output layer and $\vec{t}$ represents the wanted output, namely the bottleneck. What we want is to minimize the error function E($w$):

$$E(\mathbf{w}) = \sum_{n=1}^{N} (y_n - t_n)\phi_n \tag{3.3.1}$$

, where n is a particular input pattern, and $\varphi_n$ is the sigmoid logistic function. This procedure can be think as an information that is sent alternatively forwards and backwards through the net for every possibly combination of paths (N).

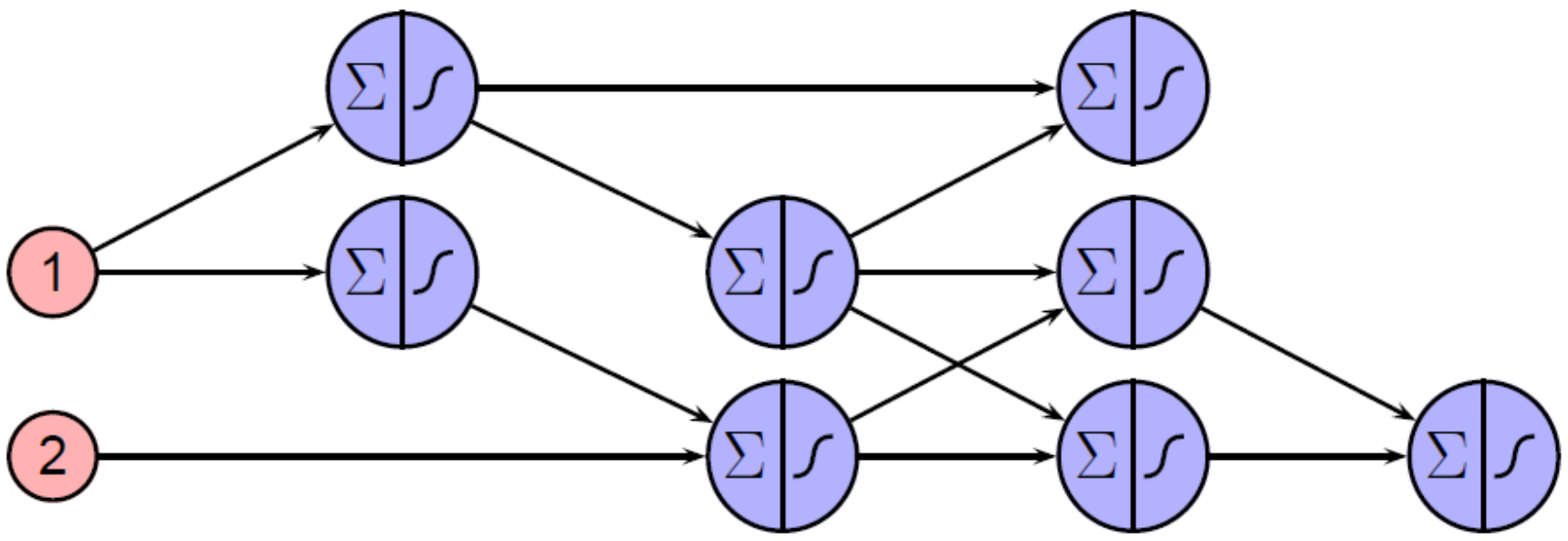

*Figure 18: Forward and backward process of Backpropagation method*

*Figure 18*, illustrates the forward and the backward processes of the Backpropagation method. Precisely, with Σ are noted the activation functions of every node that are generated going forward to the neural net until to meet the final node, representing the error function and with S are noted the derivatives of the nodes, travelling backward from the last node to the initial node as to adjust the weight of this path. This procedure is iterative until all paths are covered [20].

**Step 1**

**Forward propagation:** We use equations (3.1.1) and (3.1.2)

$$a_j = \sum_{i=1}^{D} w_{ji}^{(1)} x_i + w_{j0}^{(1)}$$

Which represents the computed weighted sum of the inputs of a j unit.

$$z_j = h(a_j)$$

with $z_j$ stating the activation of this unit, that can be both input to another unit or the final output. This step represent the flow of the information through the neural net. We can now consider the derivative of $E_n$ with respect to a weight $w_{ij}$. The outputs of the various units will depend on the particular input pattern n. However, in order to keep the notation uncluttered, we shall omit the subscript n from the network variables. First we note that $E_n$ depends on the weight $w_{ij}$ only via the summed input $\alpha_j$ to unit j. We can therefore apply the chain rule for partial derivatives to give:

$$\frac{\partial E_n}{\partial w_{ji}} = \frac{\partial E_n}{\partial a_j}\frac{\partial a_j}{\partial w_{ji}} \qquad (3.3.2)$$

**Backward propagation:**

Where we note the first part of (3.3.2) equation, often referred as error below:

$$\delta_j \equiv \frac{\partial E_n}{\partial a_j} \qquad (3.3.3)$$

and by using (3.1.1) equation we transform the second part of (3.3.2) as:

$$\frac{\partial a_j}{\partial w_{ji}} = z_i \qquad (3.3.3)$$

Consequently (3.3.2) becomes:

$$\frac{\partial E_n}{\partial w_{ji}} = \delta_j z_i$$

(3.3.4)

Equation (3.3.4) tells us that the required derivative is obtained simply by multiplying the value of δ for the unit at the output end of the weight by the value of z for the unit at the input end of the weight. Z is formed as a simple linear model as seen in Section 3.1 of this chapter, so as to evaluate the derivatives we only need to calculate δ. For the output units δ is: $\delta_k = y_k - t_k$ (3.3.5)
For the hidden layers we use the chain rule for partial derivatives, similarly as in (3.3.2), so δ in hidden layers be expresses as:

$$\delta_j \equiv \frac{\partial E_n}{\partial a_j} = \sum_k \frac{\partial E_n}{\partial a_k}\frac{\partial a_k}{\partial a_j} \tag{3.3.6}$$

, where the sum runs over all units k to which unit j sends connections. Units labelled as k could include other hidden units and/or output units. Accordingly to (3.1.1), (3.1.2) and (3.3.6) δ can be written:

$$\delta_j = h'(a_j)\sum_k w_{kj}\delta_k \tag{3.3.7}$$

, which tells us that the value of δ for a particular hidden unit can be obtained by propagating the δ's backwards from units higher up in the network. As we already know the values of δ's for the output units now, if we recursively apply (3.3.7) we can evaluate the δ's for all of the hidden units in a feed-forward network, regardless of its topology.
We can then assume step 1 of backpropagation, often called error propagation can be presented as follows:

1. Apply an input vector $x_n$ to the neural net and forward propagate through the network as to find the activation functions of all the hidden and output units.
2. Evaluate $\delta_k$ for all the output units.
3. Backpropagate the δ for each hidden unit.

**Step 2**

As to update the weights using *gradient descent* as an optimization method we choose a learning rate λ. We want to calculate the gradient of the error towards the network parameters. Moving the parameters in this direction is the quickest way to increase the error; we want to minimize the error so we move in the opposite direction of the gradient. Thus, the change in weight, which is added to the old weight, is equal to the product of the learning rate and the gradient, multiplied by -1:

$$\Delta w_{ij} = -\lambda\frac{\partial E}{\partial w_{ij}} \tag{3.3.8}$$

A difficult task is to choose the right learning rate. If they are too high the neural net will not converge and learning will be unstable, if they are too low convergence takes very long. We scale the learning rates for every layer proportional to $\frac{1}{\sqrt{n^i}}$, where $n^i$ is the number of inputs of layer i as proposed by LeCun et al. [46].
We start by using a learning factor λ factor and calculate the learning rate from that:

$$\lambda_{weight} = \frac{\lambda_{factor}}{\sqrt{n^i}} \tag{3.3.9}$$

, where $\lambda_{weight}$ is the learning rate for the weights and $\lambda_{bias}$ the learning rate for the bias. The learning rates are used to slowly adjust the parameters of the network to a good solution. We use factor ten smaller learning rate for the bias: $\lambda_{bias} = \lambda_{factor}/10\sqrt{n^i}$. We do this because there are more weights than biases which means the training of the bias should settle earlier. Therefore, a lower learning rate is preferable.

# Chapter 4

# 4. Feature Representation

In this chapter we analyze how features can be represented. Representation of texts is very important for the performance of many real word applications. They can be represented either as local representations with n-grams and as bag of words, either as continuous representations with LSA, LDA and Distributed Representations the most representative methods [6]. First, we are going to see how we can represent our corpus, composed from words in vectors (word2vec). Moreover, we are going to analyze some further methods for obtaining word vectors proposed by Mikolov et al. for Google searching machines [6] that arise from the bag of words assumption and present astonishing results, as they detect the presence of phrases in documents, something that was thought as impossible for the bag of words. Furthermore, we see how documents can be represented respectively to the word2vec model (doc2vec), a method the Autoencoder uses to train the documents. Last, we see how the Latent Dirichlet Allocation obtain the paragraph vectors as to represent its features, a different aspect for the doc2vec model of Autoencoder, but for the same goal: represent the documents of the corpus.

## 4.1 Word Vectors (word2vec models)

Word and Paragraph Vectors are distributed representations for words. The notion of distributed representations of words was first proposed in 1986 by Rumelhart et al. [47] and has a fundamental role in statistical language modeling [48]. Word vectors have been used in NLP applications such as word representation, named entity recognition, word sense disambiguation, parsing, tagging and machine translation [49] [50] [51] [52] [53] [54] [55]. The main intuition behind word vectors is the ability to predict a word given other words in a context. Such vectors are trained via autoencoder-style models.
The notion of distributed representations of words and paragraphs is that the information about the word is distributed all along the vector.

Distributed representations of words in a vector space help learning algorithms to achieve better performance in natural language processing tasks by grouping similar words. In this Section we present how word vectors are being formed.

The first step of the process is to create a binary vector V with the corpus vocabulary, where each $V_i$ represents a word of the corpus.

$$V = \begin{bmatrix} el & gato & canta & negro & es & bellisimo & pato \end{bmatrix} \quad (4.1.1.1)$$

Using this vector we will represent each of the given sentences, by turning on and off the corresponding index of each word. For example the sentence "el gato canta", "el gato negro es belissimo" and "el pato canta" are equal to:

$$\begin{aligned} sentense_1 &= \begin{bmatrix} 1 & 1 & 1 & 0 & 0 & 0 & 0 \end{bmatrix} \\ sentense_2 &= \begin{bmatrix} 1 & 1 & 0 & 1 & 1 & 1 & 0 \end{bmatrix} \\ sentense_3 &= \begin{bmatrix} 1 & 0 & 1 & 0 & 0 & 0 & 1 \end{bmatrix} \end{aligned} \quad (4.1.1.2)$$

Several researchers instead of using binary values for each word use the times of occurrence in the given sentence. However, still these inputs by themselves are not very useful, as we need linear features to represent each word, where now each word is represented in binary format (or something very similar). The idea is based on (Stanford NLP Group 2013) and (Lauly, Boulanger, and Larochelle 2014) [56], and is to create a higher level of abstraction that will represent each word. Using the sentences of (4.1.1.2) as our input, we will feed them to an Autoencoder model, a feed forward neural network, with a single hidden layer, for feature extraction. Therefore, the Autoencoder will have V input neurons, where in our case the vocabulary length is 7 and d neurons in its hidden layer. In the training phase the Autoencoder will try to learn weights to reproducing the input. Training the Autoencoder we will create a higher level of abstraction for the dataset, as depicted in *Figure 19*.

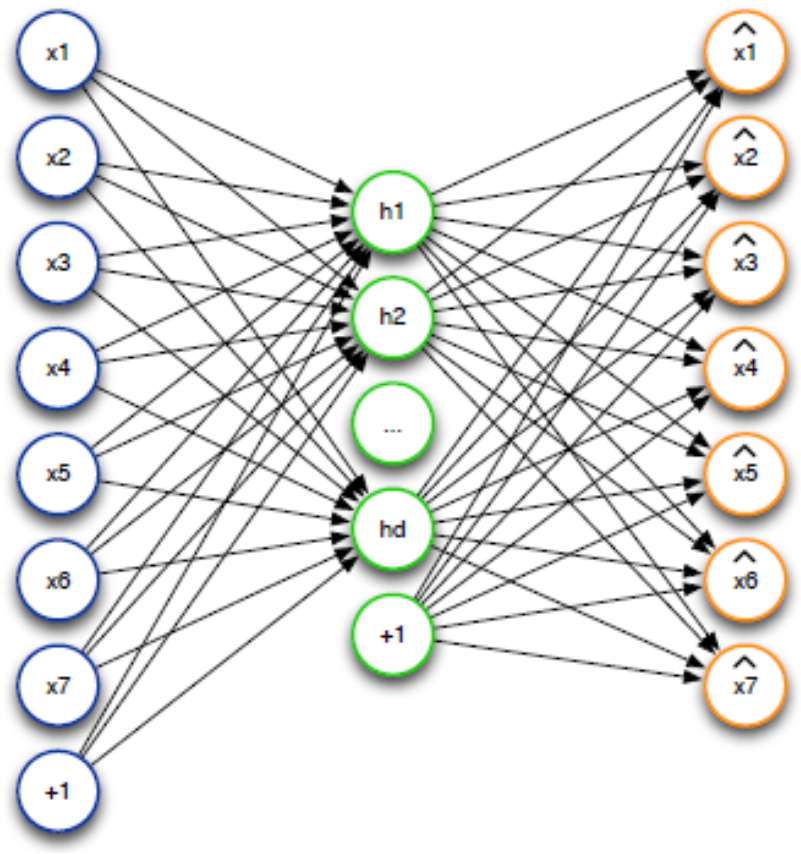

*Figure 19: Autoencoder model for learning word features*

After training the Autoencoder with the given sentences, each of the hidden neurons will learn a weight for each of the inputs (the 7 words), and these weights can be translated to features learnt for each one of the words. Therefore, in the end we will have d features for each of the words as there are d hidden neurons in our network that will have learnt a weight for each one of the inputs. This would form a matrix sized $d \times |V|$ with all the weights/features learnt as depicted in *Figure 19*, where each column represents the weights/features learnt for each of the words of our vocabulary [56].

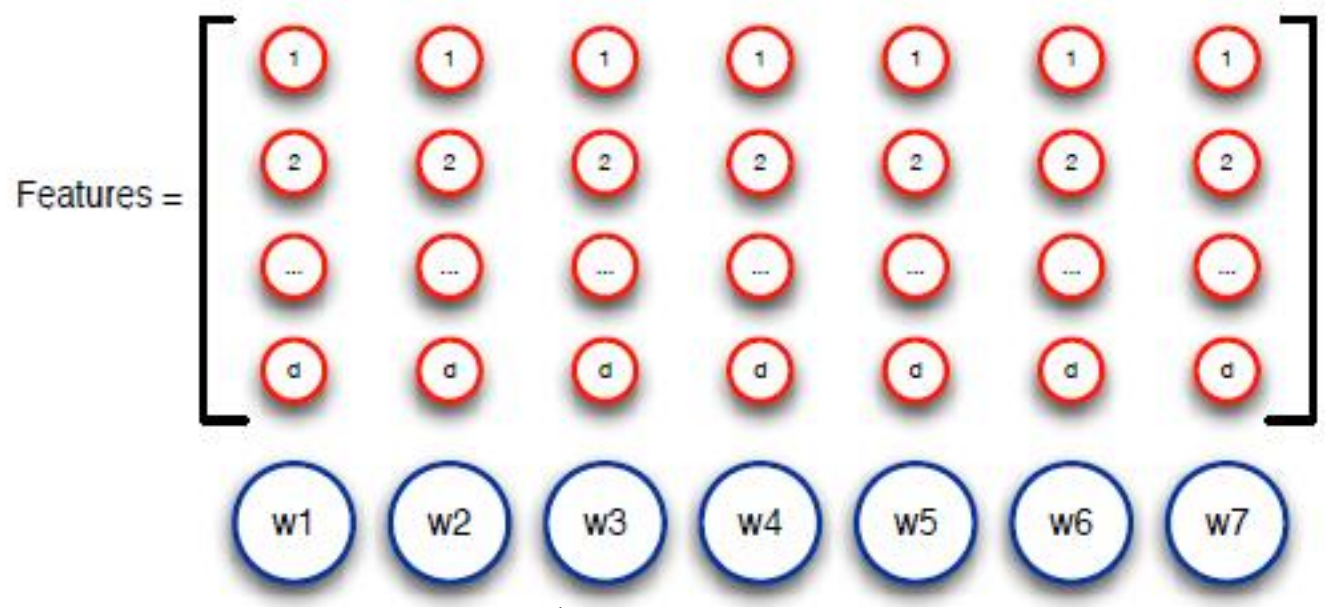

*Figure 20: Word features/weights learnt from the d hidden neurons*

A different approach that is used, and is considered to be more effective is, instead of forming a training set with sentences, to use the continuous **skip-gram (sg)** or the **continuous bag-of-words (cbow)** models to form it, as the Google word2vec tool does [57], that can provide an additional accuracy [6]. For example, these models from the second sentence "el gato negro es bellisimo" would create the following smaller ones for training, resulting a more flexible dataset:

*Continuous bag-of-words with 3-gram*

"el gato negro", "gato negro es", "negro es bellisimo"

*Skip-gram with 1-skip and 3-grams*

"el gato negro", "el gato es", "gato negro es", "gato negro bellisimo", "negro es bellisimo"

Therefore, our training set would be enlarged and the weights/features learnt from each word would represent more cases that are slightly different, producing a more versatile training set. More precisely the skip-gram model is an efficient method for learning high-quality distributed vector representations that capture a large number of precise syntactic and semantic word relationships.

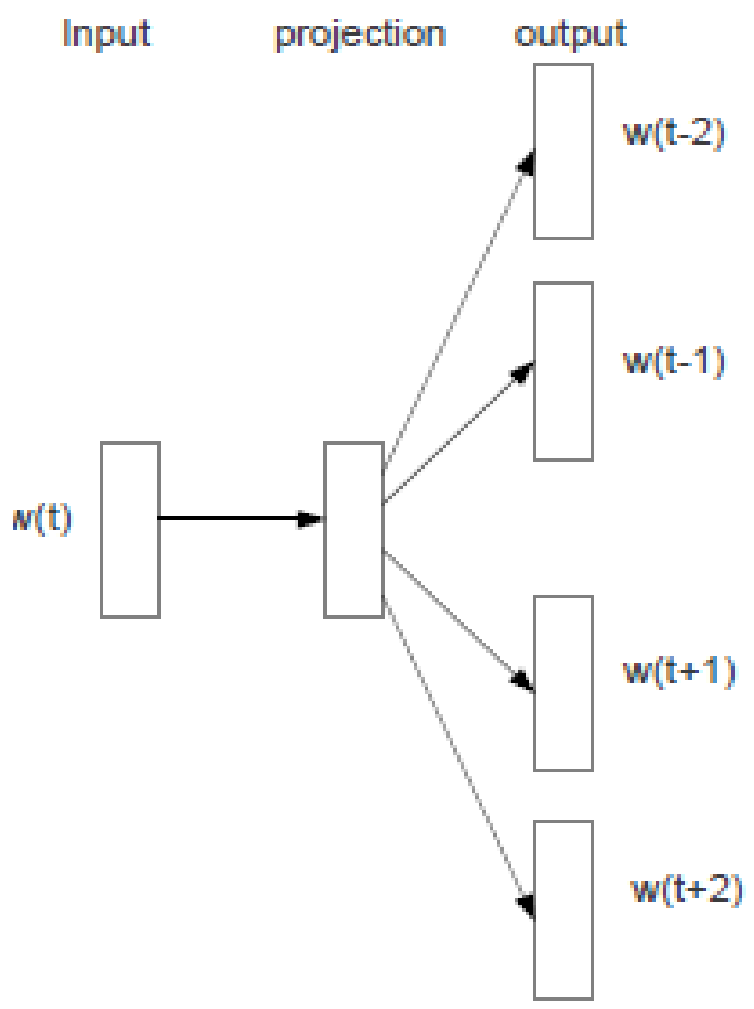

*Figure 21: **Skip-gram** model architecture. The training objective is to learn word vector representations that are good at predicting the nearby words[55].*

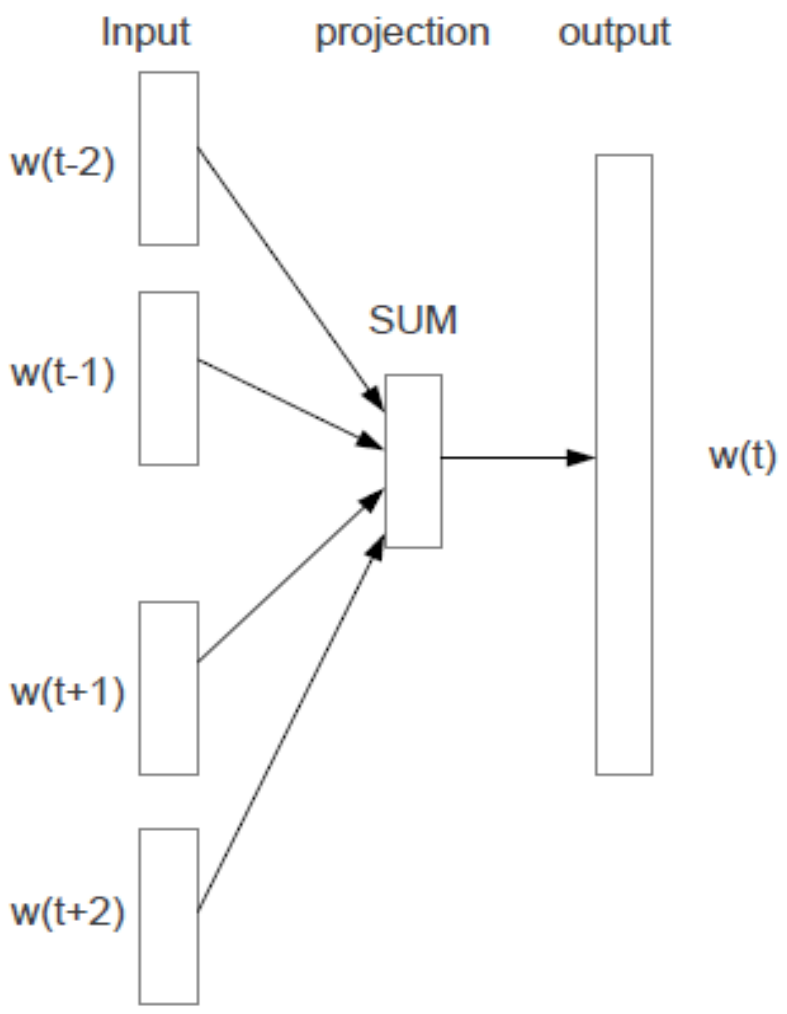

*Figure 22: **bag-of-words** model architecture. Predicts the current word given the context[55].*

In *Figure 21*, we notice the skip gram to be a generalization of n-grams whose words need not be consecutive in the text under consideration, but may leave gaps that are skipped over [58]. They provide one way of overcoming the data sparsity problem found with conventional n-gram analysis.
In *Figure 22*, in the back of words model we can assume that the context of the four words in the input layer is used to predict the fifth word in the output layer, capturing the initial intuition of using word vectors to predict a word [59]. More precisely, given a sequence of training words $w_1, w_2, \dots, w_T$ we find the objective of the word vector model to maximize the average log probability of:

$$\frac{1}{T}\sum_{t=k}^{T-k} \log p(w_t|w_{t-k}, \dots, w_{t+k}) \tag{4.1.1.3}$$

The prediction of the word is done usually with a multiclass classifier, such as softmax (a generalization of the logistic function):

$$p(w_t|w_{t-k}, \dots, w_{t+k}) = \frac{e^{y_{w_t}}}{\sum_i e^{y_i}} \tag{4.1.1.4}$$

Each of $y_i$ is un-normalized log-probability for each output word i, computed as:

$$y = b + Uh(w_{t-k}, \dots, w_{t+k}; W) \tag{4.1.1.5}$$

, with ‘U’, ‘b’ the softmax parameters. ‘h’ is constructed by a concatenation or average of word vectors extracted from W.

This efficient multi-threaded implementation of the new models highly reduces the training complexity. Of course the quality of word representations improves significantly the training data increase. The astonishing with such models, even though they are so simple to understand and applied is that we can importantly increase the degree of language understanding by using basic mathematical operations on the word vector representations.

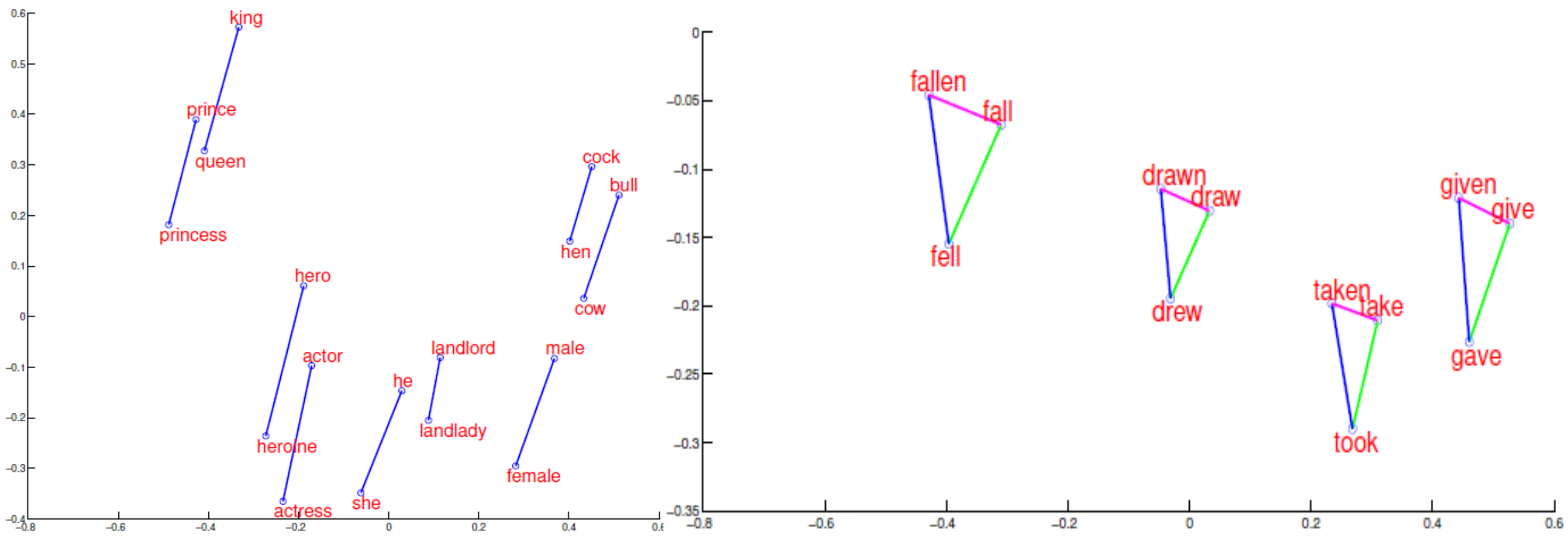

*Figure 23: The word vector space implicitly encodes many regularities*

As we can observe in *Figure 23*, when the distributed representations of words are visualized in a 2d space (here, with PCA tool) they demonstrate surprisingly a lot of syntactic and semantic information. For example as seen above, “KING” is similar to “QUEEN” as “MAN” is to “WOMAN”, while all male-female concepts have almost the same distance. At right, we can also see that different verbs in their present-past-past participle form, are evenly depicted. Subsequently we can take very intuitive results with simple vector operations. We should also state that this is also the way Google Translate works [6], as its computers generate translations based on patterns found in large amounts of text. We can see that below:

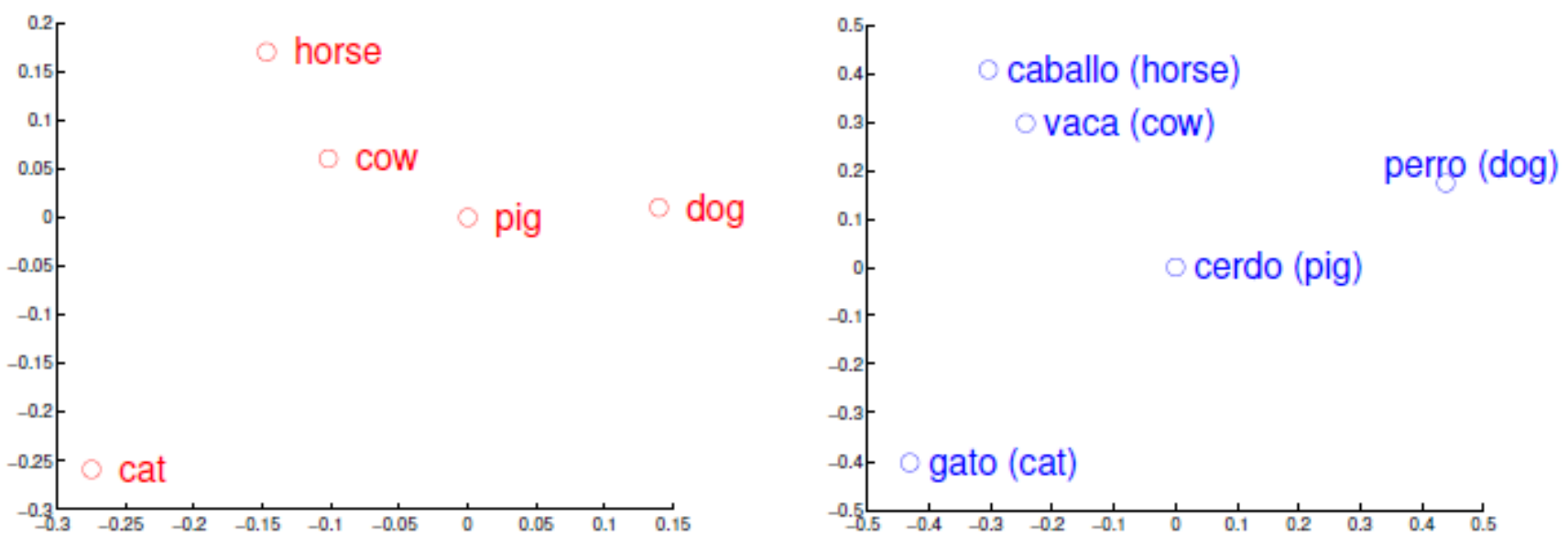

*Figure 24: English to Spanish translation, google translate (http://deeplearning4j.org/word2vec.html)*

In *Figure 24*, the Spanish translation comes if the English vector is rotated and scaled. Generally, for translating one language's vector to that of another language, the statistical machine translation, uses a linear projection, performing rotation and scaling. These linear projections are used via training large corpora with autoencoder-style models (https://code.google.com/p/word2vec/).

Intuitively, what skip grams trying to manage is to push "close" neighbors of the d-dimensional feature space to also have "close" y values. More specifically, by minimizing the objective function we attempt to ensure that if $x_i$ and $x_j$ are "close" then their $y_i\ (x_i,\theta\ )$ and $y_j(x_j,\theta\ )$ are also "close". This problem is also known as semi-supervised Label Propagation and was introduced by (Zhu and Ghahramani 2002) [60]. In contrast to this algorithm, the potential solution is obtained using deep networks in our case. The approach creates a fully connected graph and the nodes are the data points, in our case words, of both labeled and unlabeled sets, where the edges between the nodes i, j are weighted so that nodes having "close" Euclidian distance $\|x_i - x_j\|^2$ have the larger weight. The larger weight is assigned so that the objective function is forced to minimize their $\|y_i\ (x_i, \theta\ )\ - y_j(x_j, \theta\ )\|^2$ distance. Therefore, in order to build the Weight Matrix we need a measure of distance similarity for data points, and the most common one is the Gaussian Kernel similarity measure [61] [62]. Therefore, our Weight Matrix will formed as follows:

$$W_{ij} = \exp\left(-\frac{\|x_i - x_j\|^2}{2\sigma^2}\right) \qquad (4.1.1.3)$$

As it can be seen, the weights are dependent to $\sigma^2$, where the selection of $\sigma^2$ will be discussed later, and for the moment we will assume that σ = 1 / $\sqrt{2}$ and as a result the weight matrix will become $W_{ij} = \exp\left(-\|x_i - x_j\|^2\right)$. Now as $\|x_i - x_j\|^2\ \geq 0$, this means that the weight for "close" neighbors will be close to 1 and as the Euclidean distance grows the weight will decrease exponentially tending to infinity. This means that data points with long d-dimensional Euclidean distance will not be pushed that much to also have similar y distance.

About the choice of σ Bengio, Delalleau, and Le Roux [61] refer to it as the *bias-variance dilemma*, and there are many approaches that could be used. The simplest and most naïve one is to assume that is equal to σ

= 1 / $\sqrt{2}$ or $\sigma$ = 1 [63]. Another approach that is much more efficient was suggested by Zhu, Ghahramani, and Lafferty in 2003 [62], is to construct the minimum spanning tree of all the data-points, in our case words, find an edge that connects two different labels and set $\sigma = \|x_i - x_j\|^2 / 3$ . Moreover, they also state that a more complex approach would be to have a different σ for each of the dimensions. Last but not least, the selection could also be done using Bayesian Optimization and, as we are using partial supervision and we can try to evaluate some results that are known.

## 4.2 Using Autoencoder to obtain paragraph vectors (doc2vec)

As described in Section 3.2 the Autoencoder is an unsupervised neural network that assumes the output Y as known. Considering that the goal is to reconstruct the input-vector in the output layer as well as possible the network is back-propagated with the error between the reconstruction and the original pattern with the hidden layer representing a reduced dimension of the data. The question is now transferred to what the input form should be. Of course, our corpus consist the input and consequently has the assumption of a bag of words, but which is the more efficient way to represent our data as an input to the autoencoder neural network?

Autoencoder uses a paragraph to vector model, known as doc2vec, which was proposed by T. Mikolov last year, 2014 [59]. Paragraph Vector is an unsupervised algorithm that learns fixed-length feature representations from pieces of texts like sentences, paragraphs or whole documents, extending the word2vec model seen in Section 4.1 [61] [57]. Respectively to word2vec model, doc2vec model, represents each document (our paragraph here) by a dense vector which is trained to predict words in the document. Both word and paragraph vectors are trained by the stochastic gradient descent and backpropagation (See Section 3.3). The paragraph vector is preferred to word vector model for its capability to construct representations of variable length input sequences. The doc2vec architectures represents two algorithms: the **distributed memory** (dm) and the **distributed bag of words** (dbow).

### Distributed Memory Model

In the distributed memory model, respectively to word vectors that capture semantics as they are used in the prediction task of a word given the others, the paragraph vectors DM participate in the same prediction task, *given many contexts (words)* sampled from the paragraph.

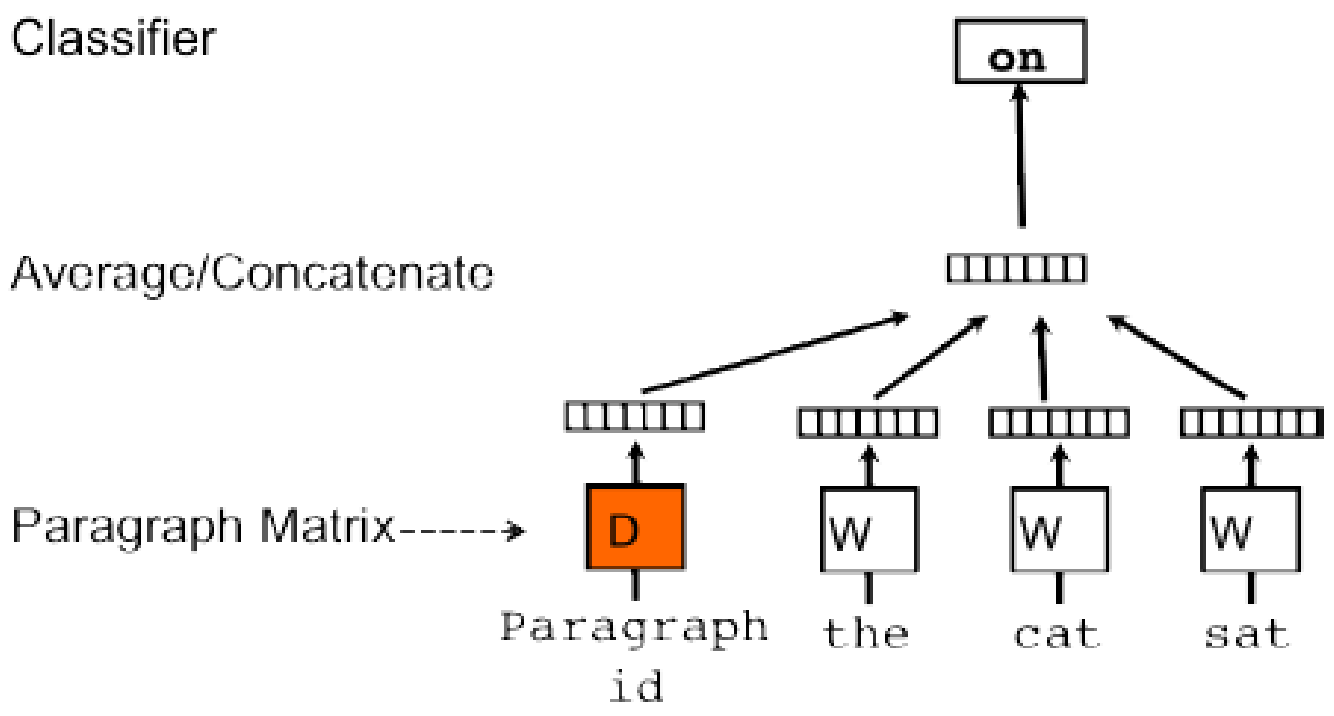

*Figure 25: Paragraph Vector: A distributed memory model (PV-DM)*

As we can see in *Figure 25*, the distributed model architecture is very closely to the back of words word vector model. Precisely, every paragraph is mapped to a unique vector, represented by a column (id) in matrix D, while, every word is also mapped to a unique vector-col in matrix W. Subsequently, the difference to the (BOW) word vector falls to the different 'h' in (4.1.1.5) equation, as now its constriction is depended on both D and W. Importantly, the contexts (D, W) are fixed-length with the goal to predict the next word and are represented in a sliding window over the paragraph. The intuition is that matrix *D demeans* as a memory that captures what is missing from the current context or *the topic of the paragraph.* This is why the model is called Distributed Memory Model of Paragraph Vectors (PV-DM) [59].
Assuming that we have N paragraphs and M words as our vocabulary, and each paragraph is modelled in a p-dimension and each word is modeled to a q dimension, then the model will have N x p + M x q parameters, without the softmax parameters.

We should state that the paragraph vector is shared only on the contexts (W) generated by the same paragraph, while the word vector matrix (W) is global across all paragraphs.
The algorithm of the PV-DM model is presented below:

1. Training to get word vectors W, softmax weights U, b and D on already seen paragraphs (W and D are trained using stochastic gradient descent and the gradient is obtained via backpropagation).
2. Inference to get D paragraph vector for new paragraphs (we add columns in D and we apply gradient descend on D, while keeping W, U, b stable. The prediction is used using a standard classifier, usually logistic regression h).

### Distributed Bag of Words

The Distributed Bag of Words (PV-DBOW) model for obtaining paragraph vectors is very close to the skip-gram model of the word vectors models. Precisely, PV-DBOW forces the model to predict words randomly *samples from the paragraph* in the output, while ignoring the context words in the input, as it can be seen in Figure 26:

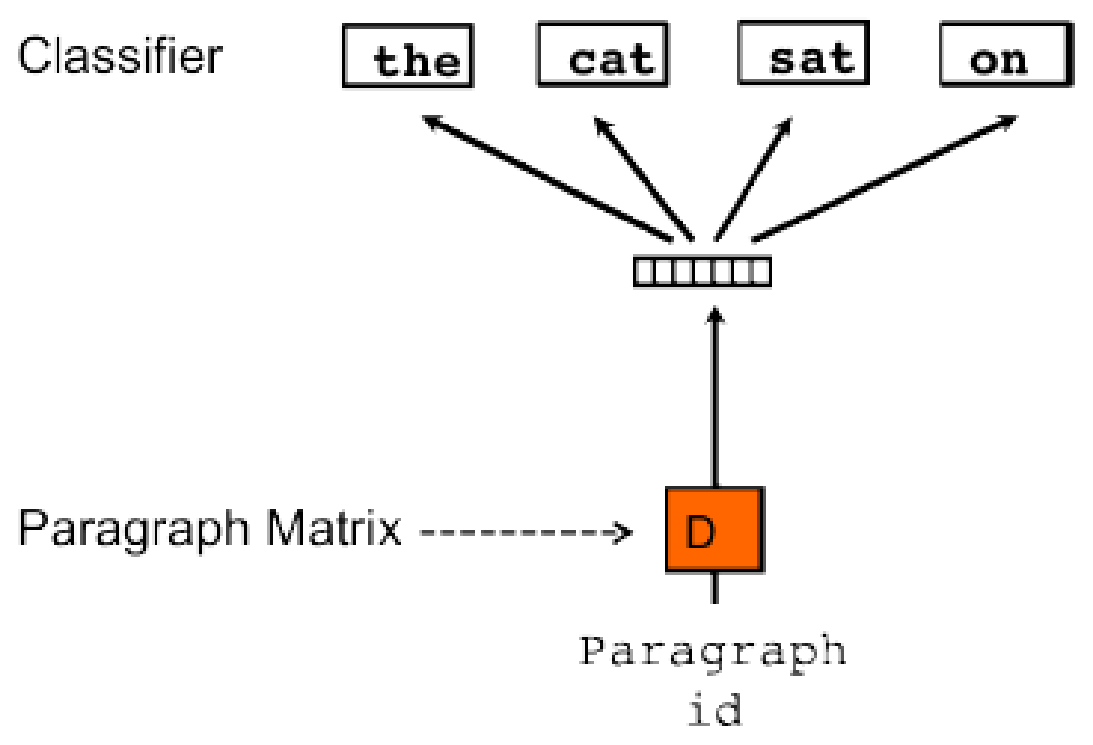

*Figure 26: Paragraph Vector: Distributed Bag Of Words (PV-DBOW)*

The algorithm of the PV-DBOW model is presented below:

1. Training to get word paragraph D, softmax weights U, b on already seen paragraphs (D is trained using stochastic gradient descent and the gradient is obtained via backpropagation).
2. For each iteration of stochastic gradient descent, we sample a text window, then sample a random word from the text window and form a classification task given the Paragraph Vector.

The advantages of using paragraph vectors instead of other models for feature representation are mainly held in their capability to confront some of the weaknesses of bag-of-words models, namely the semantics of the words. As presented in Section 4.1, word vectors are able to capture strong relationships among words and concepts. Moreover, paragraph vectors are able to consider, even in a small context the word order as n-gram does (N-grams save a lot information about a corpus and accordingly to their N, can capture also the word order). Though, n-grams models creates a very high-dimensional representation, which according to Mikolov in [59] tends to generalize poorly.

As far as the two paragraph models are concerned, generally, Mikolov's experiments in [59] [57] have shown that the Distributed Memory model outperforms the Distributed Bag of Words model. Meanwhile, the Distributed Bag of Words model stores less data than the Distributed Memory model, as the only parameters to be stored are the softmax weights and not the word vectors W.

## 4.3 Using LDA to obtain paragraph vectors

In this section we point out the differences of the probabilistic LDA model and the neural network of Autoencoder, as method for representing documents' features. In this thesis, we assume paragraphs as the documents of a corpus. Latent Dirichlet Allocation is extensively covered in Section 2.2.

Latent Dirichlet Allocation as discussed in Chapter 2 is a generative probabilistic graphical model. By that, LDA's main characteristic that differentiates it from the Autoencoder neural net, seen in Section 4.2 and in Chapter 3, is the sense of 'probabilistic' as LDA assumes that the generation of the observed data comes from data from a larger population. Precisely, LDA assumes a Bayesian Statistical method and each document as a distribution over topics and each topic as a distribution over terms (observed words) and tries to figure at the posterior distribution as represented in (2.2.3.1) equation:

$$p(\theta, z|w, \alpha, \beta) = \frac{p(\theta, z, w|\alpha, \beta)}{p(w|\alpha, \beta)}$$

In this way, Latent Dirichlet allocation captures each paragraph-document-(movie plot in our experiment) as a topic distribution via a probabilistic procedure. Manly in order to find these features (2.2.3.1) LDA uses Sampling methods (Gibbs Sampling, in this thesis). On the other hand, Autoencoder captures the topics of a paragraph by learning distributed representation for the data, namely it assumes that every output is a non-linear function of the linear combination of all the hidden units and finds these features as a result of local optimization methods as gradient descent. The risk with local optimization methods is that we may end up with a local instead of a global minima as my function is non convex, in contrast with sampling methods that presents higher possibilities to lead us to a global minima or a good local optima [64]. We can observe the results of these methods in feature representation in our Case study where we train a corpus of 25203 movie plots in Chapter 6. Moreover, LDA as a topic modeling algorithm tries to capture salient statistical patterns in the co-occurrence of words within documents [65], while autoencoder as a neural net tries to predict features, given some data (documents and/or words) via non-linear functions.

Moreover, the most significant difference of LDA and Autoencoder is that the LDA is the state of the art parametric model, while Autoencoder is a non-parametric model, as LDA lies on the hyperparameters $\alpha$ and $\beta$, while Autoencoder simply to the word weights and biases.

# Chapter 5

# 5. Case Study: Movies Modelling

We have chosen to implement the Latent Dirichlet Allocation and the Autoencoder in a corpus dataset, precisely a movie dataset as to test our hypothesis and illustrate the extracted features for a better human apprehension. Moreover, though we are waiting Autoencoder to give better conceptual results than LDA, we mainly study the two algorithms on movies as to find out how and what movies do they propose respectively to IMDB that uses for this purpose collaborative filtering. Is there a better recommendation among the three? The results are presented in Chapter 6.

## 5.1 Movie Database

For our case study, we chose the CMU Movie Summary Corpus, an excellent 46 MB dataset of 42.306 movie plot summaries extracted from Wikipedia and also of associated metadata. The data was collected by Bamman et al. [66] in use for theirs paper about finding personas in film characters, a very interesting paper and was supported in part by U.S. National Science Foundation grant IIS-0915187. At the beginning we were trying to find out a dataset with IMDB summaries, but we end up that IMDB plots were much too big incorporating our algorithms with features that do not give precise categories. Wikipedia summaries are perfect and more appropriate for our scope. The movie metadata the dataset offers helps us a lot I our preprocessing. Precisely, the metadata includes the following: Wikipedia movie ID, Freebase movie ID, movie name, movie release date, movie box office revenue, movie runtime, movie languages, movie countries, movie genres.

## 5.2 Preprocessing

The preprocessing concludes the several steps to bring our data in the appropriate format to be inputs of our algorithms. The preprocessing is an important process in data mining and machine learning as the representation and the quality of the data is first and foremost before running an analysis [67].
So, first we use java to remove stop words form the summaries. The stop words are the most common words occurring in text that give no additional concept and can cause problem in the performance of a model. Precisely we have removed most common stop words from a MySQL stop word list that displayed the bigger list. Moreover, we excluded numbers, non-asci characters, and most common occurring names. Surprisingly enough, when at first we had not yet removed the most common names, the LDA had shaped several topics with relevant names, grouping common names together.
Remaining words were filtered by frequency using the Term Frequency - Inverse Document Frequency score (TF-IDF). TF-IDF measures the importance of a word in a corpus as seen in Section 2.1.1. It increases with the number of occurrences in the document and decreases with the frequency in the corpus. We compute TF-IDF for each word of each document-plot in the corpus and keep the 100,000 words with the highest score (from the 113617 total no. of words in the corpus).

Then for a further proper dataset, we use the imdbpy python library. As far as we would finally compare our results with the IMDB proposed movies, we removed movies that were not found, by double checking those movies from the dataset that had not IMDB movie ID, and then with imdbpy if they indeed nonexistent.
Last we kept the movies that had more than 4.0/10 as IMDB rating.

All tests are carried out on this configuration: Personal Computer with 8.00 GB Ram, processor Intel®Core™ i7 CPU 64- bit-4 running at 2.93 GHz.
We used Java for the preprocessing and python for LDA and Autoencoder as to use the python's library, gensim, which is highly proposed as it is really simple and fast https://radimrehurek.com/gensim/.

## 5.3 Learning features using LDA

For the Latent Dirichlet Allocation implementation we used the genism library from python. It is a remarkable, easy library that implements both LDA and Autoencoder and it is used in most feature extraction applications. In Appendix A we can see the algorithm used.
We use the symmetric version of LDA as after running several examples, the symmetric gave us better topic interpretations for our corpus. Precisely, the symmetric values for the LDA are:
$\alpha$ = 1.0/ num. of Topics = 1.0/50 = 0.02
$\beta$ = 0.1 (The $\beta$ hyperparameter is noted as $\eta$ (eta) in genism)
Reviewing the theory of LDA in Chapter 2, the $\alpha$ is a K dimension vector of positive reals and represents the prior weight of topic K in a document. Namely $\alpha$ affects the $\Theta$ (document-topic) distribution.
Symmetric LDA assumes that these weights are similar for all topics and it is usually a number less than 1. The closer to 0 the $\alpha$ is, the more sparse will be the topics, i.e. we are going to have fewer topics per document. Accordingly, parameter $\beta$ is V-dimension vector of positive reals and represents the prior weight of each word in a topic. Namely $\beta$ affects the $\Phi$ (topic-word) distribution. Respectively, gensim library sets symmetric $\beta$ as 1.0/num. of topics = 0.02. The smaller the $\beta$, the more sparse will be the word distribution, so fewer words per topic.

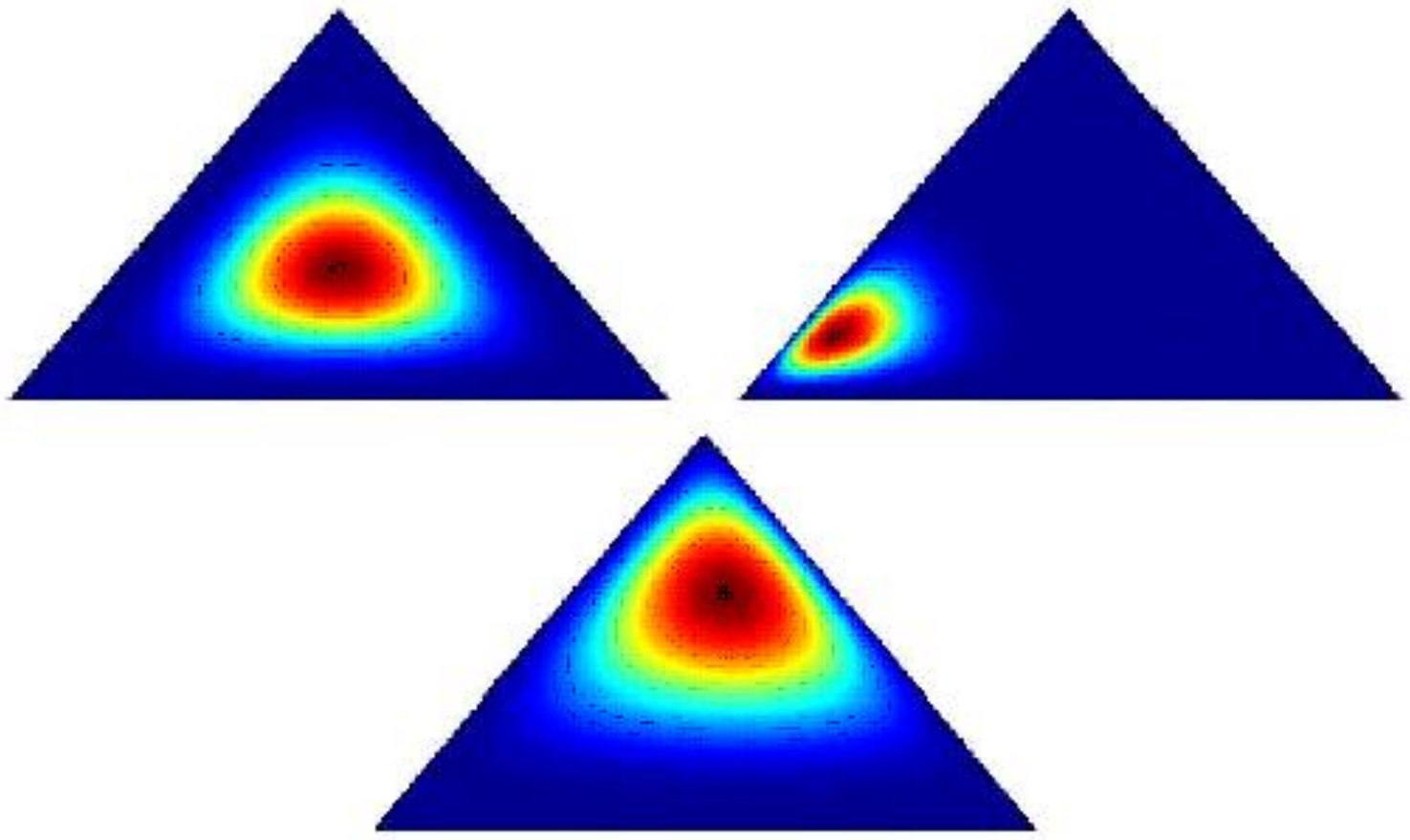

*Figure 27: 3-dimensional Dirichlet distributions with different α parameters. In the first case a=(4; 4; 4); in the second a= (8; 1; 1); while in the third a= (1; 4; 1). Colors with more red indicate larger probabilities [25].*

The algorithm runs in constant memory and the size of the training corpus does not affect memory footprint. Subsequently, it can process corpora larger than RAM.

## 5.4 Learning features using Autoencoder

As in LDA the same for Autoencoder we use the gensim python library for feature extraction. Precisely Autoencoder makes use of the new class named Doc2Vec in the latest gensim release of 0.10.3, developed by Mikolov et al. [59]. Doc2vec (aka paragraph2vec, aka sentence embeddings) modifies the word2vec algorithm to unsupervised learning of continuous representations for larger blocks of text, such as sentences, paragraphs or entire documents. The Doc2Vec class extends gensim's original Word2Vec class. If we review the theory of representing features with Autoencoder in Chapter 4, we will find information about the paragraph vectors, namely the doc2vec algorithm we are going to use for our implementation. Mikolov et al. [59] in their experiments showed that the distributed memory algorithm performs noticeably better than the distributed bag of words algorithm, so the PV-DM is the default algorithm when running Doc2vec. We left this value default.

The input to Doc2Vec is an iterator of LabeledSentence objects. Each such object represents a single sentence, and consists of two simple lists: a list of words and a list of labels (here the id of each doc-movie):

```
sentence = LabeledSentence(words=[u'some', u'words', u'here'], labels=[u'SENT_1'])
```

The algorithm then runs through the sentences iterator twice: once to build the vocabulary and once to train the model on the input data, *learning a vector representation for each word and for each label* in the dataset.

With the current implementation, all label vectors are stored separately in RAM. In the case above with a unique label per sentence, this causes memory usage to grow linearly with the size of the corpus, which may or may not be a problem depending on the size of your corpus and the amount of RAM available on your box.

## 5.5 K-Nearest Neighbors

K-Nearest Neighbors is a non-parametric algorithm for classification or regression [68] and is highly used in pattern recognition, as it is one of the simplest machine learning algorithms.

KNN is usually selected as a method for supervised learning, with the output of KNN classification to be a class membership and the output of KNN regression to be the property value of our object. In this thesis we use sklearn's library implementation of Nearest Neighbors (*see Appendix C*, K Nearest Neighbors implementation) that uses an unsupervised learner for implementing neighbor searches.

The algorithm receives as input-vectors the probabilities of each movie plot (an array where rows represent one movie plot of size 25,203 and columns represent the probabilities of each movie for all topics (topics=50)) and the number of K nearest neighbors. In general a larger K suppresses the effects of noise, but makes the classification boundaries less distinct. Here, we use K=20 to capture the most relevant movies.

As a Nearest Neighbor Algorithm we use the Brute Force Brute, which computes the distances between all pairs of points in the dataset, with each force query time to grows as O[DN], where N are the number of movie plots and D the dimensions, namely the fifty (50) topics. We mainly choose the Brute force algorithm because the query time in this algorithm is largely unaffected by the value of K and is unchanged by data structure, contradictory to the Ball tree and KD tree algorithms that the sklearn library offers. In the tree algorithms, if D>= 20 leads to queries that are slower than brute force algorithm http://scikit-learn.org/stable/.

Moreover the distances are measured with cosine similarity instead of the automatic option of the Euclidean distance as in the last option all features, here movies, could be represented in the same radius from the chosen item-movie. With cosine similarity we manage to capture more accurate distances [69].

The cosine of two vectors can be derived by using the Euclidean formula:

$$\mathbf{a} \cdot \mathbf{b} = \|\mathbf{a}\| \, \|\mathbf{b}\| \cos\theta \qquad (5.5.1)$$

Supposing we have two vectors A and B, we compute the cosine similarity, cos(θ) as:

$$\text{similarity} = \cos(\theta) = \frac{A \cdot B}{\|A\|\|B\|} = \frac{\sum_{i=1}^{n} A_i \times B_i}{\sqrt{\sum_{i=1}^{n} (A_i)^2} \times \sqrt{\sum_{i=1}^{n} (B_i)^2}} \qquad (5.5.2)$$

## 5.6 t-SNE

T-Distributed Stochastic Neighbor Embedding constitutes an awarded method for dimensionality reduction, perfect suited for visualization of high-dimensional datasets that was proposed by L.V. Maaten and G.E.Hinton in 2008 [70] as an amelioration of Stochastic Neighbor Embedding first developed by Hinton and Roweis in 2002 [71]. T-SNE differs to SNE as it gives better visualizations by reducing the central crowd points and by an easier way of optimization. Moreover in [70] is shown that T-SNE visualizations are much better than other non-parametric visualization techniques, including Sammon mapping, Isomap, and Locally Linear Embedding. It has been widely used in many areas such as in music analysis [72], cancer research [73], and bioinformatics [74]. T-SNE represents each object by a point in a two-dimensional scatter plot, and arranges the points in such a way that similar objects are modeled by nearby points and dissimilar objects are modeled by distant points. L.V. Maaten assumes in [70] that when you construct such a map using t-SNE, you typically get much better results than when you construct the map using something like principal components analysis or classical multidimensional scaling, because primarily t-SNE mainly focuses on appropriately modeling small pairwise distances, i.e. local structure, in the map and then because t-SNE has a way to correct for the enormous difference in volume of a high-dimensional feature space and a two-dimensional map. As a result of these two characteristics, t-SNE generally produces maps that provide much clearer insight into the underlying (cluster) structure of the data than alternative techniques.

Unlikely to the widely known and used dimensionality reduction technique of PCA, which is a linear technique, T-SNE is a non-linear method and subsequently it keeps the low-dimensional representations of very similar datapoints close together, when linear methods focus just on keeping the low-dimensional representations of dissimilar datapoints far apart.
A simple version of t-Distributed Neighbor Embedding is presented below [70]:

**Algorithm 2**: Simple version of t-Distributed Stochastic Neighbor Embedding.

**Data**: data set $\mathcal{X} = \{x_1, x_2, \ldots, x_n\}$,
cost function parameters: perplexity $Perp$,
optimization parameters: number of iterations $T$, learning rate $\eta$, momentum $\alpha(t)$.
**Result**: low-dimensional data representation $\mathcal{Y}^{(T)} = \{y_1, y_2, \ldots, y_n\}$.
**begin**
    compute pairwise affinities $p_{j|i}$ with perplexity $Perp$ (using Equation 1)
    set $p_{ij} = \frac{p_{j|i} + p_{i|j}}{2n}$
    sample initial solution $\mathcal{Y}^{(0)} = \{y_1, y_2, \ldots, y_n\}$ from $\mathcal{N}(0, 10^{-4} I)$
    **for** $t=1$ **to** $T$ **do**
        compute low-dimensional affinities $q_{ij}$ (using Equation 4)
        compute gradient $\frac{\delta C}{\delta \mathcal{Y}}$ (using Equation 5)
        set $\mathcal{Y}^{(t)} = \mathcal{Y}^{(t-1)} + \eta \frac{\delta C}{\delta \mathcal{Y}} + \alpha(t) \left( \mathcal{Y}^{(t-1)} - \mathcal{Y}^{(t-2)} \right)$
    **end**
**end**

In Algorithm 2 we see that T-SNE constitutes of two main stages. First, it constructs a probability distribution over pairs of high dimensional objects as to ascribe high probability in similar objects and infinitesimal probability to dissimilar ones.

The pairwise similarities in the low-dimensional map $q_{ij}$ are given by

$$q_{ij} = \frac{\exp\left(-\|y_i - y_j\|^2\right)}{\sum_{k \neq l} \exp\left(-\|y_k - y_l\|^2\right)}, \tag{5.6.1}$$

while the pairwise similarities in the high-dimensional space $p_{ij}$ is

$$p_{ij} = \frac{\exp\left(-\|x_i - x_j\|^2 / 2\sigma^2\right)}{\sum_{k \neq l} \exp\left(-\|x_k - x_l\|^2 / 2\sigma^2\right)} \tag{5.6.2}$$

Sometimes a datapoint $x_i$ can be an outlier, namely the values of $p_{ij}$ are small for all j as for the location in its low low-dimensional map point $y_i$ has very little effect on the cost function. We solve this problem by defining the joint probabilities $p_{ij}$ in the high-dimensional space to be the symmetrized conditional probabilities, so we set $p_{ij} = \frac{p_{j|i} + p_{i|j}}{2n}$.

Second, t-SNE defines a similar probability distribution over the points in the low-dimensional map, and it minimizes the Kullback–Leibler divergence between the two joint distributions:

$$KL(P||Q) = \sum_{i \neq j} p_{ij} \log \frac{p_{ij}}{q_{ij}} \tag{5.6.3}$$

The minimization of the cost function in Equation (5.6.3) is performed using a gradient descent method, where:

$$\frac{\delta C}{\delta y_i} = 4 \sum_j (p_{ij} - q_{ij})(y_i - y_j) \tag{5.6.4}$$

# Chapter 6

# 6. Models Evaluation

In this Chapter we will see how the three models perform in our movie corpus. For each algorithm, first we will present characteristic captured topics. Most often seen words in every topic testify the topic theme. Observing the word probabilities in each topic we can further understand the power of the topic theme, while more significantly we manage to capture movie notions that the simple movie categories cannot capture. Moreover, we can see same words in different topics but observed in a different proportion, capturing in that way the intuition behind LDA that documents exhibit multiple topics, with different word distributions. Then, we are going to dive into the movies' plots to see further relationships among movies, captured in a 2-dimensional space. Moreover, for every algorithm we will compare its recommendation power via KNN with that of the IMDB that performs collaborative filtering for movies recommendation. Last we compare the three models in all the above.

## 6.1 Symmetric LDA Evaluation

In this section, we are going to review the outcomes of the symmetric Latent Dirichlet Allocation. As described in Section 5.4 we run LDA with 50 topics, $\alpha = 0.02$ (1.0/num. of topics=50) and $\beta = 0.1$. It is very interesting to wait to see the differences in topics' sparsity between the results of the symmetric and the asymmetric LDA in Section 6.2. The source code is presented in Appendix A.

### 6.1.1 Evaluated Topics

Below, we present the most characteristic topics of the model with the twenty most likely words in each represented topic (*source code*: Appendix A, lines: 80-88). We point out the most astonishing observations in the topics and in their word distributions.

| 1 | Topic 6 | Topic 11 | Topic 12 | Topic 15 |
|---|---|---|---|---|
| 2 | 0.106 school | 0.034 wife | 0.015 world | 0.029 father |
| 3 | 0.027 high | 0.028 husband | 0.012 professor | 0.027 family |
| 4 | 0.024 students | 0.013 life | 0.009 one | 0.018 mother |
| 5 | 0.024 teacher | 0.012 married | 0.006 project | 0.013 life |
| 6 | 0.023 student | 0.011 marriage | 0.005 dream | 0.012 daughter |
| 7 | 0.021 college | 0.011 affair | 0.005 human | 0.010 children |
| 8 | 0.018 class | 0.009 woman | 0.005 even | 0.009 young |
| 9 | 0.014 girls | 0.008 relationship | 0.005 new | 0.008 home |
| 10 | 0.009 friends | 0.008 couple | 0.004 dreams | 0.008 brother |
| 11 | 0.008 boys | 0.006 divorce | 0.004 time | 0.007 years |
| 12 | 0.008 university | 0.005 lover | 0.004 program | 0.007 old |
| 13 | 0.008 year | 0.005 daughter | 0.004 computer | 0.007 sister |
| 14 | 0.007 principal | 0.005 becomes | 0.004 basketball | 0.007 parents |
| 15 | 0.007 anderson | 0.005 man | 0.004 real | 0.006 one |
| 16 | 0.005 san | 0.005 new | 0.003 ability | 0.006 time |
| 17 | 0.005 kids | 0.004 two | 0.003 society | 0.005 man |
| 18 | 0.005 group | 0.004 ex | 0.003 research | 0.005 two |
| 19 | 0.005 professor | 0.004 women | 0.003 experiment | 0.005 child |
| 20 | 0.004 classmates | 0.004 suicide | 0.003 state | 0.005 girl |
| 21 | 0.004 parents | 0.004 friend | 0.003 become | 0.005 marriage |

As we can see Topic 6 is the “strongest” movie captured in our corpus from the symmetric LDA as the term “school” holds almost the 10% of the topic vocabulary, declaring that this topic has to do with school life, and more precisely high school life as the next more likely word in topic is the term “high”. We can observe similar word-topic-theme relationships in all the presented topics. Accordingly, Topic 11 is related to relationships as the most occurent words are wife, husband marriage but is also highly to bad relationships as in the most likely words in topic are words like divorce , lover, suicide with 0.4% topic occurrence. More surprisingly, LDA manages to capture ideas behind movies. For example in Topic 11 again, we may have to do with wife and husband, but “wife” has 1,4% bigger occurrence in this topic, so someone could estimate that movies related to this topic, observe relationship more from the feminin aspect. Respectively Topic 15 that mainly is a family movie categoy, shows “father” to have the most likely occurrence in this topic 1,1% bigger rate than this of the “mother”. “Father” in these movies seems to have a bigger role than “mother”.

| 1 | Topic 18 | Topic 20 | Topic 22 | Topic 24 |
|---|---|---|---|---|
| 2 | 0.034 horse | 0.024 ship | 0.046 team | 0.013 mother |
| 3 | 0.026 brown | 0.017 island | 0.034 game | 0.011 old |
| 4 | 0.024 black | 0.013 crew | 0.016 fight | 0.010 young |
| 5 | 0.021 white | 0.012 boat | 0.011 coach | 0.010 baby |
| 6 | 0.019 ranch | 0.011 captain | 0.010 match | 0.009 relationship |
| 7 | 0.018 dragon | 0.010 find | 0.010 football | 0.009 sex |
| 8 | 0.018 doc | 0.010 sea | 0.009 play | 0.009 new |
| 9 | 0.013 johnson | 0.009 water | 0.008 competition | 0.008 woman |
| 10 | 0.011 indians | 0.007 back | 0.008 players | 0.007 becomes |
| 11 | 0.010 horses | 0.006 one | 0.007 player | 0.007 girl |
| 12 | 0.009 cattle | 0.006 rescue | 0.007 training | 0.007 home |
| 13 | 0.008 ride | 0.006 castle | 0.007 first | 0.007 man |
| 14 | 0.007 rider | 0.005 return | 0.007 baseball | 0.006 year |
| 15 | 0.007 land | 0.005 group | 0.007 final | 0.006 boy |
| 16 | 0.006 winter | 0.005 cave | 0.006 one | 0.006 women |
| 17 | 0.006 indian | 0.004 moon | 0.006 wins | 0.006 life |
| 18 | 0.006 blue | 0.004 ocean | 0.006 tournament | 0.006 two |
| 19 | 0.006 knight | 0.004 land | 0.006 winning | 0.006 begins |
| 20 | 0.006 rides | 0.004 two | 0.006 arts | 0.006 one |
| 21 | 0.006 farm | 0.004 magic | 0.005 boxing | 0.006 child |

| 1 Topic 27 | 1 Topic 30 | 1 Topic 31 | 1 Topic 32 | 1 Topic 34 |
|---|---|---|---|---|
| 2 0.039 money | 2 0.015 house | 2 0.032 princess | 2 0.070 mr | 2 0.025 band |
| 3 0.017 company | 3 0.010 room | 3 0.025 lord | 3 0.049 mrs | 3 0.025 music |
| 4 0.014 transaction | 4 0.008 back | 4 0.019 palace | 4 0.035 church | 4 0.019 song |
| 5 0.012 business | 5 0.008 car | 5 0.011 slave | 5 0.021 priest | 5 0.011 sing |
| 6 0.010 pay | 6 0.008 body | 6 0.009 ling | 6 0.021 god | 6 0.010 singing |
| 7 0.007 work | 7 0.007 find | 7 0.009 kingdom | 7 0.016 lady | 7 0.009 singer |
| 8 0.006 deal | 8 0.007 one | 8 0.008 snake | 8 0.005 sister | 8 0.009 concert |
| 9 0.006 sell | 9 0.007 finds | 9 0.007 baba | 9 0.005 st | 9 0.008 playing |
| 10 0.006 casino | 10 0.007 man | 10 0.006 slaves | 10 0.005 catholic | 10 0.008 play |
| 11 0.006 buy | 11 0.006 night | 11 0.006 bush | 11 0.005 sisters | 11 0.006 piano |
| 12 0.006 workers | 12 0.006 dead | 12 0.006 wish | 12 0.004 reverend | 12 0.006 club |
| 13 0.005 gambling | 13 0.006 door | 13 0.006 aladdin | 13 0.004 convent | 13 0.005 record |
| 14 0.005 million | 14 0.006 goes | 14 0.006 magic | 14 0.004 cousins | 14 0.005 stage |
| 15 0.005 vegas | 15 0.005 tells | 15 0.006 sultan | 15 0.004 nun | 15 0.005 songs |
| 16 0.005 race | 16 0.005 next | 16 0.005 wishes | 16 0.003 angels | 16 0.005 performance |
| 17 0.005 owner | 17 0.005 later | 17 0.005 evil | 17 0.003 heaven | 17 0.005 radio |
| 18 0.005 factory | 18 0.005 inside | 18 0.005 mo | 18 0.003 pope | 18 0.005 group |
| 19 0.004 wins | 19 0.005 head | 19 0.005 abu | 19 0.003 pastor | 19 0.005 perform |
| 20 0.004 manager | 20 0.005 sees | 20 0.004 divine | 20 0.003 children | 20 0.005 members |
| 21 0.004 bet | 21 0.005 away | 21 0.004 masters | 21 0.003 nuns | 21 0.005 plays |

Topic 27 represents in its most part (3.9%) movies about money and watching the next shown terms, we take further information for the movies plots. Topic 31 presents is linked to animations that mainly have to do with princesses and palaces. Topic 32 is the religious topic, while topic 34 deals with music.

| 1 Topic 36 | 1 Topic 37 | 1 Topic 38 | 1 Topic 43 | 1 Topic 49 |
|---|---|---|---|---|
| 2 0.024 show | 2 0.035 plane | 2 0.020 chinese | 2 0.017 dog | 2 0.015 earth |
| 3 0.016 new | 3 0.019 air | 3 0.020 master | 3 0.012 bugs | 3 0.007 city |
| 4 0.009 play | 4 0.018 flight | 4 0.020 sword | 4 0.012 back | 4 0.007 planet |
| 5 0.009 director | 5 0.017 pilot | 5 0.018 emperor | 5 0.008 gets | 5 0.007 space |
| 6 0.009 stage | 6 0.014 race | 6 0.018 china | 6 0.007 tries | 6 0.007 alien |
| 7 0.007 actor | 7 0.013 aircraft | 7 0.013 japanese | 7 0.007 away | 7 0.006 team |
| 8 0.007 television | 8 0.012 crash | 8 0.013 li | 8 0.007 head | 8 0.006 power |
| 9 0.007 hollywood | 9 0.012 flying | 9 0.011 ma | 9 0.006 tree | 9 0.006 time |
| 10 0.007 career | 10 0.011 fly | 10 0.010 japan | 10 0.006 mouse | 10 0.006 world |
| 11 0.007 tv | 11 0.008 engine | 11 0.009 blade | 11 0.005 little | 11 0.006 control |
| 12 0.007 studio | 12 0.007 passengers | 12 0.008 clan | 12 0.005 runs | 12 0.006 machine |
| 13 0.007 role | 13 0.007 airport | 13 0.008 yang | 13 0.005 one | 13 0.006 destroy |
| 14 0.007 york | 14 0.006 crew | 14 0.008 battle | 14 0.005 comes | 14 0.005 human |
| 15 0.007 actress | 15 0.006 crashes | 15 0.008 feng | 15 0.004 starts | 15 0.005 bomb |
| 16 0.006 dance | 16 0.006 hawk | 16 0.008 samurai | 16 0.004 goes | 16 0.005 using |
| 17 0.006 producer | 17 0.006 helicopter | 17 0.006 fight | 17 0.004 around | 17 0.005 new |
| 18 0.006 performance | 18 0.006 jet | 18 0.006 liu | 18 0.004 rabbit | 18 0.005 use |
| 19 0.006 audience | 19 0.006 landing | 19 0.006 man | 19 0.004 hole | 19 0.005 nuclear |
| 20 0.005 act | 20 0.006 new | 20 0.006 wu | 20 0.004 porky | 20 0.004 escape |
| 21 0.004 set | 21 0.005 airplane | 21 0.006 iron | 21 0.004 finally | 21 0.004 powers |

Accordingly, Topic 36 is linked to shows, directors, Hollywood and Topic 37 with airplane plots. In Topic 38 we see Asian movies. In Topic 43 we assume it involves kid movies, respectively to Topic 31. Finally, the vocabulary of Topic 49 reveals movies where earth is threatened.

What we further observe is that the 20th term of every topic appears in the 0.5% of the topic vocabulary, but for the most likely term there is a highly divergence. From the first two-three most likely terms we can understand the basic thematic structure of the topics. Moreover, by capturing a number of most likely terms in a category of movie, we give a better intepretation of its contexts. In the above presented topics we have yet present six topics that mainly talk about wars. By capturing a number of terms in every topic, we manage to capture subcategories. Let watch these topics about war:

| | | | | | | | | | | | | | | |
|---|---|---|---|---|---|---|---|---|---|---|---|---|---|---|
| 1 | Topic | 16 | 1 | Topic | 21 | 1 | Topic | 25 | 1 | Topic | 35 | 1 | Topic | 41 |
| 2 | 0.021 | president | 2 | 0.029 | camp | 2 | 0.025 | murder | 2 | 0.012 | men | 2 | 0.025 | police |
| 3 | 0.014 | united | 3 | 0.019 | german | 3 | 0.024 | police | 3 | 0.009 | kill | 3 | 0.015 | gang |
| 4 | 0.014 | states | 4 | 0.018 | war | 4 | 0.017 | killer | 4 | 0.008 | escape | 4 | 0.011 | gets |
| 5 | 0.013 | american | 5 | 0.014 | nazi | 5 | 0.014 | death | 5 | 0.008 | two | 5 | 0.009 | one |
| 6 | 0.013 | government | 6 | 0.011 | jewish | 6 | 0.011 | detective | 6 | 0.007 | killed | 6 | 0.009 | drug |
| 7 | 0.010 | political | 7 | 0.011 | germany | 7 | 0.010 | killed | 7 | 0.007 | one | 7 | 0.008 | prison |
| 8 | 0.010 | miller | 8 | 0.009 | italian | 8 | 0.010 | evidence | 8 | 0.006 | back | 8 | 0.007 | brother |
| 9 | 0.009 | people | 9 | 0.009 | berlin | 9 | 0.009 | found | 9 | 0.006 | kills | 9 | 0.007 | kill |
| 10 | 0.009 | minister | 10 | 0.009 | world | 10 | 0.009 | murdered | 10 | 0.006 | take | 10 | 0.006 | money |
| 11 | 0.009 | country | 11 | 0.008 | hitler | 11 | 0.008 | investigation | 11 | 0.006 | gun | 11 | 0.006 | goes |
| 12 | 0.008 | pooja | 12 | 0.008 | ii | 12 | 0.008 | crime | 12 | 0.005 | man | 12 | 0.006 | crime |
| 13 | 0.008 | state | 13 | 0.007 | prisoners | 13 | 0.007 | dead | 13 | 0.005 | shoots | 13 | 0.006 | jail |
| 14 | 0.007 | white | 14 | 0.007 | italy | 14 | 0.007 | murders | 14 | 0.004 | help | 14 | 0.005 | officer |
| 15 | 0.006 | british | 15 | 0.007 | american | 15 | 0.006 | body | 15 | 0.004 | find | 15 | 0.005 | killed |
| 16 | 0.006 | black | 16 | 0.007 | paris | 16 | 0.006 | trial | 16 | 0.004 | away | 16 | 0.005 | help |
| 17 | 0.006 | african | 17 | 0.006 | escape | 17 | 0.006 | suspect | 17 | 0.004 | shot | 17 | 0.005 | take |
| 18 | 0.005 | prime | 18 | 0.006 | formula | 18 | 0.006 | kill | 18 | 0.004 | takes | 18 | 0.005 | comes |
| 19 | 0.005 | south | 19 | 0.006 | nazis | 19 | 0.006 | court | 19 | 0.004 | gang | 19 | 0.005 | man |
| 20 | 0.005 | english | 20 | 0.006 | resistance | 20 | 0.005 | killing | 20 | 0.004 | fight | 20 | 0.005 | criminal |
| 21 | 0.005 | america | 21 | 0.006 | communist | 21 | 0.005 | man | 21 | 0.004 | later | 21 | 0.004 | tries |

| | | |
|---|---|---|
| 1 | Topic | 45 |
| 2 | 0.020 | war |
| 3 | 0.015 | army |
| 4 | 0.012 | soldiers |
| 5 | 0.009 | men |
| 6 | 0.008 | general |
| 7 | 0.008 | military |
| 8 | 0.007 | officer |
| 9 | 0.007 | mission |
| 10 | 0.007 | attack |
| 11 | 0.006 | colonel |
| 12 | 0.006 | british |
| 13 | 0.006 | killed |
| 14 | 0.006 | american |
| 15 | 0.006 | battle |
| 16 | 0.006 | japanese |
| 17 | 0.006 | captain |
| 18 | 0.006 | soldier |
| 19 | 0.005 | orders |
| 20 | 0.005 | one |
| 21 | 0.005 | german |

All these six topics are related to war. By watching the terms shown in every one of them, we can observe six subcategories: Topic 16 is related to federal was, mainly linked to America. In Topic 21 we can see words like “camp”, ”war”, “ii”, “German”, “Jewish” understanding the relationship with the world war II more probably. Topic 35 is an adventure thriller with guns and fights. Topic 41 is linked to police-gangsters movies, closely related to Topic 25 linked to police investigations. For example both Topics includes the term “police” in high percentages, so subsequently they are closely related topics. Finally, topic 45 is the actual “war” topic, mostly related with topic 21.

### 6.1.2 Dataset Plot

Here, we plot with T-SNE the 25.203 movies of our dataset. We capture close neighborhoods of related movies. The close thematic stricture that was presented in the topics in Section 6.1.1 is now captured in a 2d space.

Capturing the whole dataset in the 2-dimensional space, we first get this representation:

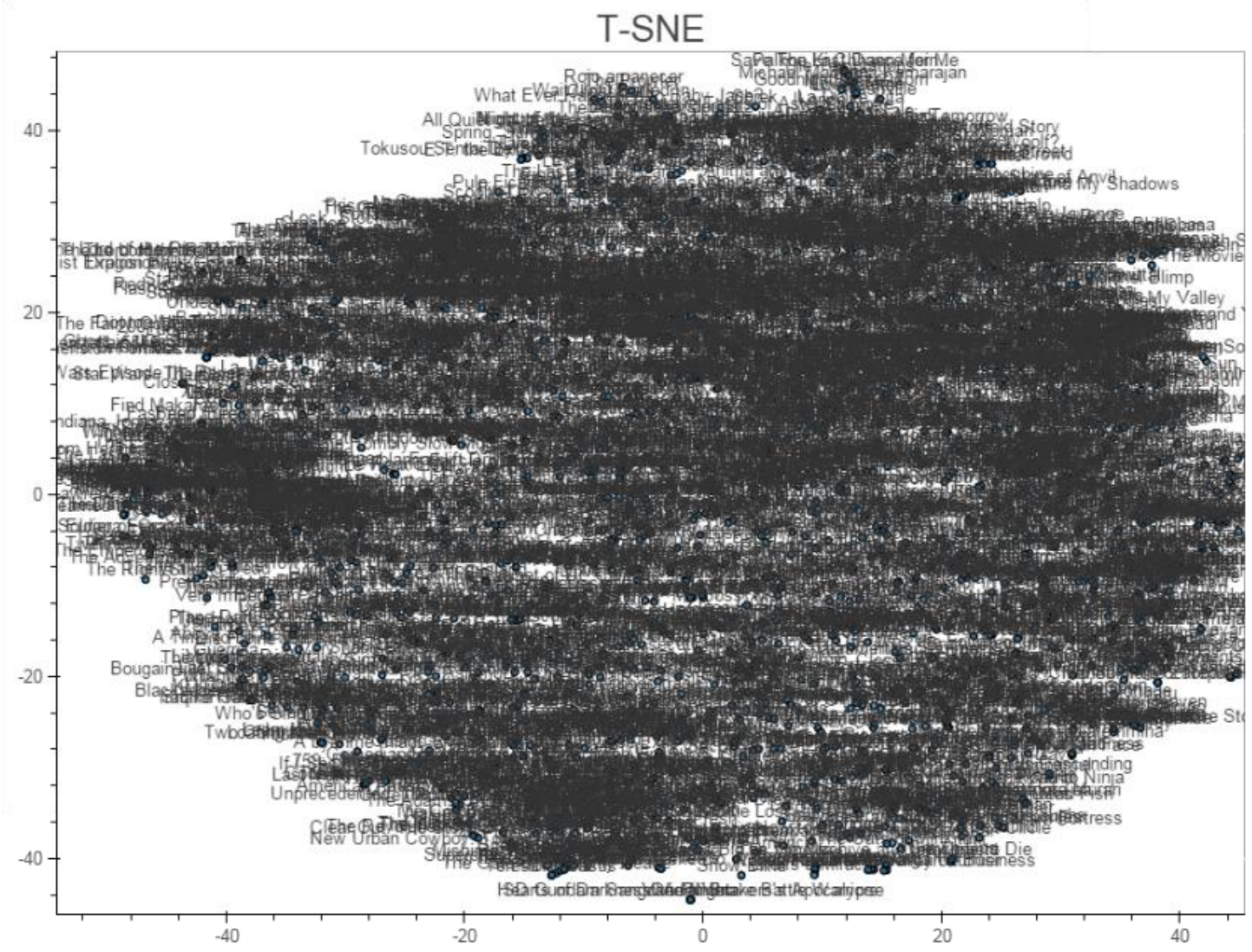

*TSNE-symLDA 1*

From the initial movies map we cannot understand that much, but we are able to see big categories of movies, which are represented as the most dark parts of the TSNE-symLDA 1. The more dense neighborhoods are mostly observed up from the middle of the TSNE, in the bottom and left of the middle. Below we can see some well-known movies' neighborhoods:

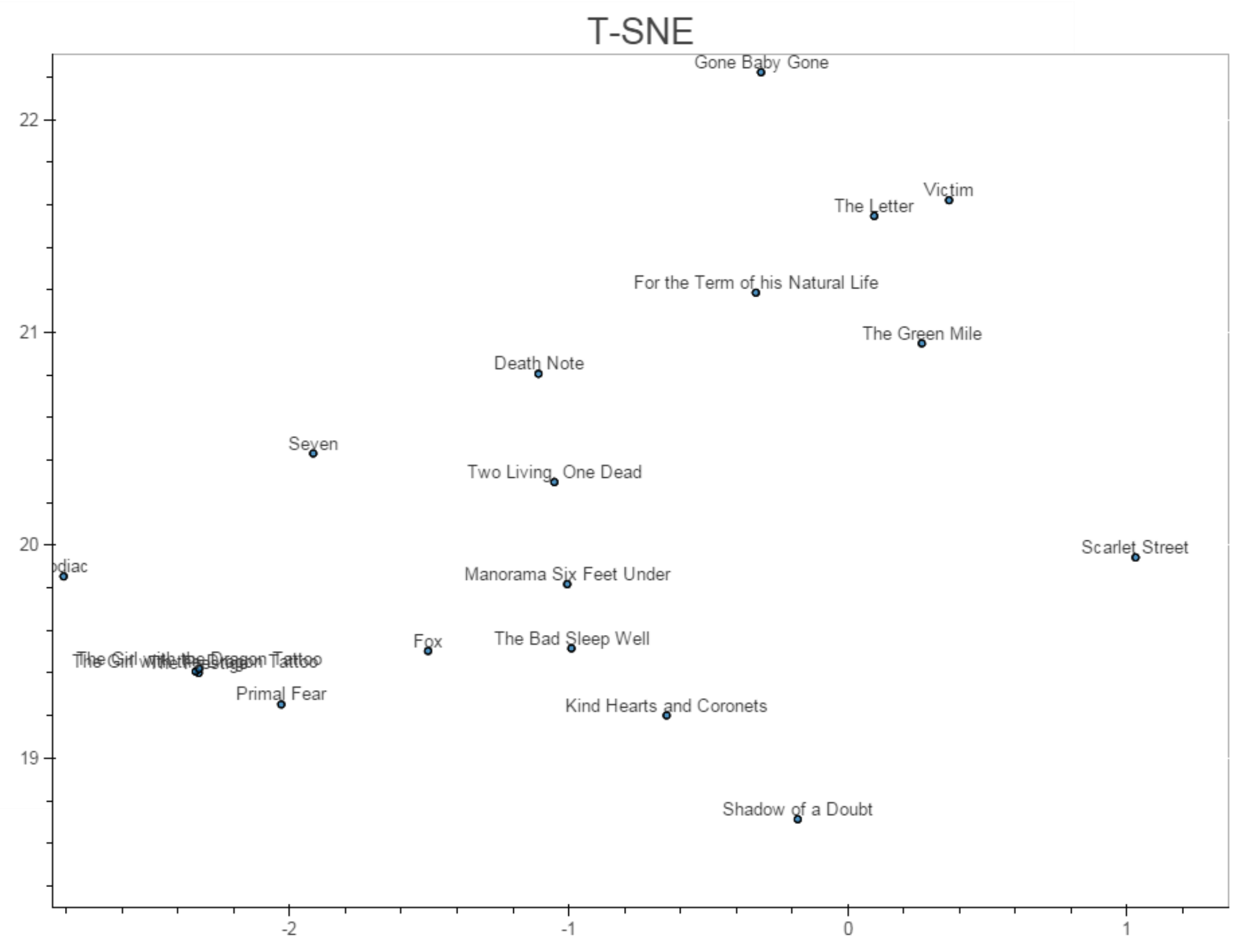

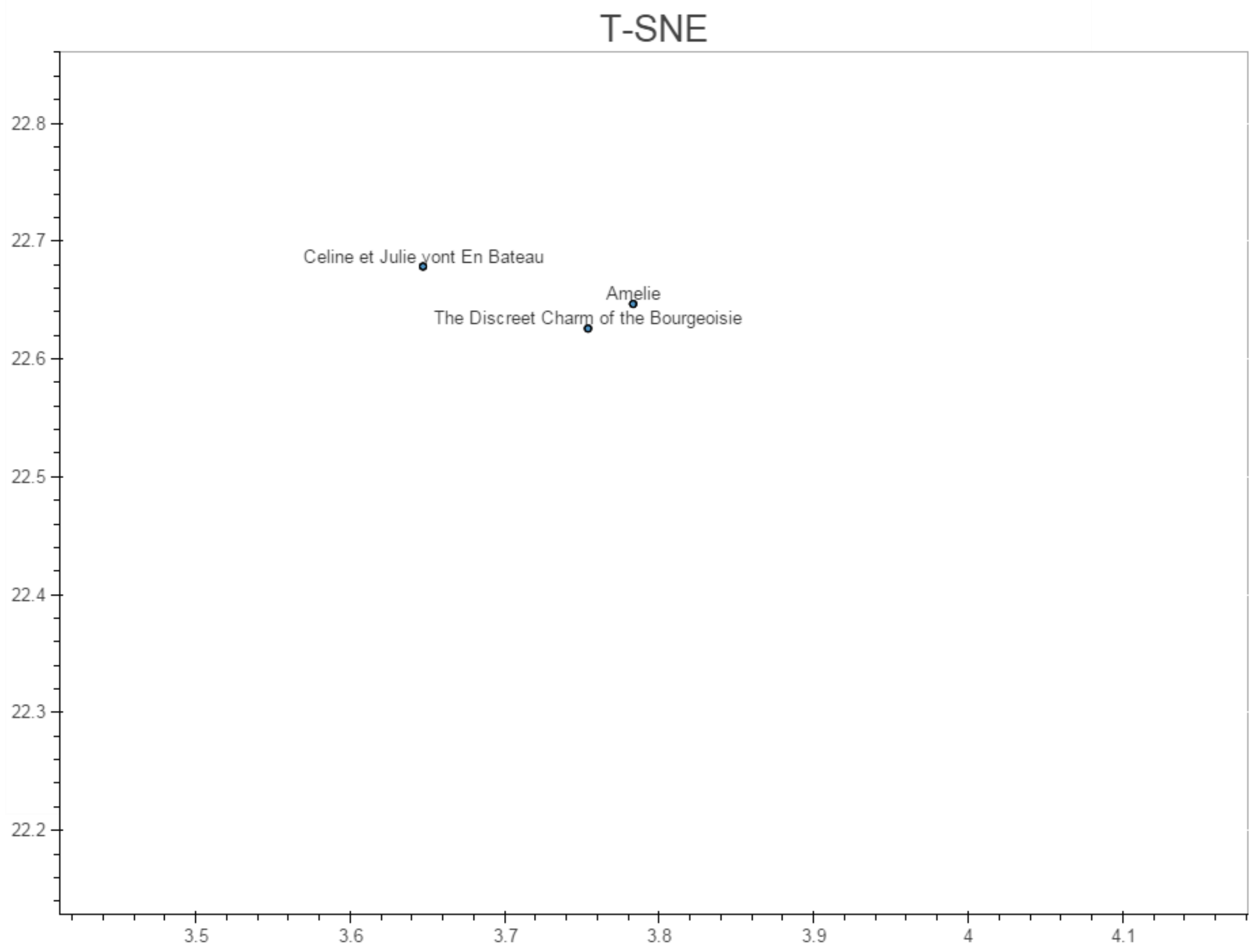

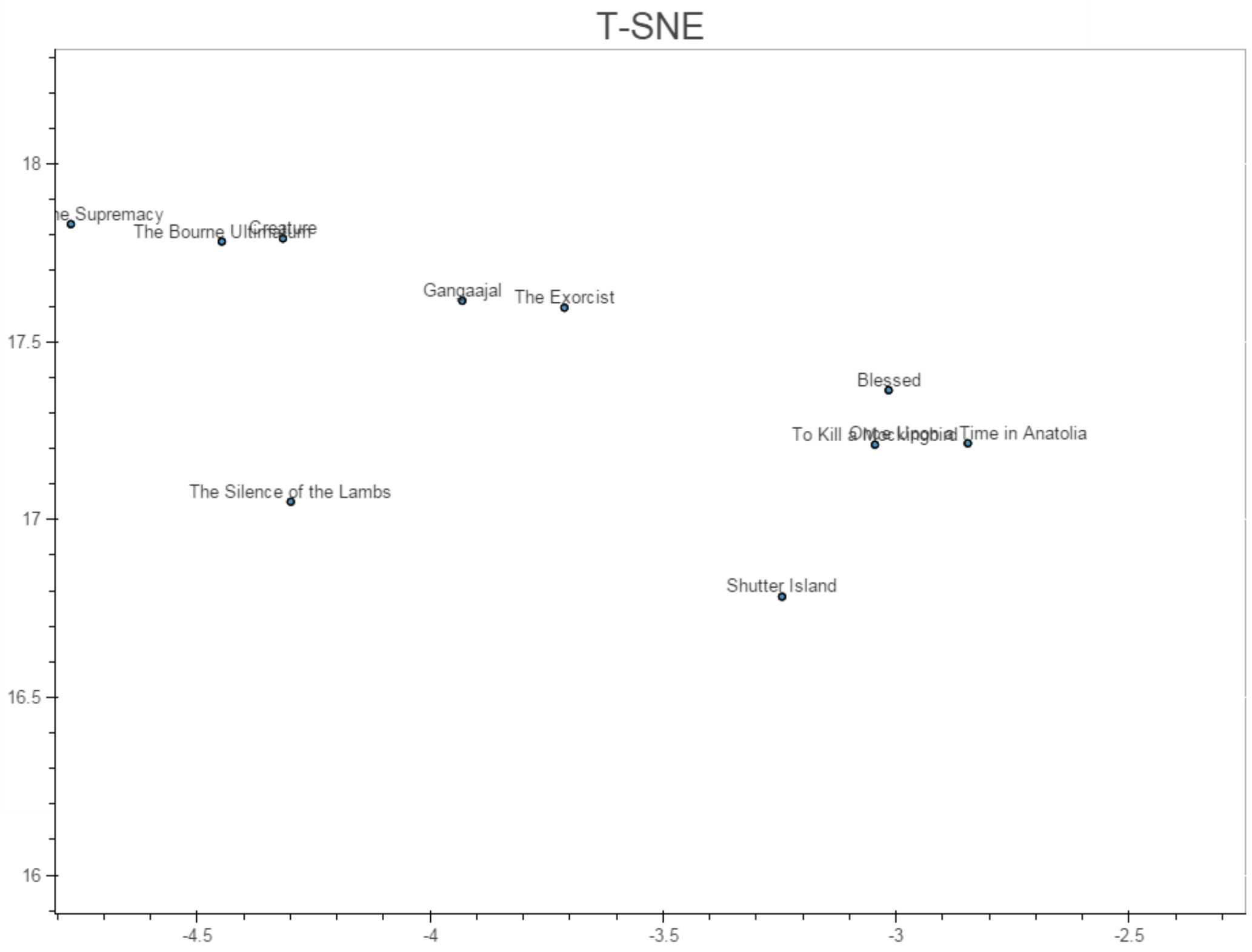

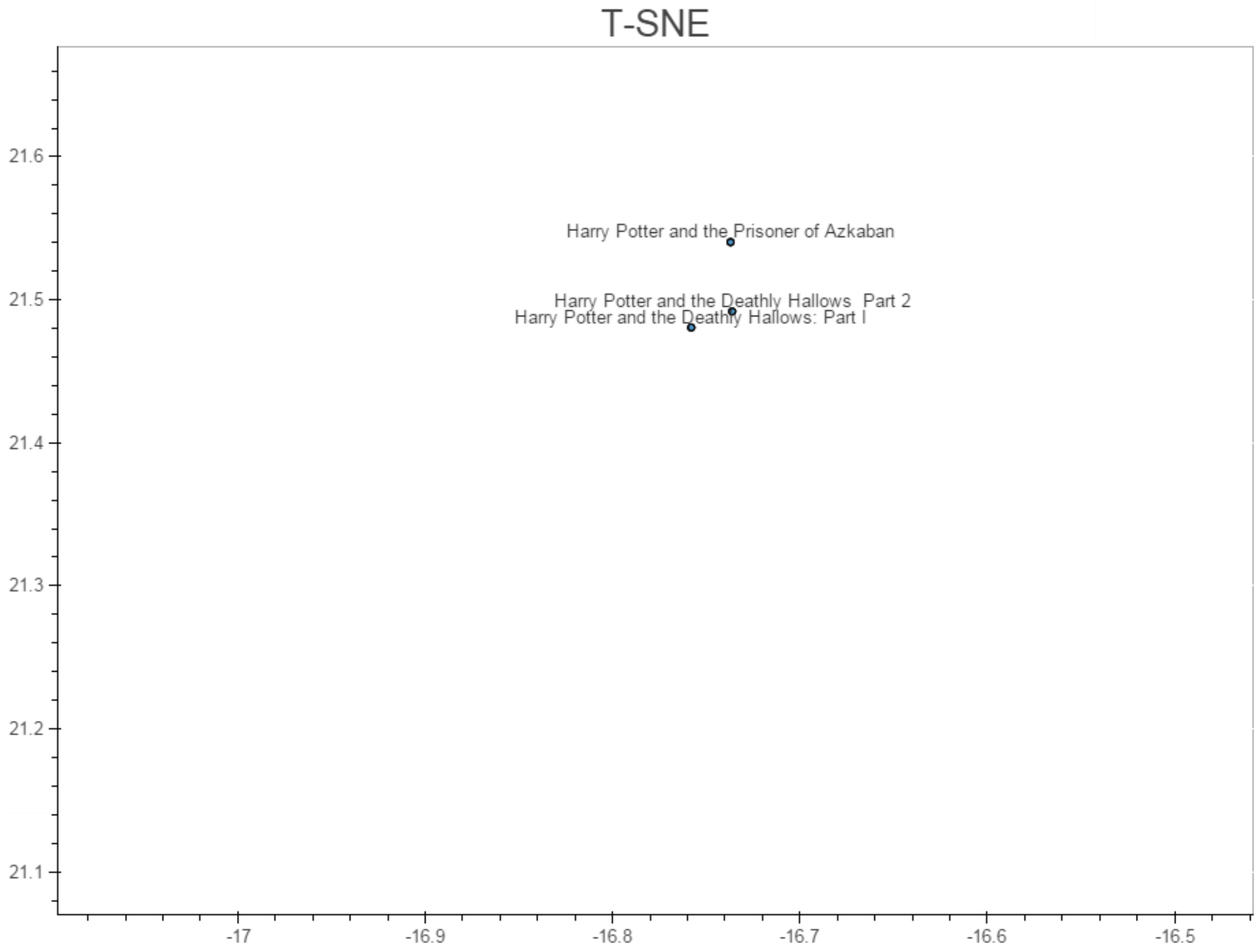
T-SNE
Harry Potter and the Prisoner of Azkaban
Harry Potter and the Deathly Hallows Part 2
Harry Potter and the Deathly Hallows: Part I
21.6
21.5
21.4
21.3
21.2
21.1
-17
-16.9
-16.8
-16.7
-16.6
-16.5

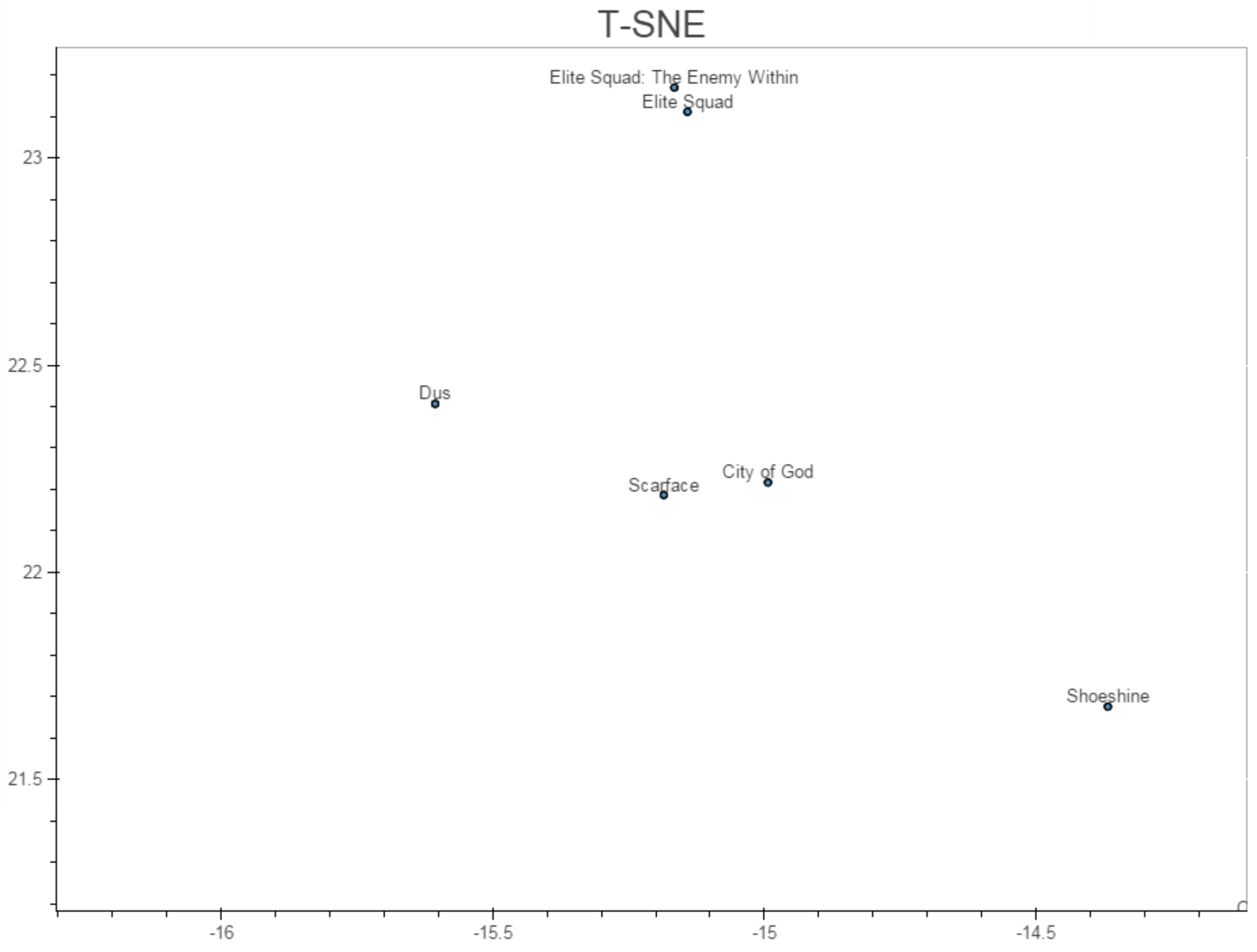

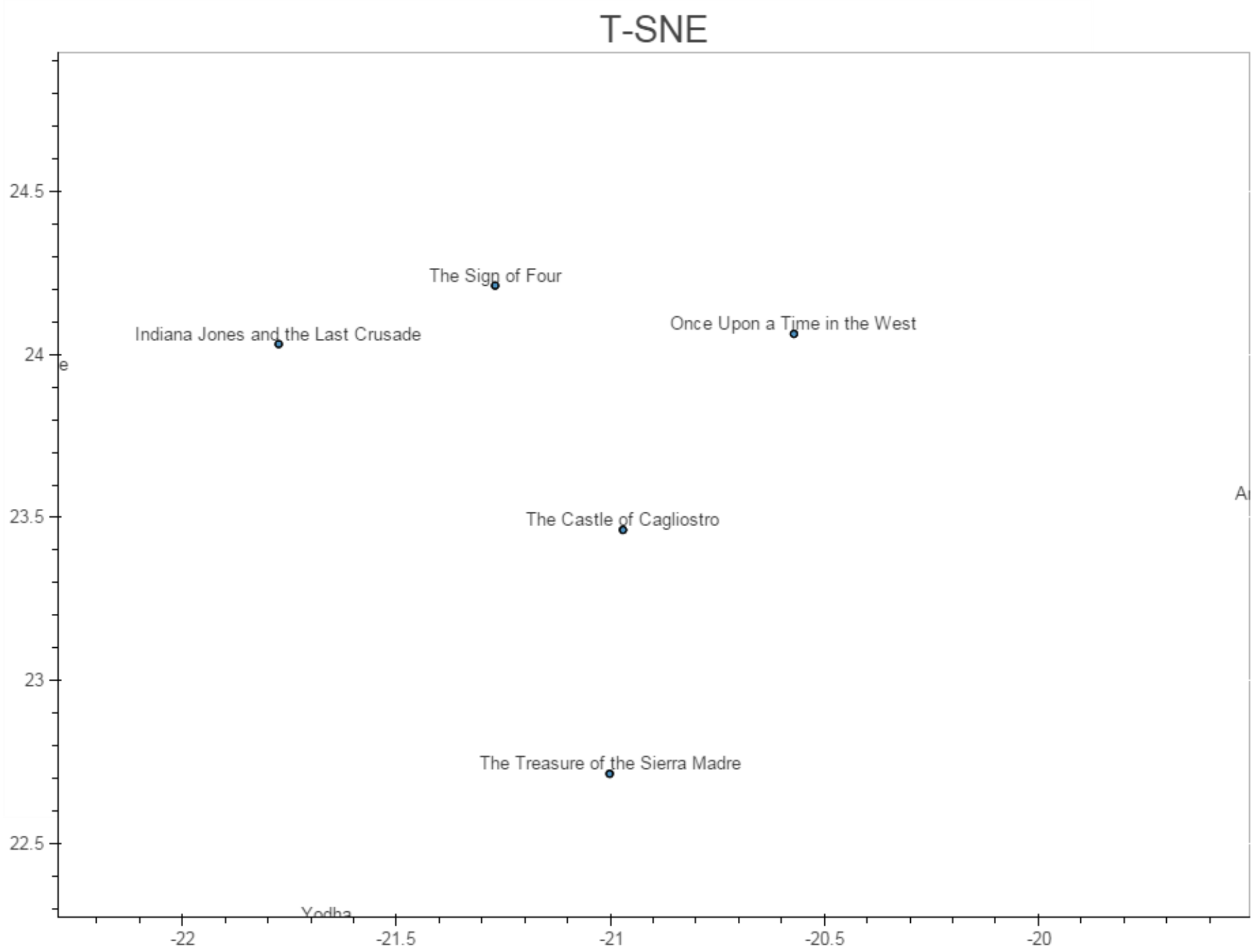
T-SNE
The Sign of Four
Indiana Jones and the Last Crusade
Once Upon a Time in the West
The Castle of Cagliostro
The Treasure of the Sierra Madre
24.5
24
23.5
23
22.5
-22
-21.5
-21
-20.5
-20

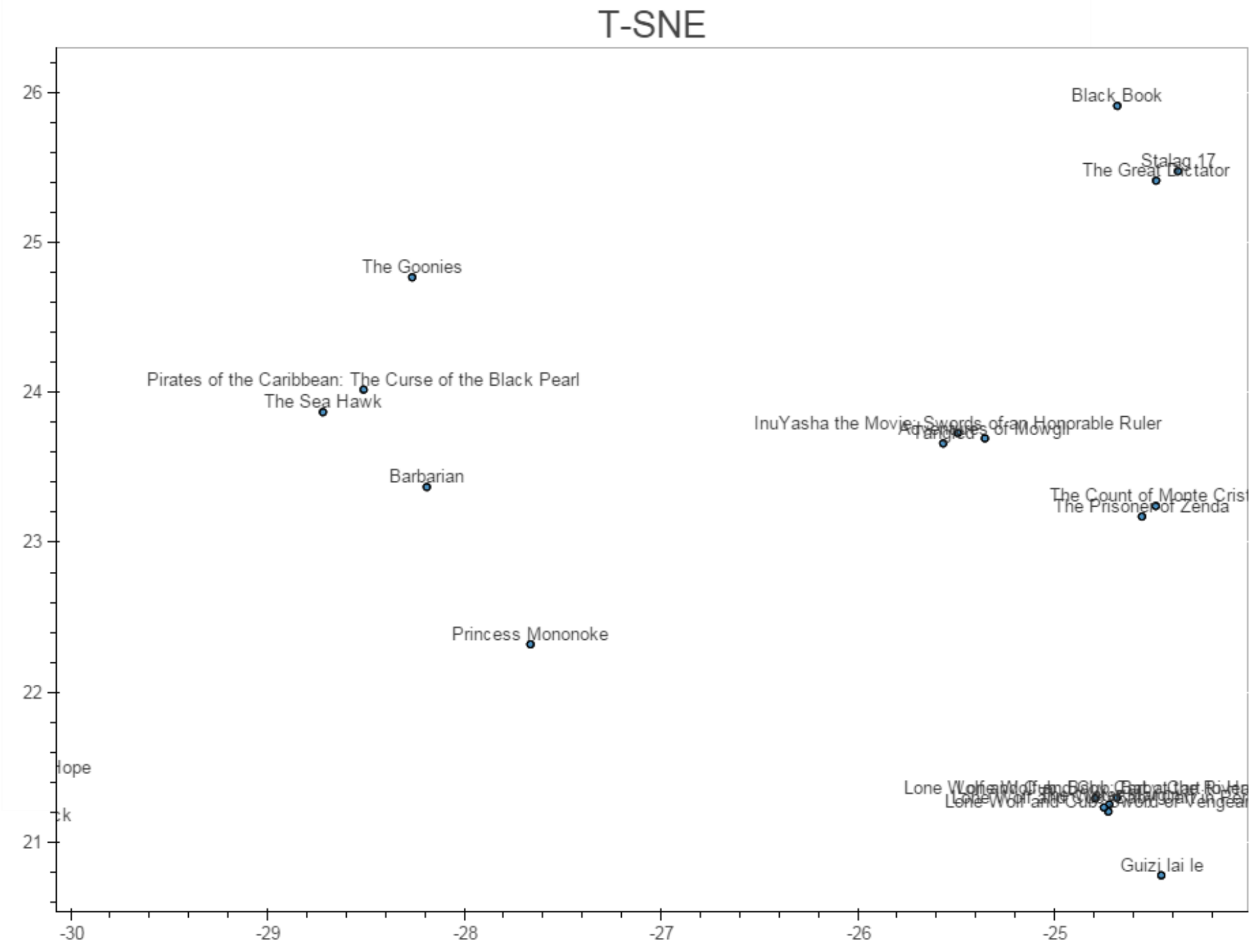
T-SNE
Black Book
Stalag 17
The Great Dictator
The Goonies
Pirates of the Caribbean: The Curse of the Black Pearl
The Sea Hawk
InuYasha the Movie: Swords of an Honorable Ruler
Barbarian
The Prisoner of Zenda
Princess Mononoke
Guizi lai le
26
25
24
23
22
21
-30
-29
-28
-27
-26
-25

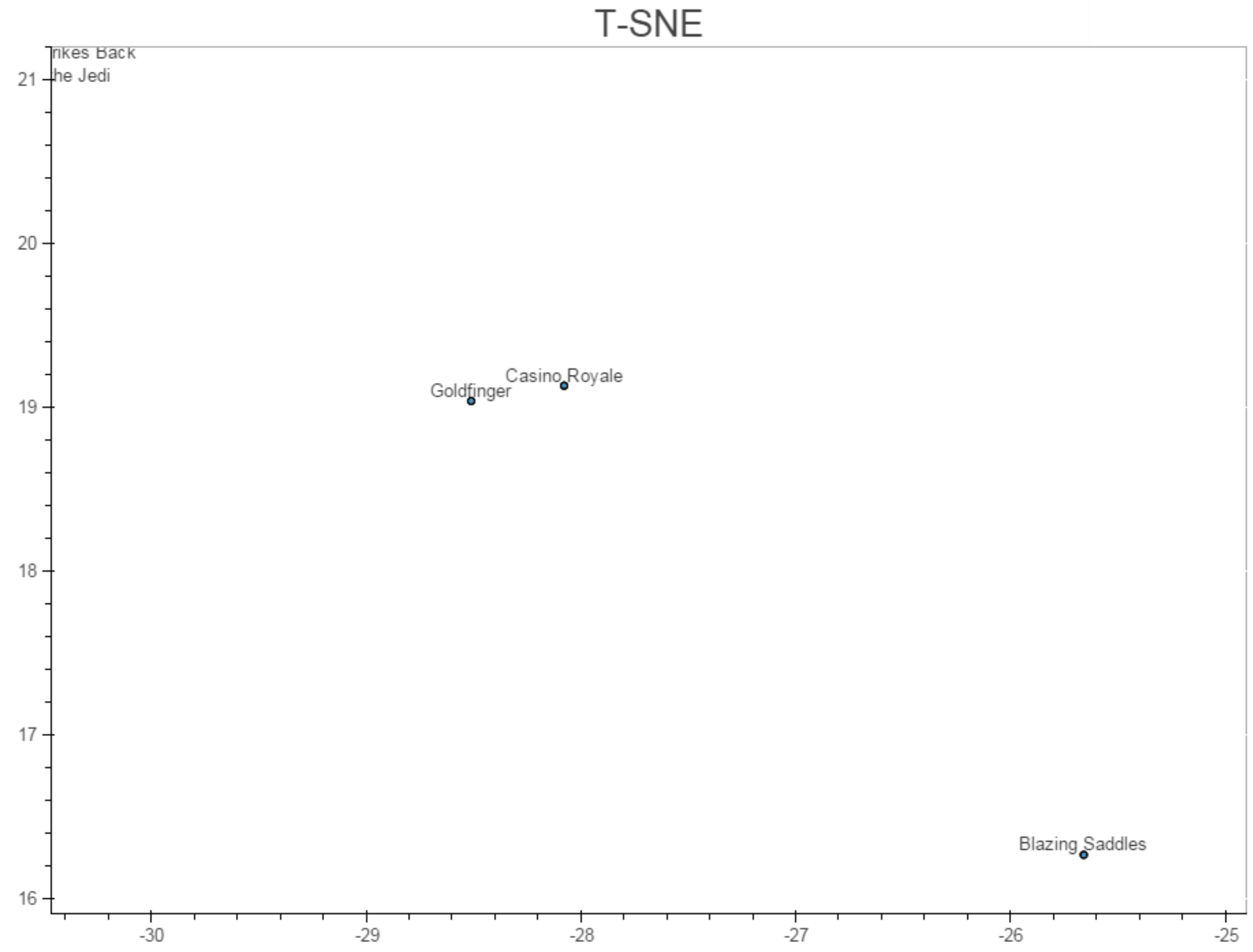

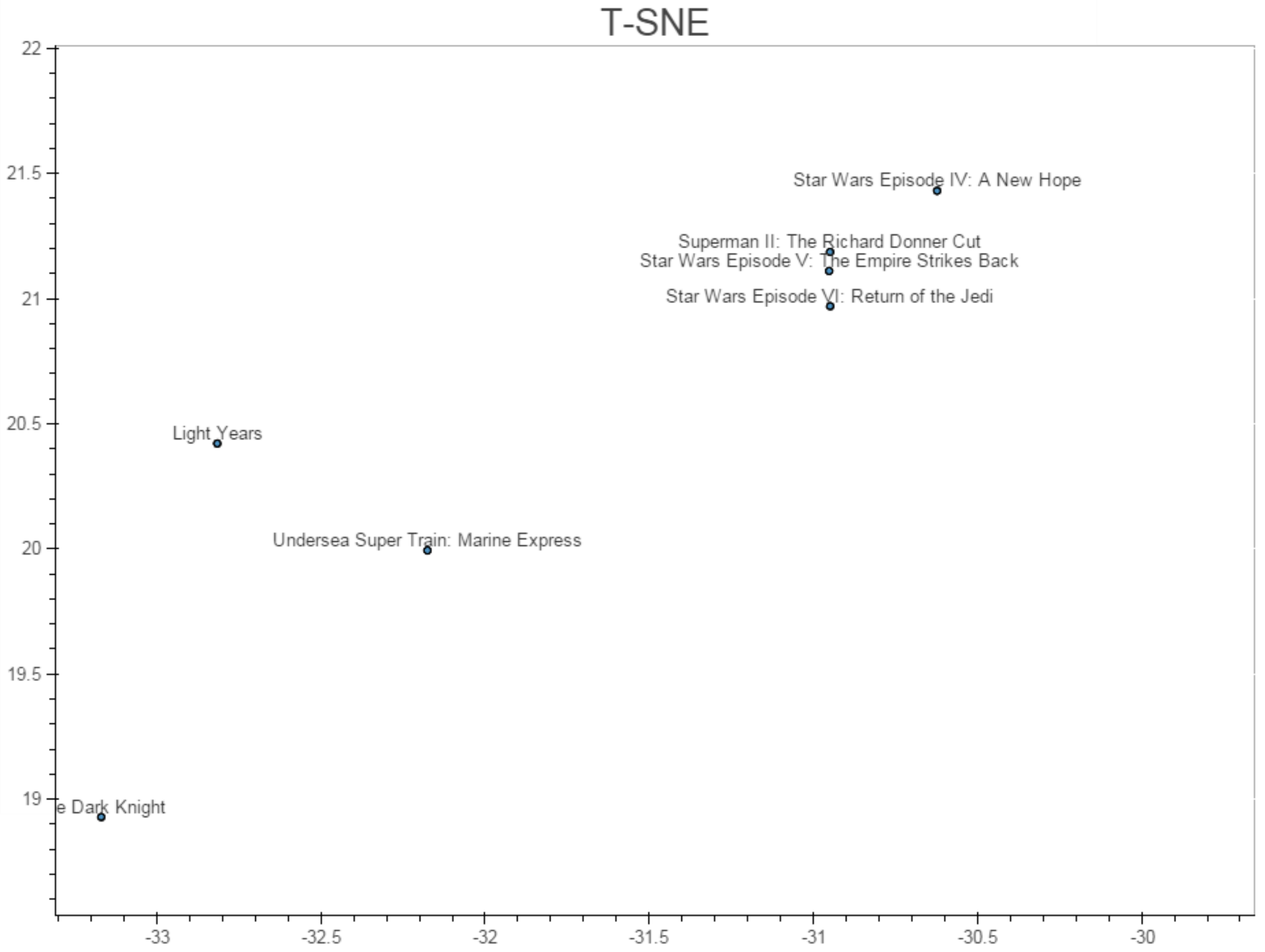
T-SNE
Star Wars Episode IV: A New Hope
Superman II: The Richard Donner Cut
Star Wars Episode V: The Empire Strikes Back
Star Wars Episode VI: Return of the Jedi
Light Years
Undersea Super Train: Marine Express
e Dark Knight
22
21.5
21
20.5
20
19.5
19
-33
-32.5
-32
-31.5
-31
-30.5
-30

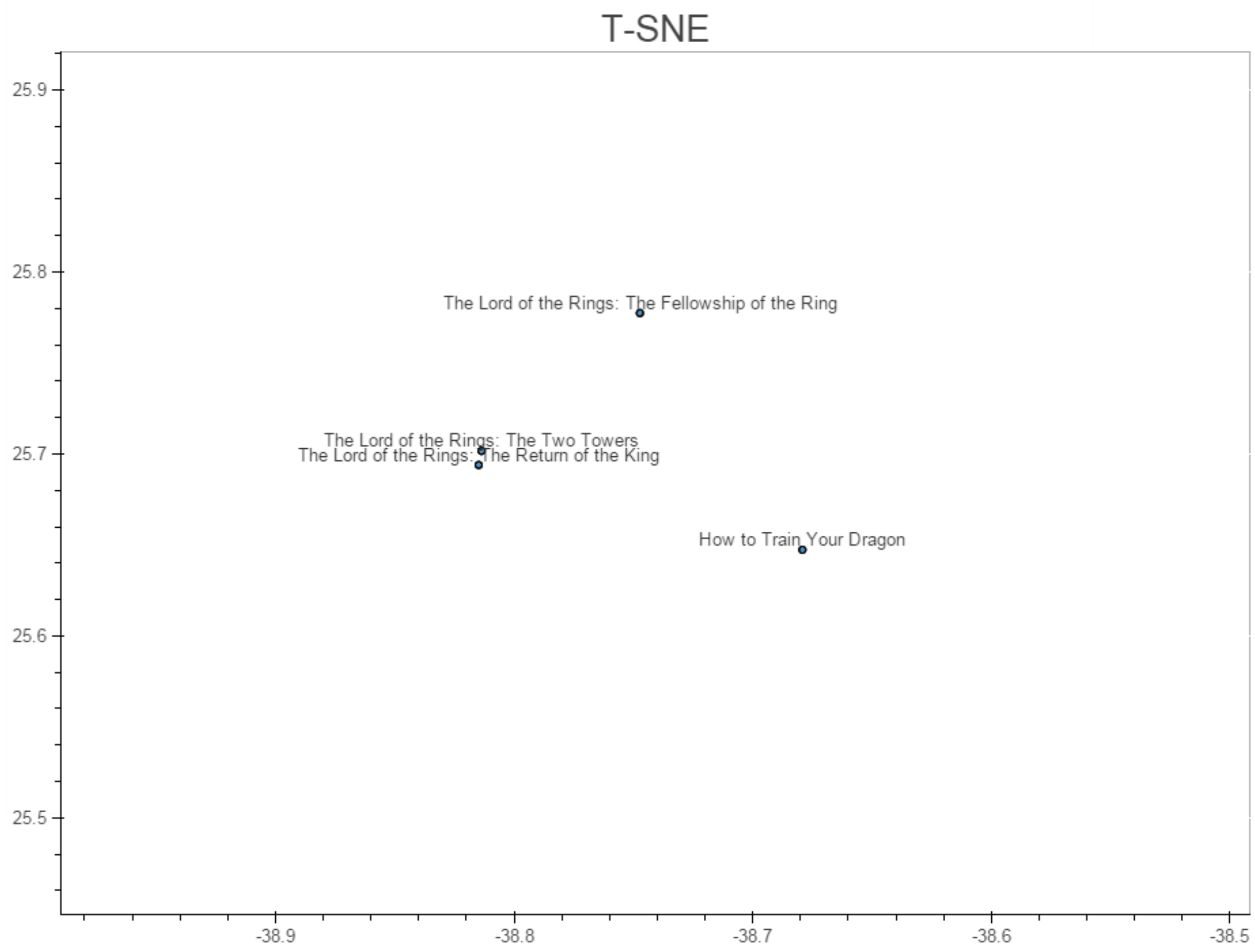
T-SNE
The Lord of the Rings: The Fellowship of the Ring
The Lord of the Rings: The Two Towers
The Lord of the Rings: The Return of the King
How to Train Your Dragon
25.9
25.8
25.7
25.6
25.5
-38.9
-38.8
-38.7
-38.6
-38.5

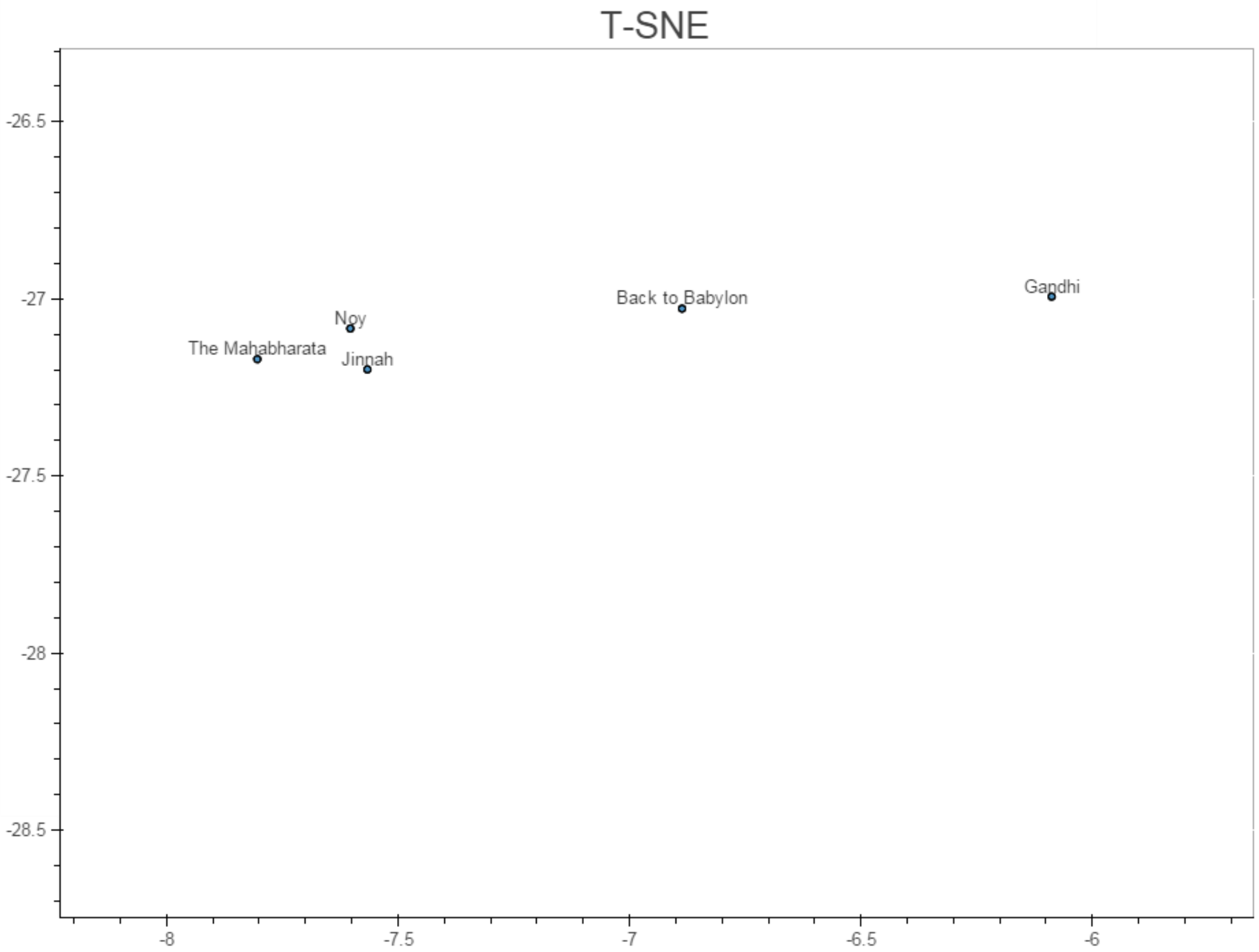

### 6.1.3 Movies Recommendation

We can better understand the above plots if we find with KNN the nearest 20 movies for a chosen movie. KNN will give us the nearest twenty movies, namely the twenty movies that have the less distance with the chosen movie. KNN is an alternate way for feature representation.
<u>Movie Selected</u>: Star Wars Episode V: The Empire Strikes Back

KNN

```
0 Star Wars Episode V: The Empire Strikes Back 0.0
1 All-Star Superman 72.5785828954
2 Superman/Batman: Apocalypse 73.436759593
3 Star Wars Episode VI: Return of the Jedi 77.9020743848
4 Superman Returns 82.6306807904
5 Superman II: The Richard Donner Cut 84.3515689817
6 Star Wars Episode IV: A New Hope 91.1640915309
7 Lego Star Wars: The Quest for R2-D2 94.0582540738
8 Superman: Doomsday 116.510269495
9 Clash of the Titans 117.532350122
10 Clash of the Titans 118.114736848
11 Wonder Woman 119.173056501
12 SpongeBob vs. The Big One 119.530075561
13 LEGO Star Wars: Revenge of the Brick 122.795088812
14 The Hitchhiker's Guide to the Galaxy 123.352669074
15 Jaws 124.998154498
16 The Silence of the Lambs 125.197199586
17 Lilo & Stitch 125.228324348
18 The Spy with My Face 125.267086381
19 Superman: Brainiac Attacks 125.530776959
20 Star Wars Episode III: Revenge of the Sith 125.553633653
```

IMDB recommendations

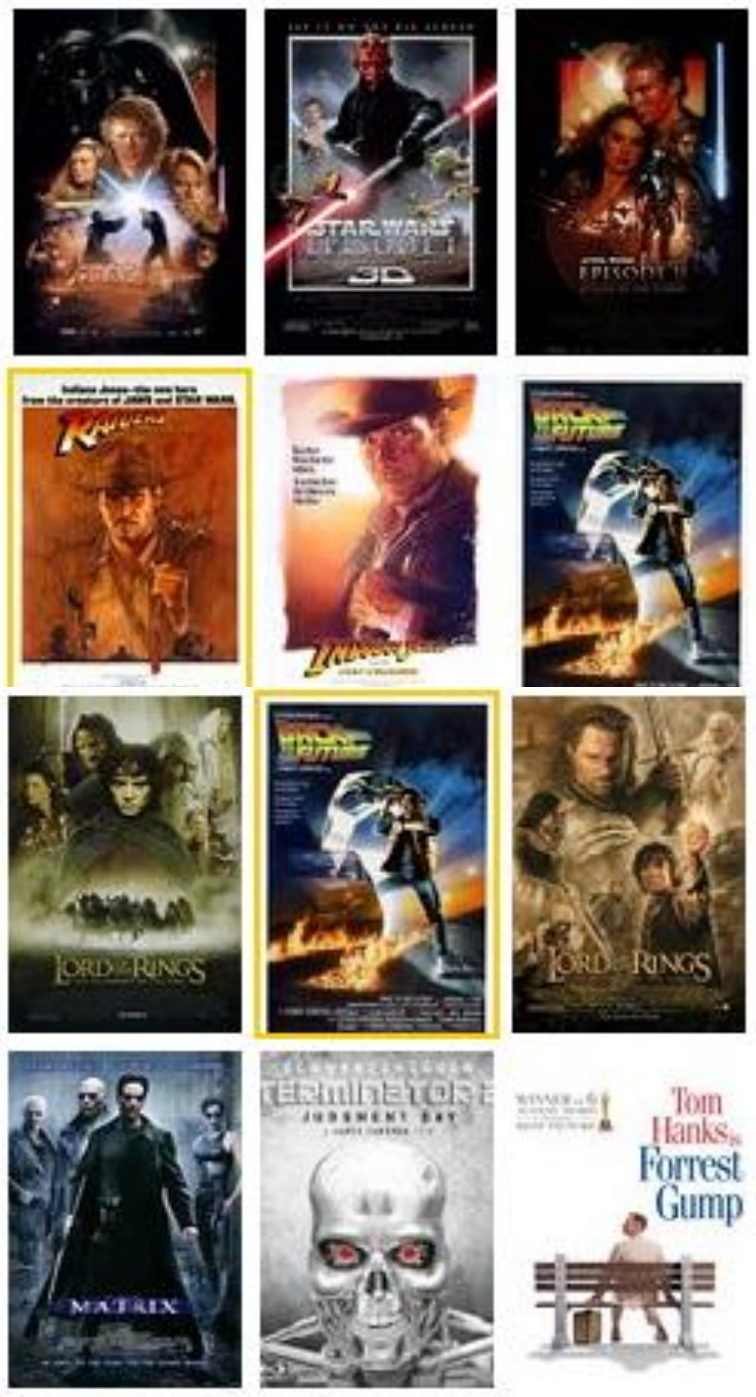

We observe that symmetric LDA depicts as closest movies to “Star Wars Episode V: The Empire Strikes Back” mainly movies relevant to space terms. IMDB proposes movies relevant with adventure which is the highly concept in Star Wars. At this point we witness the bag-of-word-assumption weakness to capture higher linguistic terminologies and semantics as discussed in 4.2. On the other hand KNN proposed movies are conceptually linked to the selected movie in a way. IMDB surprisingly proposes the Forrest Gump according to the user evaluation of collaborative filtering.
<u>Movie Selected</u>: The Lord of the Rings: The Fellowship of the Ring
IMDB recommendations

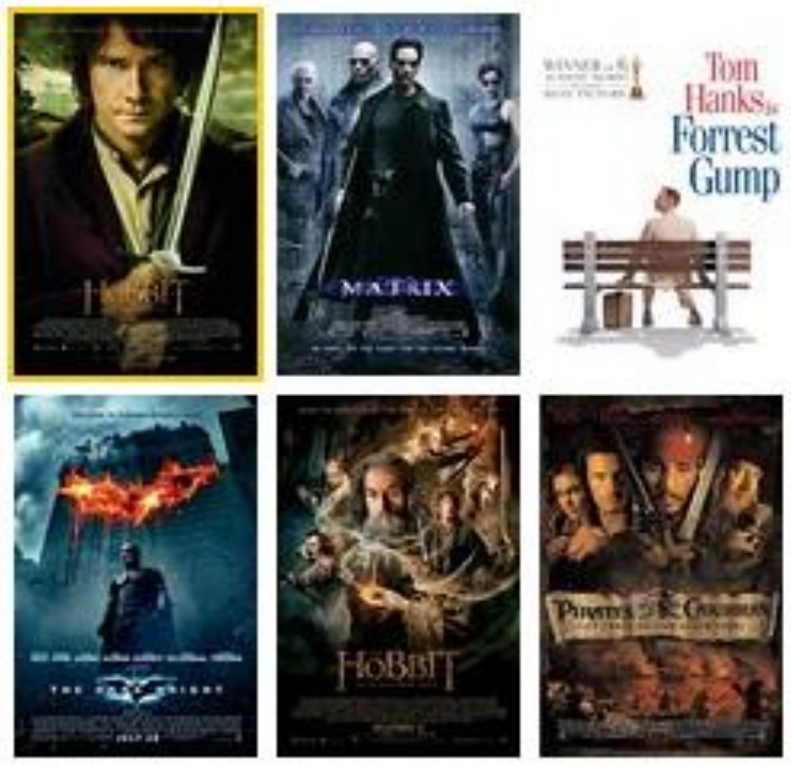

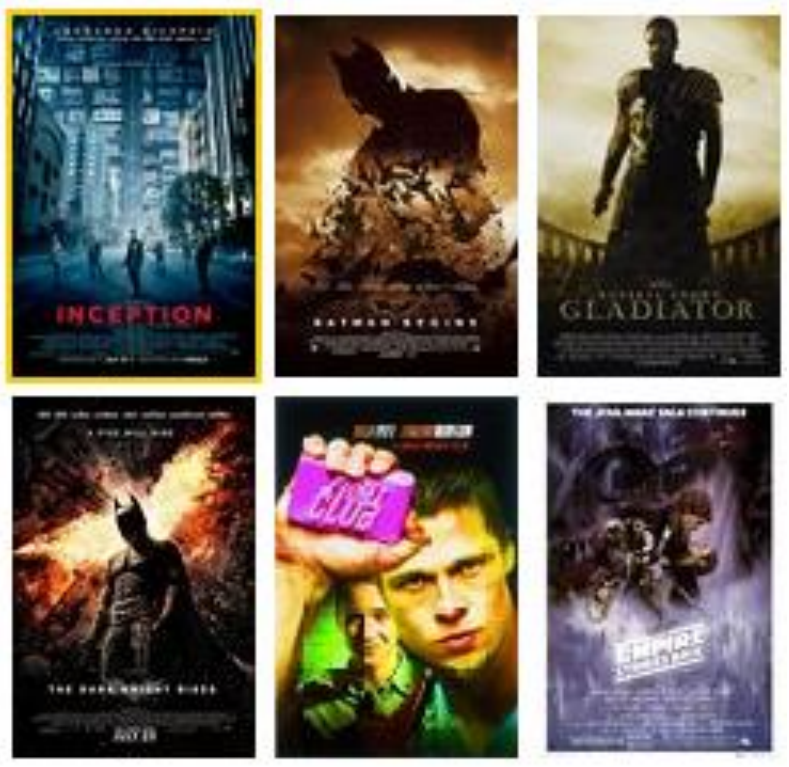

```
0 The Lord of the Rings: The Fellowship of the Ring 0.0
1 The Lord of the Rings 48.6009149697
2 The Lord of the Rings: The Two Towers 86.8926991269
3 The Lord of the Rings: The Return of the King 94.7894125033
4 Dragons II: The Metal Ages 103.207253664
5 The Hunt for Gollum 107.185494347
6 Kamen Rider Den-O & Kiva: Climax Deka 111.784477115
7 Dragons: Fire and Ice 114.026322591
8 How to Train Your Dragon 114.96679629
9 Cho Kamen Rider Den-O & Decade NEO Generations: The Onigashima Battleship 116.949142551
10 The Hunchback of Notre Dame 121.641558313
11 The Valley of Gwangi 124.925882714
12 Legend of the Boneknapper Dragon 126.853517743
13 Tsubasa Chronicle the Movie: The Princess of the Country of Birdcages 126.907288058
14 Chisum 127.261153323
15 George and the Dragon 129.473676726
16 The Black Stallion Returns 130.423277738
17 Pirates of the Caribbean: Dead Man's Chest 130.548380643
18 Dragonslayer 131.849885146
19 The Return of the King 131.876052344
20 Quest for Fire 133.440807331
```

LDA captures better representation for the “The Lord of the Rings: The Fellowship of the Ring” than the Star war movie as we observe smaller distances. Moreover we see some same recommendations for movies. The intuition of the two recommendation methods is very clear in this example: LDA captures same contexts, even with the bag assumption, while the collaborative filtering proposes most known movies but does not have a wide range in the same topic though as LDA manages to capture.

<u>Movie Selected</u>: Amelie

```
0 Amelie 0.0
1 The Discreet Charm of the Bourgeoisie 55.9100483953
2 Bussen 61.2959405662
3 Celine et Julie vont En Bateau 62.7600376841
4 Wonder Boys 64.4956710022
5 The Darjeeling Limited 66.2056767209
6 You've Got Mail 67.3791376489
7 Reversal of Fortune 69.6187586942
8 Inside Monkey Zetterland 70.6390860936
9 Four Nights of a Dreamer 71.0052658037
10 Facing Windows 73.0291710764
11 When a Woman Ascends the Stairs 73.9080591161
12 Things You Can Tell Just by Looking at Her 74.4780489978
13 The Ploughman's Lunch 74.9035749952
14 The Palm Beach Story 75.2531994509
15 The Umbrellas of Cherbourg 76.5052070925
16 Three Seasons 76.5605140931
17 Transit 76.6192407589
18 Where God Left His Shoes 76.8801967299
19 Ping Pong Playa 77.5327543882
20 Twisted Nerve 77.8766514999
```

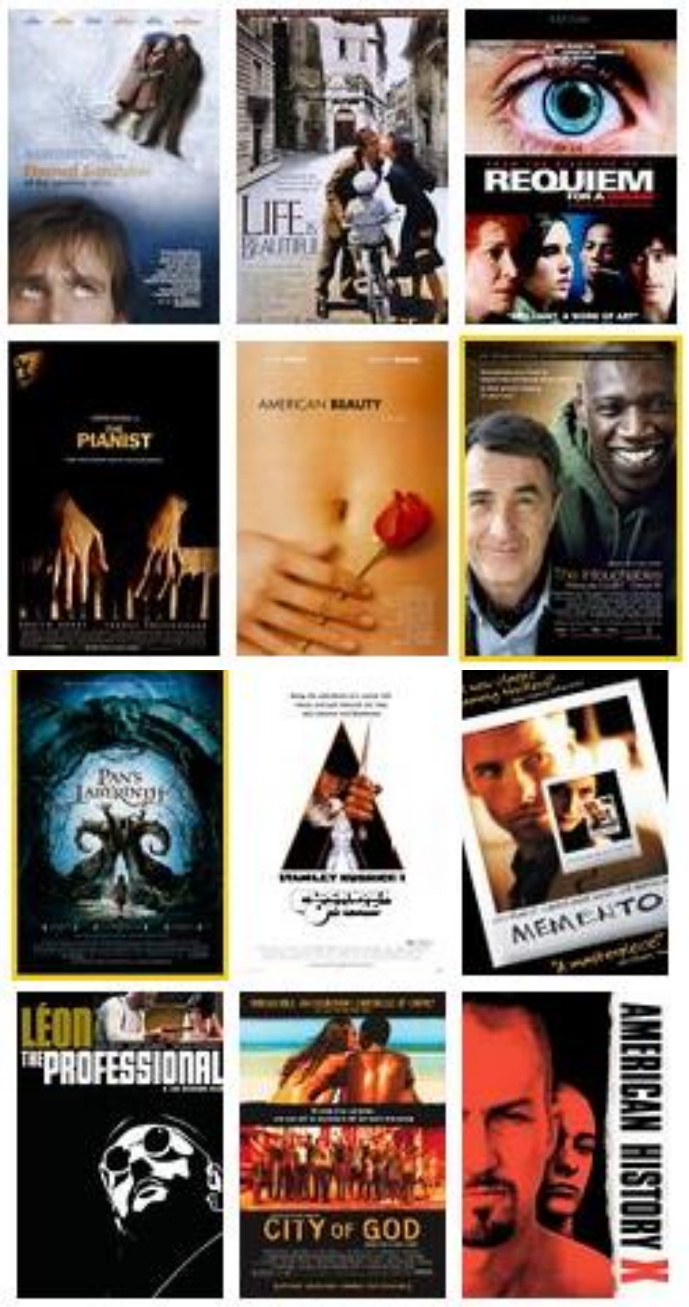

Here, we capture again the same intuition. The two methods do not have similar proposed movies. We should see how the other two algorithms map “Amelie” do estimate why this is happening.

## 6.2 Autoencoder Evaluation

Autoencoder in contrast to LDA does not estimate paragraph vectors (topics) as a distribution over words (See Section 4.2), but the compressed information in the hidden layer of the encoder manage to represent the initial information in a reduces space. Here, this space is a 50-dimensional space, namely 50-hidden topics.

### 6.2.1 Dataset Plot

First we view the initial plot of the 25203 movies in the 2d space:

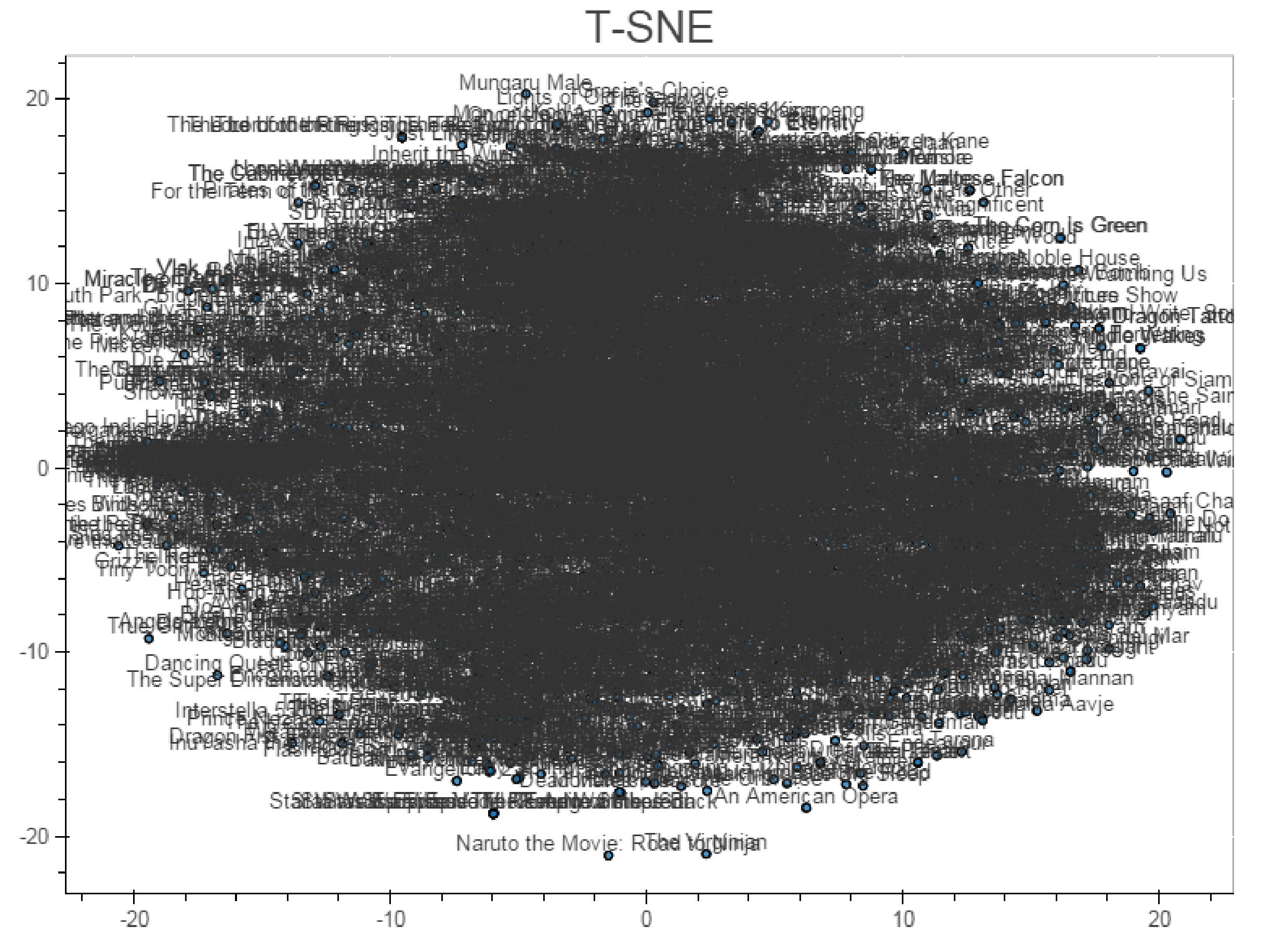

What we observe in relevance to the symmetric LDA is that Autoencoder gives a more dense result. Precisely more movies seems to be mapped in the middle. Let’s see further inside for some more representations in the 2d.

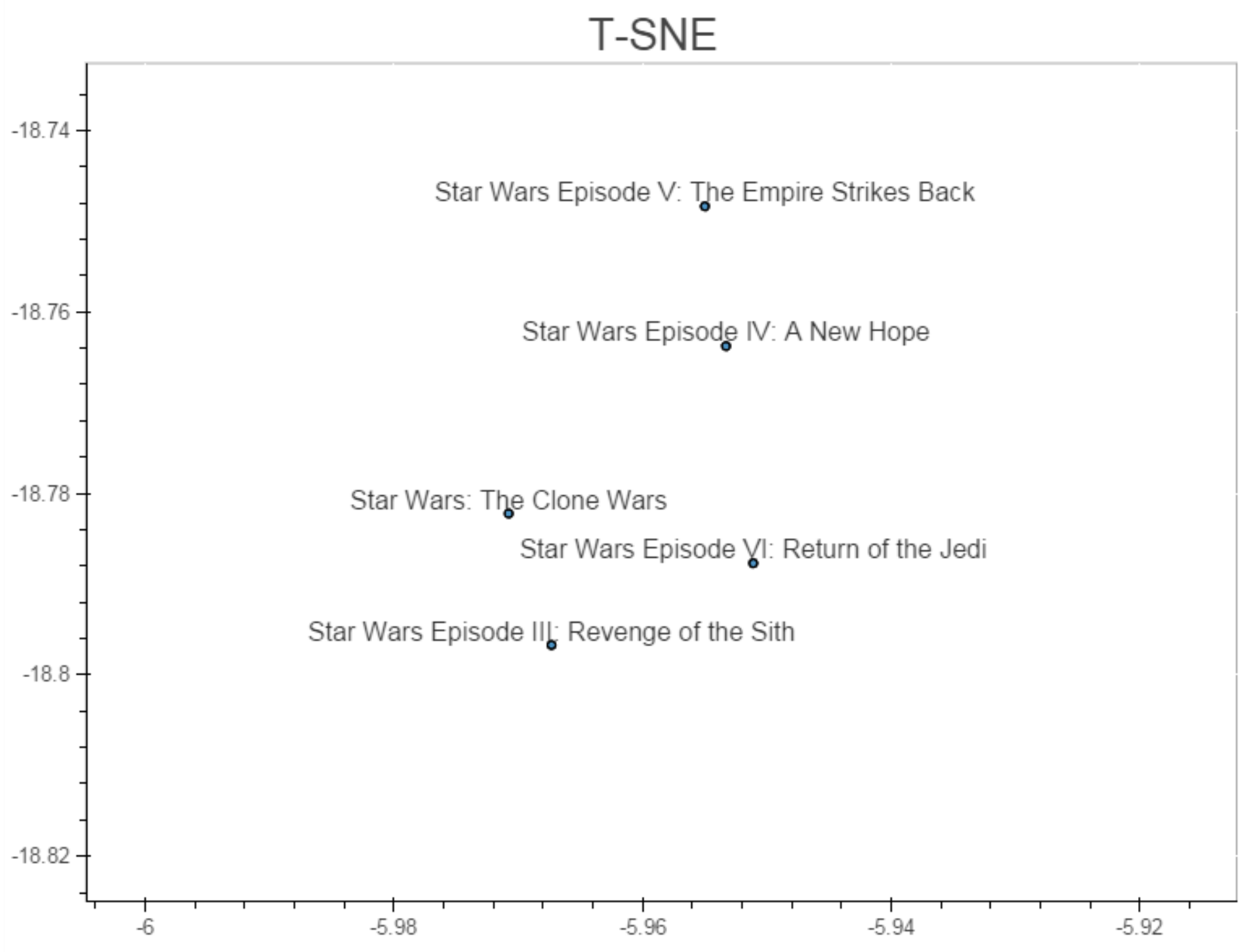
T-SNE
Star Wars Episode V: The Empire Strikes Back
Star Wars Episode IV: A New Hope
Star Wars: The Clone Wars
Star Wars Episode VI: Return of the Jedi
Star Wars Episode III: Revenge of the Sith
-18.74
-18.76
-18.78
-18.8
-18.82
-6
-5.98
-5.96
-5.94
-5.92

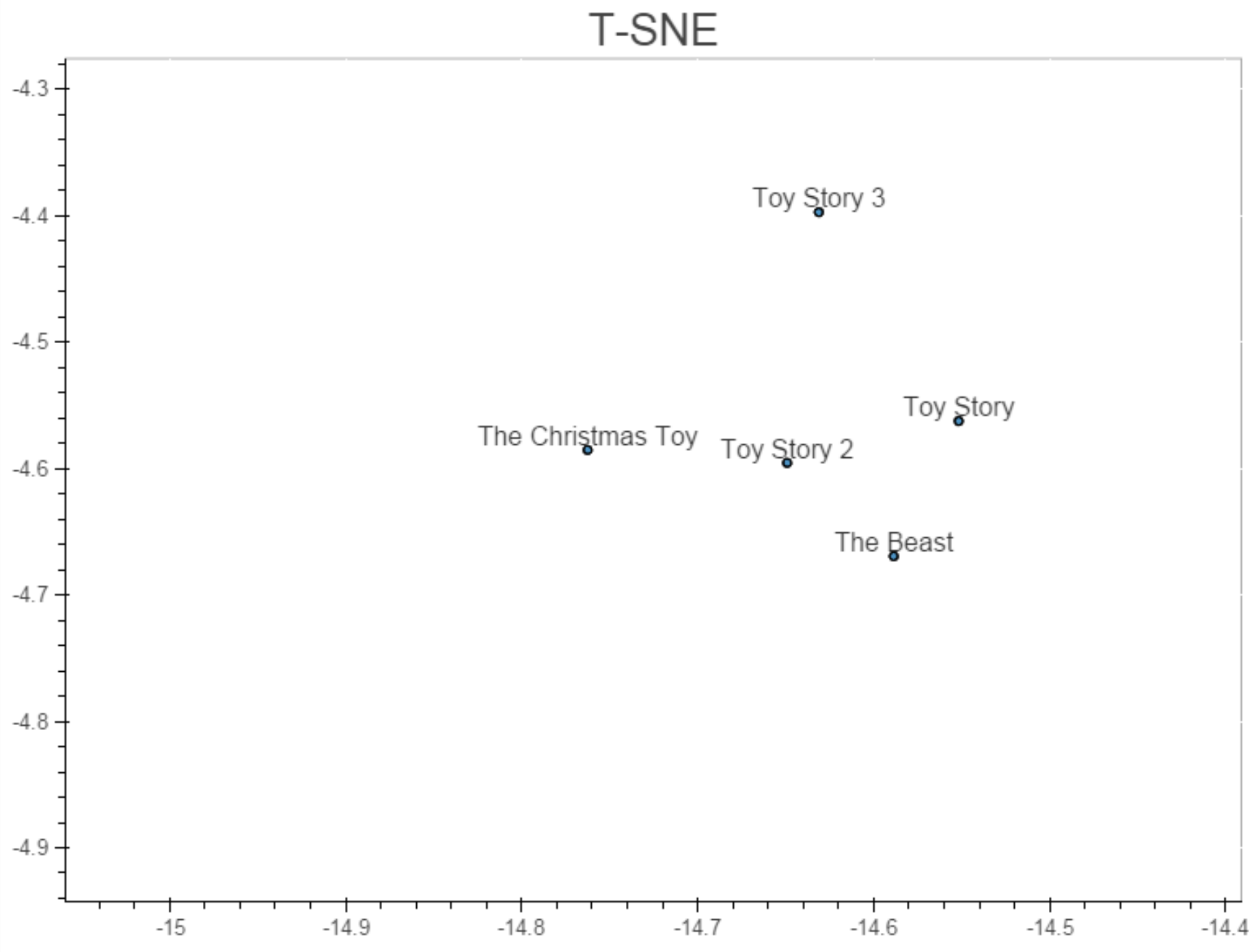

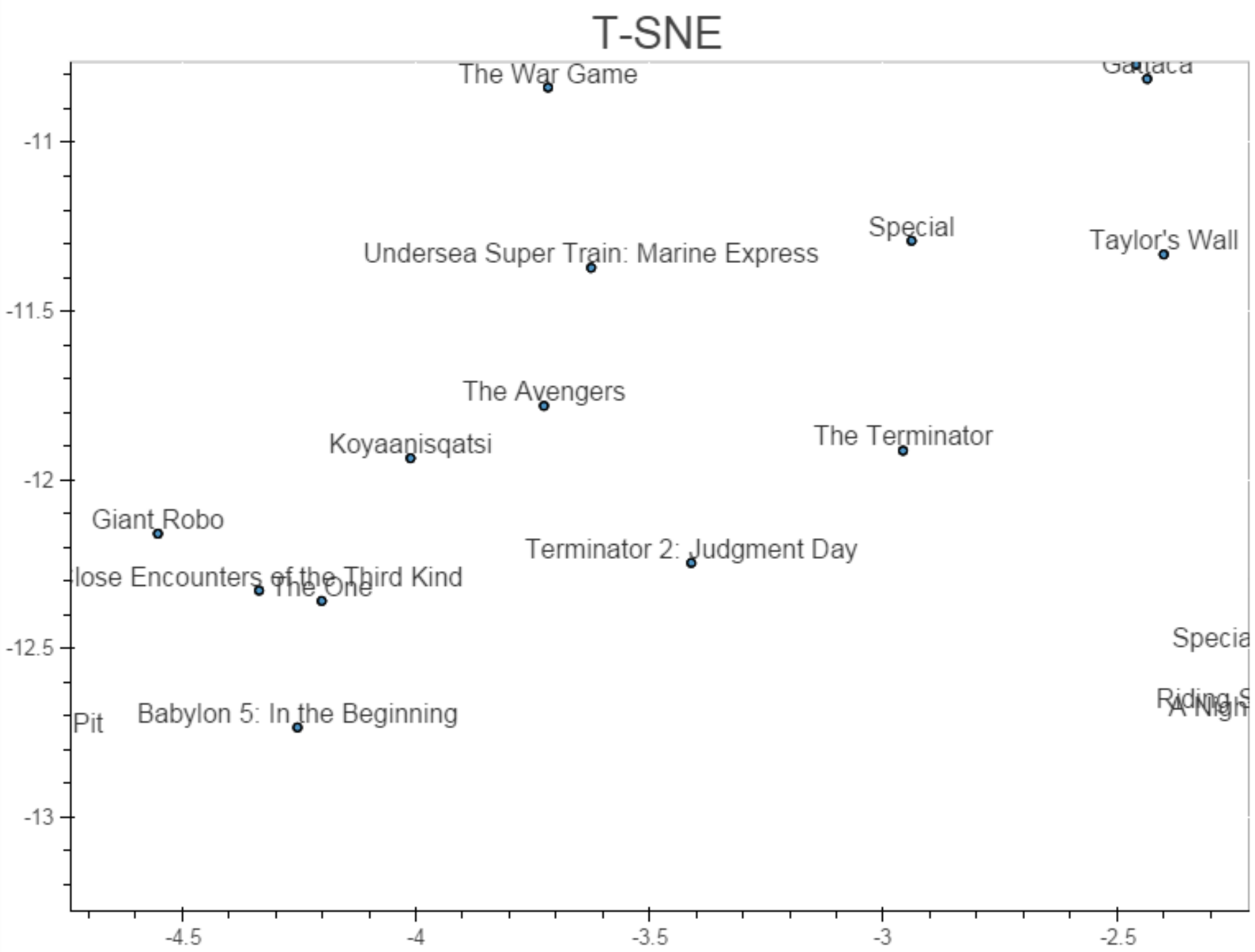

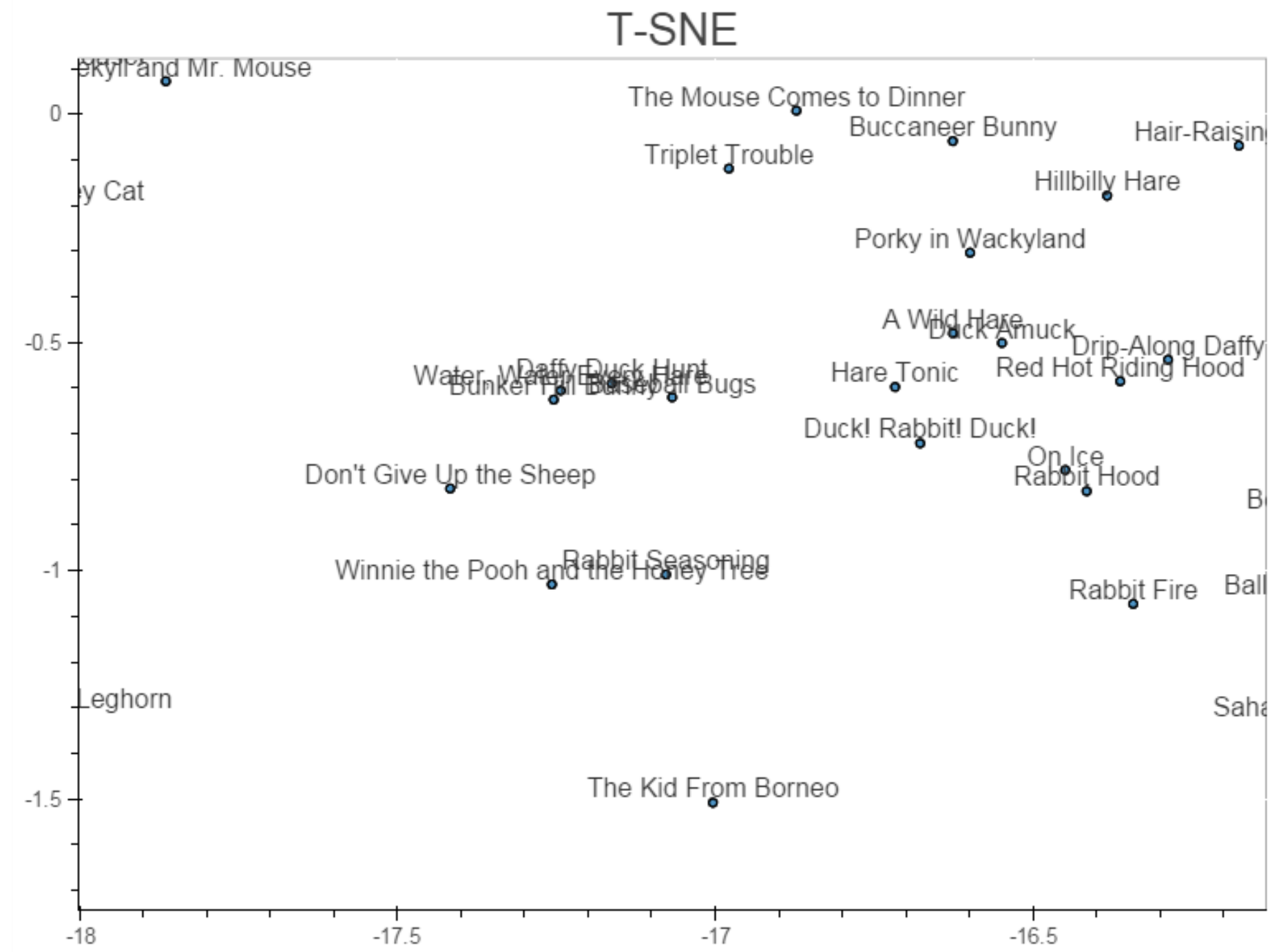

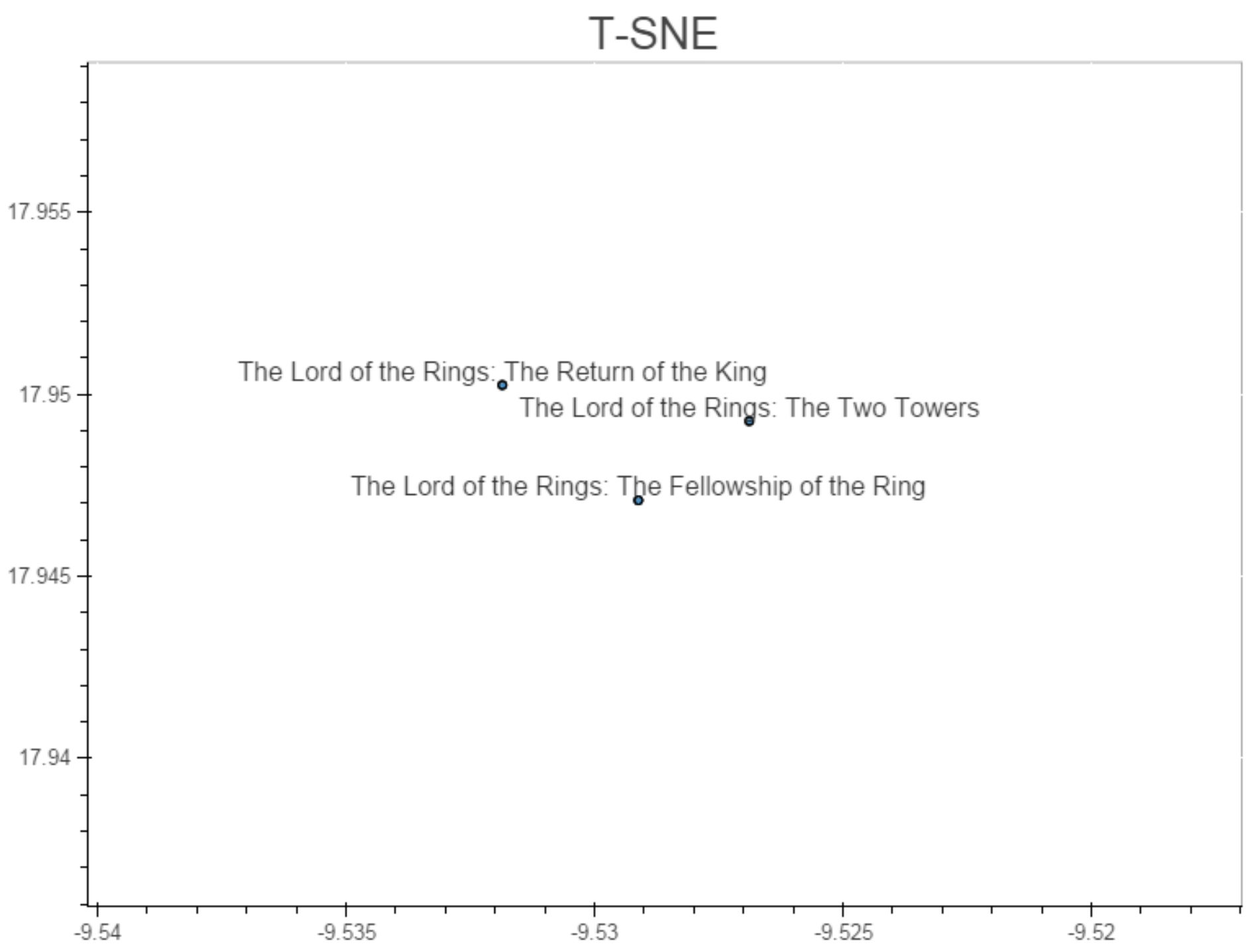
T-SNE
The Lord of the Rings: The Return of the King
The Lord of the Rings: The Two Towers
The Lord of the Rings: The Fellowship of the Ring
17.955
17.95
17.945
17.94
-9.54
-9.535
-9.53
-9.525
-9.52

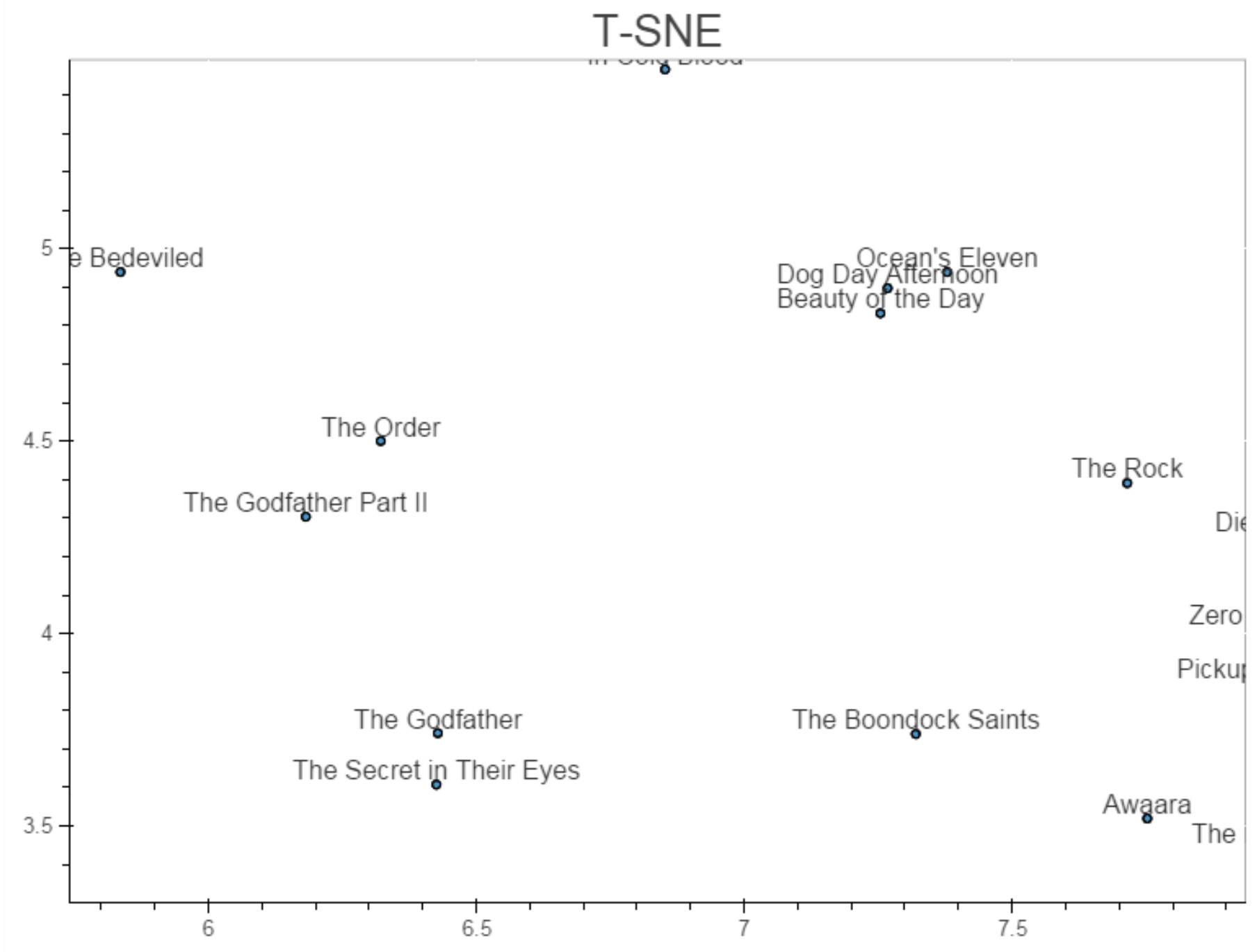

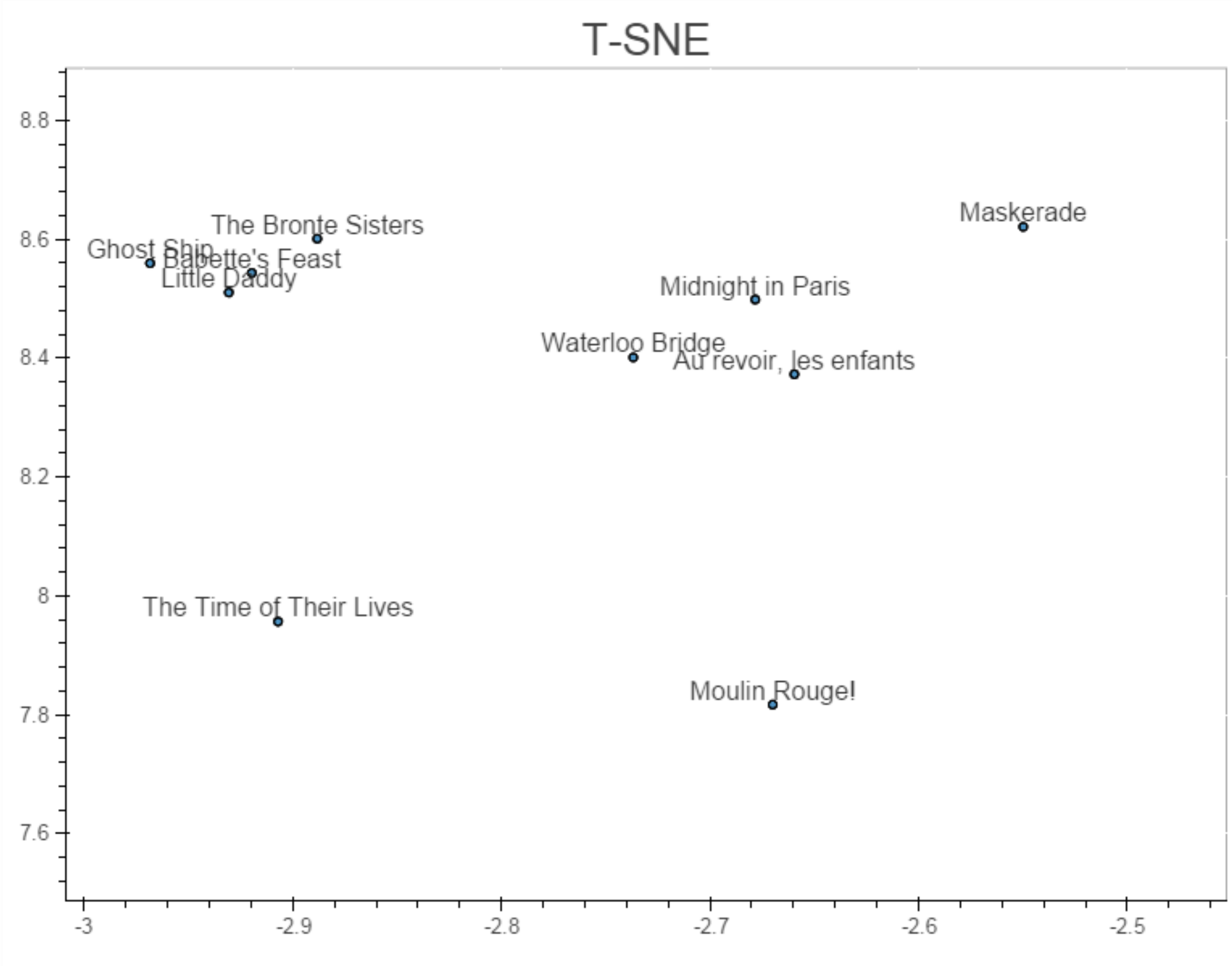

Autoencoder performs excellent representations of the features. Closely related movies such as “Star Wars”, “The lord of the rings”, “Toy Story”, are really near to each and they are encircled with movies that share the same semantics. We can better view this interpretation with the KNN.

### 6.2.2 Movies Recommendation

With KNN we can observe in detail the distances between certain movies and as a matter of fact given a certain movie propose the nearest K of it. Here, as in LDA K=20.

Movie Selected: Star Wars Episode V: The Empire Strikes Back

```
0  Star  Wars Episode V: The Empire Strikes Back 0.0
1 Star Wars Episode IV: A New Hope 2.77936185375
2 Lego Star Wars: The Quest for R2-D2 3.46439905145
3 LEGO Star Wars: Revenge of the Brick 3.53798449792
4 Star Wars Episode I: The Phantom Menace 3.675081325
5   batteries  not included 3.6847448612
6 Star Wars Episode II: Attack of the Clones 3.75624609865
7 Earth Star Voyager 3.76074996035
8 Journey to the Far Side of the Sun 3.76768187463
9 Bicentennial Man 3.78933278972
10 Alien Planet 3.85871359116
11 Star Wars Episode VI: Return of the Jedi 3.92073625569
12 Star Wars Episode III: Revenge of the Sith 3.95813588718
13 Justice League: New Frontier 3.98570175975
14 Cool Hand Luke 3.98693698094
15 Danny Deckchair 4.00230381175
16 Second Chance 4.00796022426
17 The War of the Roses 4.01018942698
18 Superman 4.01223618789
19 Threads 4.03645776123
20 De Lift 4.0557159171
```

Movie Selected: The Lord of the Rings: The Fellowship of the Ring

```
0 The Lord of the Rings: The Fellowship of the Ring 0.0
1 The Lord of the Rings: The Two Towers 3.07142153566
2 The Lord of the Rings 3.26138545554
3 The Lord of the Rings: The Return of the King 4.05568554019
4 The Return of the King 4.49416952538
5 The Hunt for Gollum 4.81783470383
6 Indiana Jones and the Temple of Doom 5.80077764753
7 The Magnificent  Seven  Ride 5.83182547582
8 Tarzan Triumphs 5.84084121763
9 Dragonslayer 5.87472325742
10 Ali Baba and the Forty Thieves 5.90008406263
11 Zorro 5.90544380931
12 Captain from Castile 5.92011896736
13 Snow White and the Huntsman 6.0019117788
14 Marie Antoinette 6.01460156822
15 The Four Feathers 6.07029940862
16 Brotherhood of the Wolf 6.07156992389
17 Gettysburg 6.12812381907
18 National Treasure 6.14079189026
19 Richard III 6.15669462358
20 Cleopatra 6.17270905918
```

Movie Selected: Amelie

```
0 Amelie 0.0
1 Being There 2.58956116322
2 Look Back in Anger 2.6785110063
3 To Rome with Love 2.71266105625
4  Vegas  Vacation 2.773942418
5 Fellini: I'm a Born Liar 2.80881142599
6 The Bucket List 2.88796922781
7 The Black Balloon 2.89357901851
8 Danny Deckchair 2.90808879966
9 Millions 2.92209675515
10 Love Actually 2.92423504692
11 Boarding Gate 2.95876512639
12 Who's Afraid of Virginia Woolf? 2.96463249552
13 Carry On Abroad 2.98562387352
14  Living  in Oblivion 2.9890768154
15 Hereafter 2.99503482127
16 A Girl Cut in Two 3.01475843945
17 Smithereens 3.02255117099
18 Tea and Sympathy 3.0266808704
19 In America 3.02889867743
20 Heavenly Creatures 3.0339444027
```

Movie Selected: The God Father

```
0 The Godfather 0.0
1 The Mack 2.90157428928
2 The Godfather Part III 3.07052919217
3 Goodfellas 3.12880415781
4 The Lucky One 3.26372181318
5 American History X 3.27191367232
6 Sneakers 3.28930866674
7 Collateral 3.29325111704
8 Heathers 3.31201638897
9 Redbelt 3.31347578031
10 Single White Female 3.35584252313
11 Boarding Gate 3.36737577633
12 Shame 3.37487810641
13 Anne of the Thousand Days 3.37563495639
14  Mad  Love 3.3773799991
15 I, the Jury 3.38704935101
16 Flightplan 3.39729700857
17 Hush 3.42476777106
18 The Good Shepherd 3.42521700182
19 Saboteur 3.42841387069
20 The Last Supper 3.43277868015
```

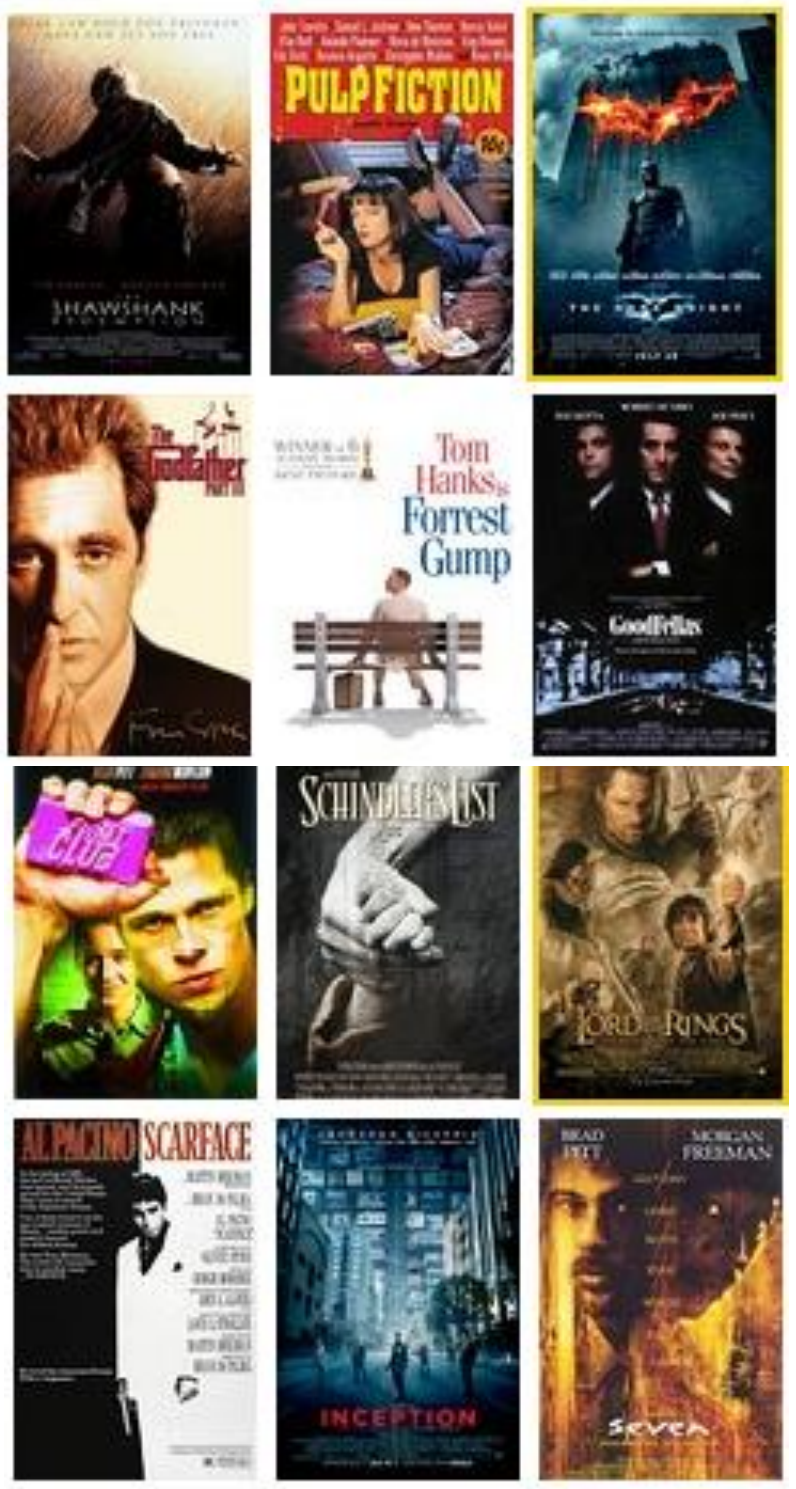

## 6.3 Comparative evaluation of models

To begin with, as we have review some movie's representation from both models, we testify what we thought we would from the theory: Autoencoder as it uses paragraph vectors outperforms LDA. Autoencoder captures more precisely the thematic structure, the semantic relations through documents. This can be further seen in parallel below:

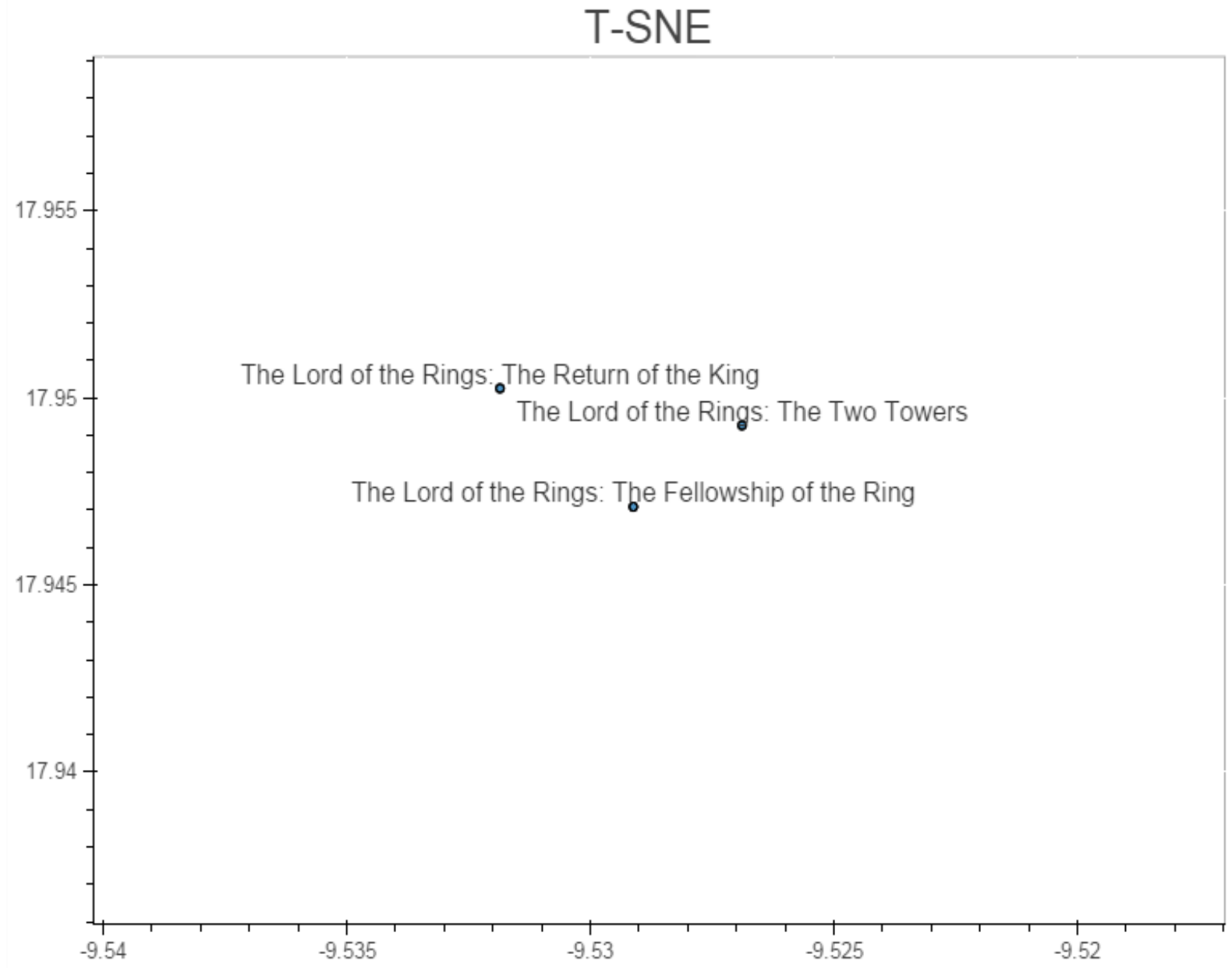

*TSNE-Autoencoder*

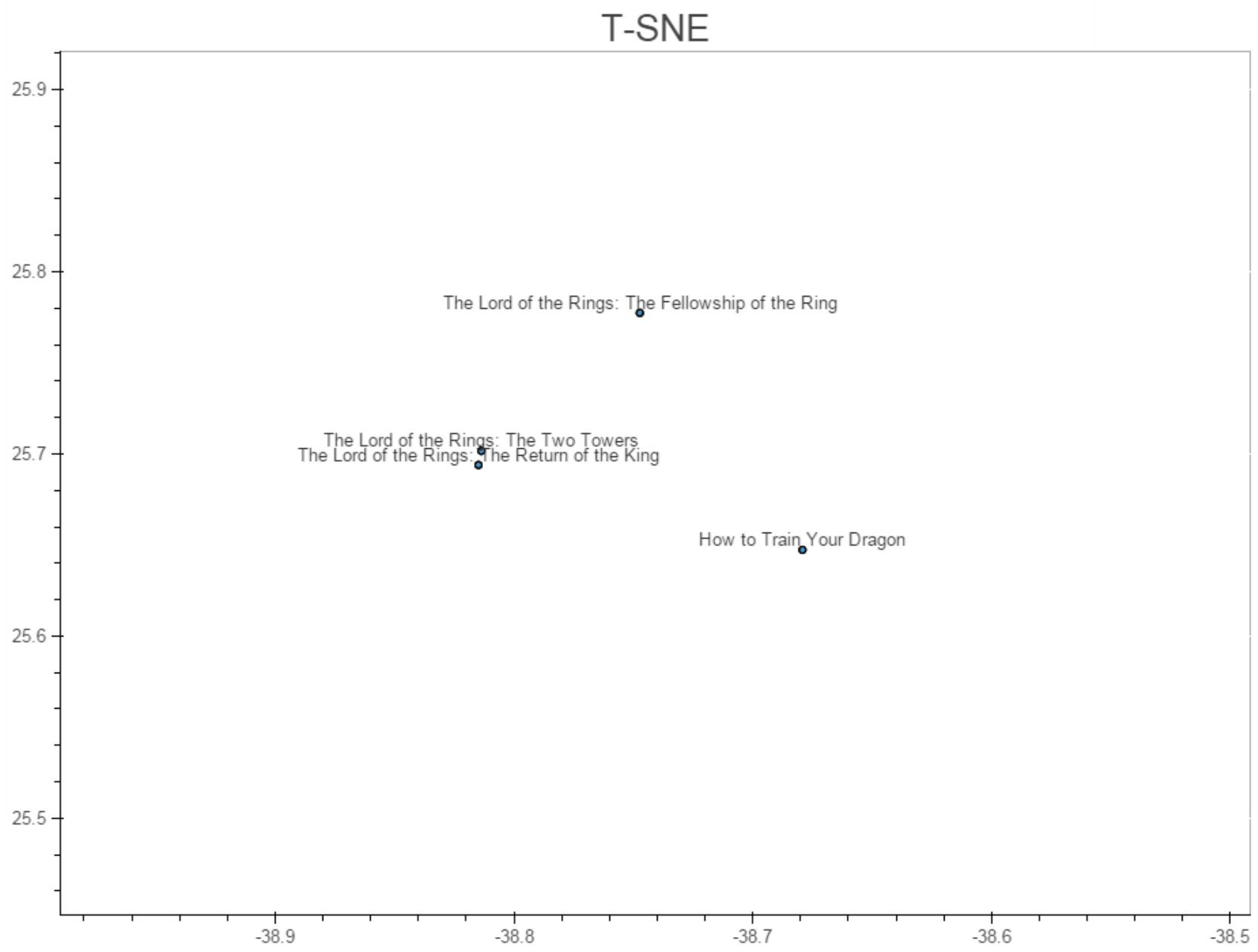

*TSNE-sLDA*

What we can see in the above Figures is that even though both Autoencoder and LDA represents the same movies as “nearest” Autoencoder represents a denser, more coherent representation. We further see that in their KNN distances:

```
0 The Lord of the Rings: The Fellowship of the Ring 0.0
1 The Lord of the Rings: The Two Towers 3.07142153566
2 The Lord of the Rings 3.26138545554
3 The Lord of the Rings: The Return of the King 4.05568554019
4 The Return of the King 4.49416952538
5 The Hunt for Gollum 4.81783470383
6 Indiana Jones and the Temple of Doom 5.80077764753
7 The Magnificent  Seven  Ride 5.83182547582
8 Tarzan Triumphs 5.84084121763
9 Dragonslayer 5.87472325742
10 Ali Baba and the Forty Thieves 5.90008406263
11 Zorro 5.90544380931
12 Captain from Castile 5.92011896736
13 Snow White and the Huntsman 6.0019117788
14 Marie Antoinette 6.01460156822
15 The Four Feathers 6.07029940862
16 Brotherhood of the Wolf 6.07156992389
17 Gettysburg 6.12812381907
18 National Treasure 6.14079189026
19 Richard III 6.15669462358
20 Cleopatra 6.17270905918
```

*KNN-Autoencoder*

```
0 The Lord of the Rings: The Fellowship of the Ring 0.0
1 The Lord of the Rings 48.6009149697
2 The Lord of the Rings: The Two Towers 86.8926991269
3 The Lord of the Rings: The Return of the King 94.7894125033
4 Dragons II: The Metal Ages 103.207253664
5 The Hunt for Gollum 107.185494347
6 Kamen Rider Den-O & Kiva: Climax Deka 111.784477115
7 Dragons: Fire and Ice 114.026322591
8 How to Train Your Dragon 114.96679629
9 Cho Kamen Rider Den-O & Decade NEO Generations: The Onigashima Battleship 116.949142551
10 The Hunchback of Notre Dame 121.641558313
11 The Valley of Gwangi 124.925882714
12 Legend of the Boneknapper Dragon 126.853517743
13 Tsubasa Chronicle the Movie: The Princess of the Country of Birdcages 126.907288058
14 Chisum 127.261153323
15 George and the Dragon 129.473676726
16 The Black Stallion Returns 130.423277738
17 Pirates of the Caribbean: Dead Man's Chest 130.548380643
18 Dragonslayer 131.849885146
19 The Return of the King 131.876052344
20 Quest for Fire 133.440807331
```

*KNN-symLDA*

IMDB recommendations

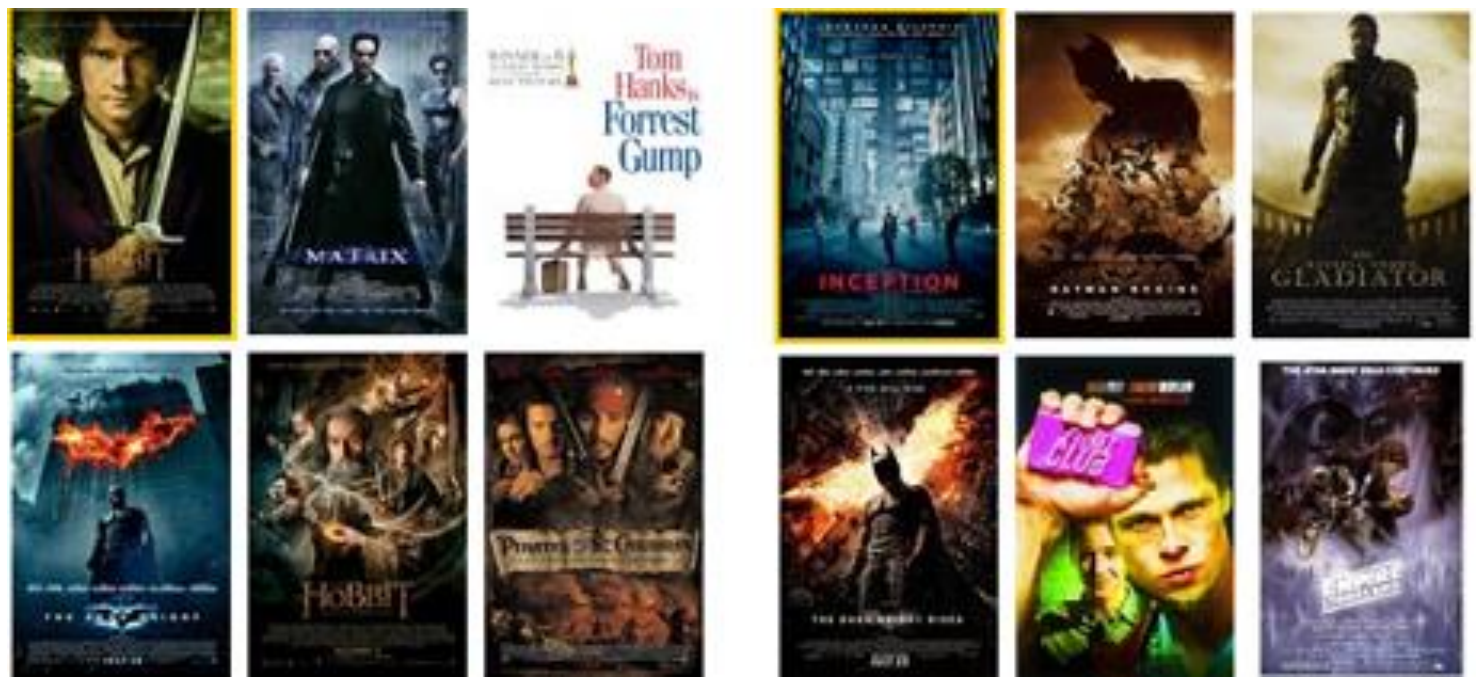

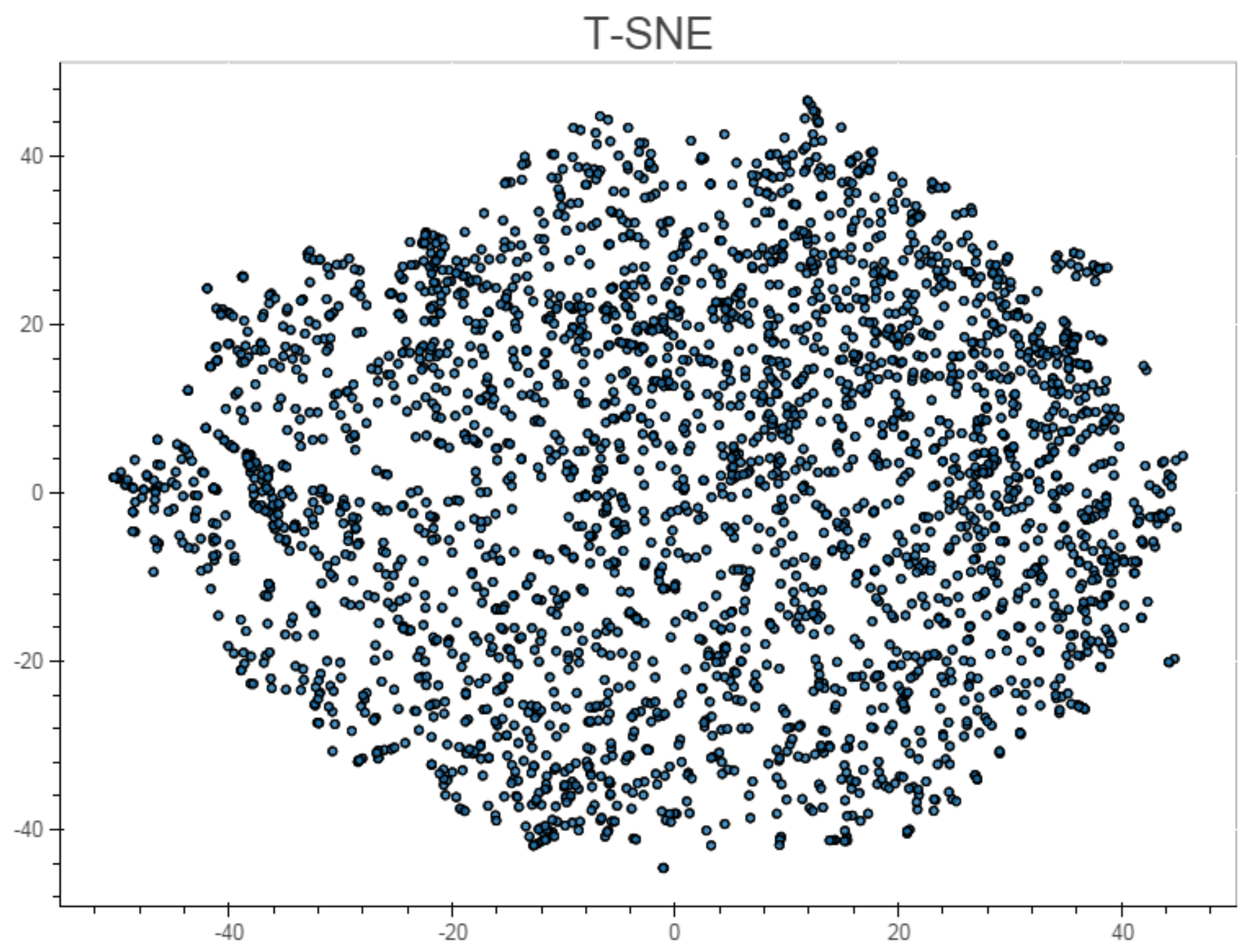

*Symmetric LDA movies representation*

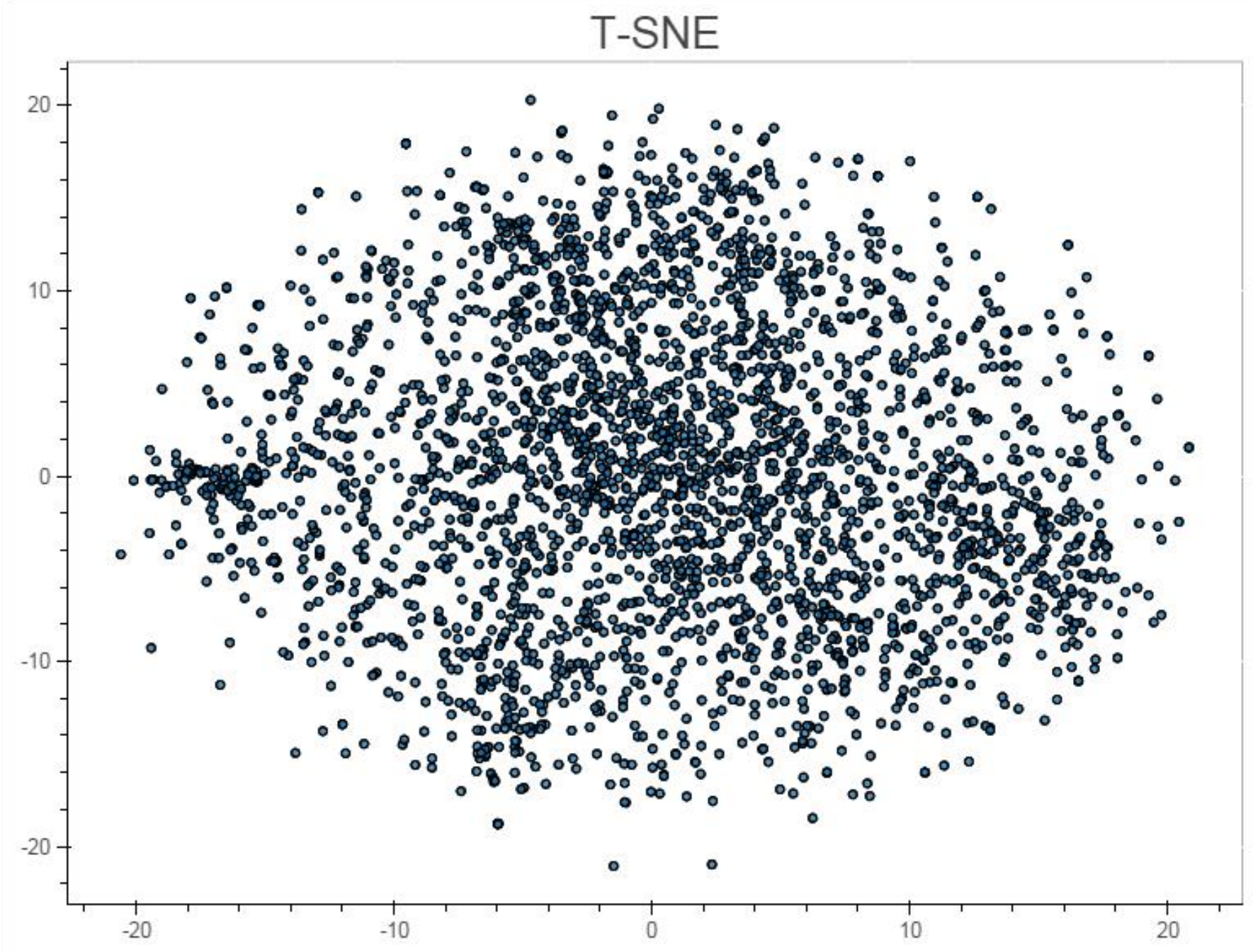

*Autoencoder movies representation*

# Chapter 7

# 7. Conclusions

## 7.1 Contributions and Future Work

In this thesis we show the differences between the LDA probabilistic and the Autoencoder neural net models in their feature extraction in a movie database, showing the relationships of the movies as the distances among them. This is captured by representing the distances of the extracted data either by KNN or T-SNE. The closer a movie is to another, the most likable to be close related and thus recommended.
The results show Autoencoder to outperform LDA as its longest distance is only the 8% of the shortest distance on LDA model (6.17 – 48.60). Subsequently, the paragraph vectors represent the data according to their hidden thematic structure, while LDA representations are words more often seen together. Meanwhile, it is exact this framework of Autoencoder that gives so close variables to close meanings that tends to leave behind movies for recommendation just because the question movie and some related in our concept movies are so strong, that do not meet each other in KNN. For example the Autoencoder do not propose the Pirates of the Caribbean as LDA does mainly because of their strong semantics that keeps more relative movies tight, too far though from other categories. Moreover, searching the TSNE map, both for Autoencoder and for LDA, we find that topics like those presented (Star wars, Lord of the Rings, Harry Potter) usually hosted to the outer layers of the map, declaring in this way the special character-plots of such movies.
Meanwhile, the LDA topic-word distribution is able to capture sub-genres that could not be expressed otherwise, as also specific attributes in a topic as discussed in Section 6.1.1.
Last the feature representation of the extracted features of the two models are compared to the respective IMDB recommending movies (IMDB uses collaborative filtering for its recommendation system). What we find out it that all three models, namely the probabilistic model, the Neural Networks and the collaborative filtering, even though manage to reflect astonishing relationships with the Autoencoder outperforming the other two, all three models were not able by their own to capture the best results. Consequently, what we show is that the best recommending system would be a combination of the three models. Moreover, as private systems with no tracking now gain increased space, the recommendation systems held in content is a field of much interest. The extension of the shown methods for feature extraction mainly used in recommendations systems is a project to be examined in the near future.

## Appendix A

```
import logging

import numpy as np

from gensim import corpora, models, similarities
from gensim.models import LdaModel

from gensim.models.doc2vec import LabeledSentence
from gensim.matutils import Sparse2Corpus
from sklearn.feature_extraction.text import CountVectorizer


logging.basicConfig(format='%(asctime)s : %(levelname)s : %(message)s', level=logging.INFO)


class LabeledLineSentence(object):
    def __init__(self, filename):
        self.filename = filename

    def __iter__(self):
        for uid, line in enumerate(open(self.filename)):
            yield LabeledSentence(words=line.split(), labels=['MOV_%s' % uid])


with open('plots_5.txt') as f:
    sentences = f.readlines()
dictionary = corpora.Dictionary(line.lower().split() for line in open('plots_5.txt'))
class MyCorpus(object):
    def __iter__(self):
        for line in open('plots_5.txt'):
            yield dictionary.doc2bow(line.lower().split())

corpus = MyCorpus()

num_topics = 50
model = LdaModel(corpus, id2word=dictionary, #corpus=vec_corpus, id2word=id2word,
                 num_topics=num_topics,
                 #  alpha ='asymmetric',
                  alpha ='symmetric',
                 # eta=0.1,
                 chunksize=2000,
                 iterations=100,
                 passes=20,
                 #eval_every=5
                 # workers=2
)

resultlist = np.array(model.inference(corpus)[0])
print resultlist  shape
np.savetxt('lda_symmetric_20.txt', resultlist, delimiter=',')

resultlist = model.show_topics(num_topics=num_topics, num_words=20, formatted=False)
out_str = ""
for i, words in enumerate(resultlist):
    out_str += "Topic %d\n" % (i)
    for perc, word in words:
        out_str += "%.3f %s\n" % (perc, word)

    out_str += "\n"

with open("lda_symmetric_20_words.txt", "w") as text_file:
    text_file.write(out_str)
```

Appendix B

```
from gensim.models import Doc2Vec
from gensim.models.doc2vec import LabeledSentence
import numpy as np


class LabeledLineSentence(object):
    def __init__(self, filename):
        self.filename = filename

    def __iter__(self):
        for uid, line in enumerate(open(self.filename)):
            yield LabeledSentence(words=line.split(), labels=['MOV_%s' % uid])


# sentences = LabeledLineSentence('PlotsWithoutStopWords_filtered.txt')
sentences = LabeledLineSentence('plots_5.txt')

model = Doc2Vec  alpha =0.025, min_alpha=0.025,
                size=50, window=5, min_count=5,
                dm=1,
                workers=8,
                sample=1e-5)  # use fixed learning rate
model.build_vocab(sentences)

for epoch in range(500):
    try:
        print 'epoch %d' % (epoch)
        model.train(sentences)
        # if epoch % 2 == 0:
            # model.alpha -= 0.0005  # decrease the learning rate
        model.alpha *= 0.99
        model.min_alpha = model.alpha  # fix the learning rate, no decay
        if model.alpha < 1e-3:
            break
    except (KeyboardInterrupt, SystemExit):
        break


# store the model to mmap-able files
model.save('doc2vec_out_50.model')


# load the model back
model = Doc2Vec.load('doc2vec_out_50.model')


resultlist = []
idx = []
for i in range(25203): #41796
    string = "MOV_" + str(i)
    # idx.append(string)
    resultlist.append(model[string].tolist())

resultlist = np.array(resultlist)
np.savetxt('autoencoder_50.txt', resultlist, delimiter=',')
```

Appendix C

# Imports

```
# -*- coding: utf-8 -*-
import numpy as np
import unicodedata

import pyprind

from bokeh.plotting import *
import pandas as pd
output_notebook()

# Import the imdb package.
import imdb
```

BokehJS successfully loaded.

# Load Feature Vectors (processed)

```
top_movies = -1
features = 50

lda_symmetric_tsne = np.loadtxt("data/lda_symmetric_"+str(features)+"_tsne.txt", delimiter=',')
lda_symmetric = np.loadtxt("data/lda_symmetric_"+str(features)+".txt", delimiter=',')
lda_asymmetric_tsne = np.loadtxt("data/lda_asymmetric_"+str(features)+"_tsne.txt", delimiter=',')
lda_asymmetric = np.loadtxt("data/lda_asymmetric_"+str(features)+".txt", delimiter=',')
autoencoder_tsne = np.loadtxt("data/autoencoder_"+str(features)+"_tsne.txt", delimiter=',')
autoencoder = np.loadtxt("data/autoencoder_"+str(features)+".txt", delimiter=',')

with open('data/titles_5.txt') as f:
    titles = np.array([line.strip() for line in f.readlines()])
with open('data/plots_5.txt') as f:
    plots = np.array([line.strip() for line in f.readlines()])
with open('data/imdb_ratings_5.txt') as f:
    imdb_ratings = np.array([line.strip() for line in f.readlines()])
```

```
# data = lda_asymmetric_tsne
# data_full = lda_asymmetric

# data = autoencoder_tsne
# data_full = autoencoder

data = lda_symmetric_tsne
data_full = lda_symmetric
```

## Plots

```
num_plot = 3000

TOOLS = "pan,wheel_zoom,box_zoom,reset,resize,save"
p = figure(tools=TOOLS, toolbar_location="left", logo="grey", plot_width=800)
p.title = "T-SNE"
# p.background_fill= "#cccccc"

p.circle(data[:num_plot,0], data[:num_plot,1], size=5, line_color="black", fill_alpha=0.8)

p.text(data[:num_plot,0], data[:num_plot,1],
   text=titles[:num_plot], text_color="#333333",
    text_align="center", text_font_size="10pt")

p.xaxis.axis_label="atomic weight (amu)"
p.yaxis.axis_label="density (g/cm^3)"
p.grid.grid_line_color="white"

show(p)
```

## k-Nearest Neighbors

```
from sklearn.neighbors import NearestNeighbors
nbrs = NearestNeighbors(n_neighbors=21).fit(data_full) #, algorithm='brute', metric='cosine'
knn_distances, knn_indices = nbrs.kneighbors(data_full)
```

```
# movie_id = titles.tolist().index("Star Wars Episode V: The Empire Strikes Back")
# movie_id = titles.tolist().index("The Lord of the Rings: The Fellowship of the Ring")
# movie_id = titles.tolist().index("Amelie")
movie_id = titles.tolist().index("The Godfather")

for i in range(knn_indices.shape[1]):
    print i, titles[knn_indices[movie_id, i]], knn_distances[movie_id, i]
```